\documentclass[runningheads]{llncs}
 
\usepackage{eccv}

\usepackage{eccvabbrv}

\usepackage{graphicx}
\usepackage{booktabs}

\usepackage[accsupp]{axessibility}  

\usepackage{hyperref}

\usepackage{standalone}
\newcommand{\secondarytitle}[1]{%
    \begin{center}
        \Large\textbf{#1}\par
        \end{center}
}

\usepackage{orcidlink}
\usepackage{graphicx}
\usepackage{amsmath}
\usepackage{amssymb}

\usepackage{dsfont} 

\usepackage[utf8]{inputenc} 
\usepackage[T1]{fontenc}    
\usepackage{hyperref}       
\usepackage{url}            
\usepackage{booktabs}       
\usepackage{amsfonts}       
\usepackage{nicefrac}       
\usepackage{microtype}      
\usepackage[dvipsnames]{xcolor}         

\usepackage{amssymb}
\usepackage{pifont}
\usepackage{fixmetodonotes}
\usepackage{colortbl}
\usepackage{rotating}
\usepackage{scalerel}

\usepackage{multirow}
\usepackage{todonotes}
\usepackage{placeins}
\usepackage{wrapfig}

\usepackage{bm}
\usepackage{adjustbox}
\usepackage{array}
\newcolumntype{R}[2]{%
    >{\adjustbox{angle=#1,lap=1.3\width-(#2)}\bgroup}%
    l%
    <{\egroup}%
}

\usepackage[linesnumbered,ruled,vlined]{algorithm2e}
\usepackage{algpseudocode}
\usepackage{flushend} 
\makeatletter 
\@namedef{ver@everyshi.sty}{}
\makeatother
\usepackage{tikz}
\usetikzlibrary{positioning,shapes}

\usepackage[]{tikz}
\usepackage{pgfplots}
\pgfplotsset{compat=1.18}
\usetikzlibrary{intersections}
\usepgfplotslibrary{fillbetween}
\usepackage[nomessages]{fp}
\usepackage{pgfkeys}
\newenvironment{customlegend}[1][]{%
    \begingroup
    \csname pgfplots@init@cleared@structures\endcsname
    \pgfplotsset{#1}%
}{%
    \csname pgfplots@createlegend\endcsname
    \endgroup
}%
\def\addlegendimage{\csname pgfplots@addlegendimage\endcsname}

\definecolor{MyGreen}{RGB}{0, 180, 0}
\definecolor{MyRed}{RGB}{180, 0, 0}
\definecolor{MyOrange}{RGB}{180, 180, 0}
\definecolor{MyBlue}{RGB}{30, 0, 180}
\definecolor{MyGrey}{RGB}{82.75, 82.75, 82.75}

\definecolor{skcar}{RGB}{100,150,245}
\definecolor{skbicycle}{RGB}{255, 200, 0}
\definecolor{skmotorcycle}{RGB}{255, 120, 0}
\definecolor{sktruck}{RGB}{80, 30, 180}
\definecolor{skotherv}{RGB}{0, 0, 255}
\definecolor{skpedestrian}{RGB}{255,30,30}
\definecolor{skdrivable}{RGB}{255, 0,255}
\definecolor{sksidewalk}{RGB}{75, 0,75}
\definecolor{skterrain}{RGB}{150, 240, 80}
\definecolor{skvegetation}{RGB}{0, 175, 0}
\definecolor{skbuilding}{RGB}{255, 200, 0}

\definecolor{tsne_source}{rgb}{0.86, 0.3712, 0.33999999999999997}
\definecolor{tsne_target}{rgb}{0.33999999999999997, 0.8287999999999999, 0.86}

\newcommand{\cmark}{{\textcolor{MyGreen}{\ding{51}}}}%
\newcommand{\xmark}{{\textcolor{MyRed}{\ding{55}}}}%
\newcommand{\oxmark}{{\textcolor{MyOrange}{\ding{58}}}}%

\newcommand{\synth}{{SL}}
\newcommand{\sposssyn}{{SP$_{13}$}}
\newcommand{\spossns}{{SP$_{6}$}}
\newcommand{\pdns}{{PD$_{8}$}}
\newcommand{\wyns}{{WO$_{10}$}}

\newcommand{\sk}{{SK}} 
\newcommand{\ns}{{NS}} 
\newcommand{\skns}{{SK$_{10}$}} 
\newcommand{\sksyn}{{SK$_{19}$}}

\newcommand{\nstosk}{{\DAsetting{\ns}{\skns}}}
\newcommand{\synthtosk}{{\DAsetting{\synth}{\sksyn}}}
\newcommand{\synthtoposs}{{\DAsetting{\synth}{\sposssyn}}}
\newcommand{\nstoposs}{{\DAsetting{\ns}{\spossns}}}

\newcommand{\nstopd}{{\DAsetting{\ns}{\pdns}}}
\newcommand{\nstowy}{{\DAsetting{\ns}{\wyns}}}

\newcommand{\methodstop}{\textbf{TTYD}$_\textit{stop}$}
\newcommand{\methodvalid}{\textbf{TTYD}$_\textit{valid}$}
\newcommand{\methodcore}{\textbf{TTYD}$_\textit{core}$}
\newcommand{\method}{\textbf{TTYD}}

\newcommand{\titletext}{Train Till You Drop: Towards Stable and Robust Source-free Unsupervised 3D Domain Adaptation}

\newcommand{\perf}[1]{{{#1}}}

\newcommand{\ssp}[1]{\,{#1}\,}

\newcommand{\src}{{\mathsf{s}}}
\newcommand{\tgt}{{\mathsf{t}}}

\newcommand{\discrim}{{\mathsf{discrim}}}
\newcommand{\simsrc}{{\mathsf{simsrc}}}
\newcommand{\model}{f}
\newcommand{\Models}{F}
\newcommand{\pmodel}[1]{\model[#1]}
\newcommand{\othermodel}{g}
\newcommand{\damodel}{g}
\newcommand{\param}{\theta}
\newcommand{\pt}{p}
\newcommand{\pc}{P}

\newcommand{\data}{x}
\newcommand{\dataset}{{\mathcal{X}}}
\newcommand{\nclasses}{K}
\newcommand{\loss}{{\mathcal{L}}}
\newcommand{\entropy}{H}
\newcommand{\distrib}{D}
\newcommand{\interval}[1]{[\![#1]\!]}

\newcommand{\KL}{\text{KL}}
\newcommand{\caa}{A}
\newcommand{\stopt}{\hat}
\newcommand{\ourscolor}{red}
\newcommand{\tentcolor}{Periwinkle}
\newcommand{\shotcolor}{RoyalBlue!70}
\newcommand{\urmacolor}{CadetBlue}

\DeclareMathOperator*{\argmax}{argmax} 
\DeclareMathOperator*{\argmin}{argmin} 

\newcommand{\DAsetting}[2]{{#1}$\rightarrow${#2}}

\newcommand{\parag}[1]{\smallskip\noindent\textbf{#1}\enspace}

\usepackage{stmaryrd}
\usepackage{trimclip}

\makeatletter
\DeclareRobustCommand{\shortto}{%
  \mathrel{\mathpalette\short@to\relax}%
}

\newcommand{\short@to}[2]{%
  \mkern2mu
  \clipbox{{.5\width} 0 0 0}{$\m@th#1\vphantom{+}{\shortrightarrow}$}%
  }
\makeatother

\begin{document}

\title{\titletext} 

\titlerunning{TTYD: Towards Stable and Robust
3D SFUDA}

\author{Bj\"orn Michele\inst{1,2}\orcidlink{0009-0004-1902-6232} \and
Alexandre Boulch\inst{1}\orcidlink{0000-0002-4196-9665} \and
Tuan-Hung Vu\inst{1}\orcidlink{0000-0002-9765-8233} 
\and 
Gilles Puy \inst{1}
\and 
Renaud~Marlet \inst{1,3}\orcidlink{0000-0003-1612-1758} \and
Nicolas Courty \inst{2}\orcidlink{0000-0003-1353-0126}}

\authorrunning{B.~Michele et al.}

\institute{valeo.ai, Paris, France  
\and
CNRS, IRISA, Univ.\ Bretagne Sud, Vannes, France
\and
LIGM, Ecole des Ponts, Univ Gustave Eiffel, CNRS, Marne-la-Vall\'ee, France}

\maketitle

\begin{abstract}
We tackle the challenging problem of source-free unsupervised domain adaptation (SFUDA) for 3D semantic segmentation. It 
amounts to performing domain adaptation on an unlabeled target domain without any access to source data; the 
available information is a model trained to achieve good performance on the source domain. 
A common issue with existing SFUDA approaches is that 
performance degrades after some training time, which is a by-product of an under-constrained and ill-posed problem. We discuss two strategies to alleviate this issue. First, we propose a sensible way to regularize the learning problem. Second, we introduce a novel criterion based on agreement with a reference model. It is used (1)~to stop the training when appropriate and (2)~as validator to select hyperparameters without any knowledge on the target domain. Our contributions are easy to implement and readily amenable for all SFUDA methods, ensuring stable improvements over all baselines. We validate our findings on various 3D lidar settings, achieving state-of-the-art performance.
The project repository (with code) is: \href{https://github.com/valeoai/TTYD}{github.com/valeoai/TTYD}

\end{abstract}

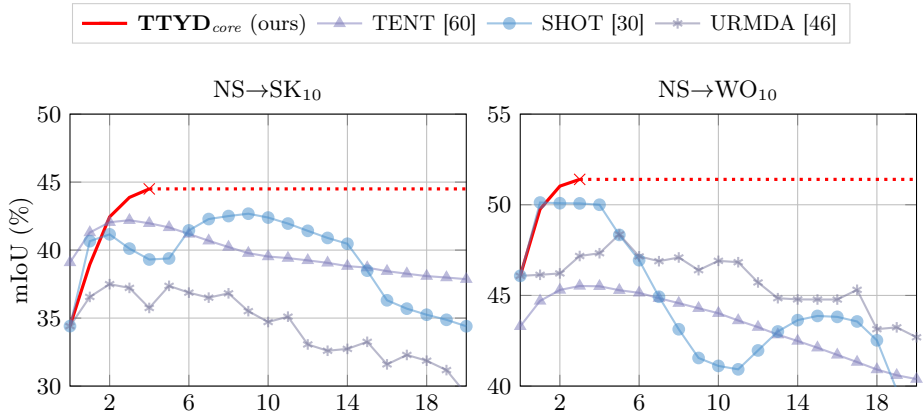
\begin{figure*}

\def\widthvar{6.85}
\def\heigthvar{5.2}

\begin{tabular}{cc}

    \multicolumn{2}{c}{
    \centering
    \begin{tikzpicture}
    \begin{customlegend}[legend columns=5,legend style={align=center,column sep=.4ex, nodes={scale=0.9, transform shape}, draw=white!80!black}, 
        legend entries={ \small {\methodcore{} (ours)}, \small TENT~\cite{wangtent}, 
        \small SHOT~\cite{shotliang20a}, \small URMDA~\cite{teja2021uncertainty}}]
        
        \addlegendimage{ very thick, \ourscolor, mark=-}
        \addlegendimage{thick, solid, mark=triangle*, color=\tentcolor, opacity=0.5, mark options ={fill=\tentcolor , fill opacity=0.5}}
        \addlegendimage{thick, solid, mark=oplus*, color=\shotcolor, opacity=0.5,  mark options ={fill=\shotcolor , fill opacity=0.5}} 
        \addlegendimage{thick, solid, mark=asterisk, color=\urmacolor, opacity=0.5,  mark options ={fill=\urmacolor , fill opacity=0.5}}
        \end{customlegend}
    \end{tikzpicture}
    }
    \\
    \begin{tikzpicture}[baseline={([yshift=\baselineskip]current bounding box.north)}]
        \tikzstyle{every node}=[font=\footnotesize]
        \begin{axis}[
            title=\nstosk,
            width=\widthvar cm,
            height=\heigthvar cm,
            font=\footnotesize,
            xtick={2,6,10,14,18},
            ylabel=mIoU ($\%$),
            ylabel shift=-5 pt,
            xmin = 0,
            xmax = 20,
            ymin = 30,
            ymax = 50,
            label style={font=\footnotesize},
            tick label style={font=\footnotesize},
            title style={yshift=-1.ex,},
            grid=major,
            legend pos=north west,
            legend style={draw=none},
            grid=major,
            legend style={at={(0.5,-0.5)},    
                anchor=north,legend columns=4},
        ]

        \addplot[ very thick, solid, color=\ourscolor]  table {figures/main_figure_data/ns2sk/ctd.dat};
         \addlegendentry{\method};
         \addplot[ very thick, dotted, color=\ourscolor]  table {figures/main_figure_data/ns2sk/ctd_2nd.dat};
         \addlegendentry{\method};
        
        \addplot[smooth,mark=x,mark size=3pt,red] 
        coordinates{
        (4,44.5)}; 
         
         \addplot [thick, mark=triangle*, color=\tentcolor, opacity=0.5, mark options ={fill=\tentcolor, fill opacity=0.5}] table {figures/main_figure_data/ns2sk/tent.dat};
         \addlegendentry{TENT};

         \addplot [thick, solid, mark=oplus*, color=\shotcolor, opacity=0.5,  mark options ={fill=\shotcolor , fill opacity=0.5}] table {figures/main_figure_data/ns2sk/shot.dat};
         \addlegendentry{SHOT};

         \addplot [thick, solid, mark=asterisk, color=\urmacolor, opacity=0.5,  mark options ={fill=\urmacolor, fill opacity=0.5}] table {figures/main_figure_data/ns2sk/ur.dat};
         \addlegendentry{URMA};

         \legend{};

         \legend{};
        \end{axis}
    \end{tikzpicture} &

\begin{tikzpicture}[baseline={([yshift=\baselineskip]current bounding box.north)}]
        \tikzstyle{every node}=[font=\footnotesize]
        \begin{axis}[
            title=\nstowy,
            width=\widthvar cm,
            height=\heigthvar cm,
            font=\footnotesize,
            xtick={2,6,10,14,18},
            ylabel shift=-5 pt,
            ymin = 40,
            ymax = 55,
            xmin = 0,
            xmax = 20,
            label style={font=\footnotesize},
            tick label style={font=\footnotesize},
            title style={yshift=-1.ex,},
            legend pos=south east,
            grid=major,
            legend pos=north west,
            legend style={draw=none},
            grid=major,
            legend style={at={(0.5,-0.5)},    
                anchor=north,legend columns=4},
        ]
         \addplot[ very thick, color=\ourscolor]  table {figures/main_figure_data/ns2wy/ctd.dat};
         \addlegendentry{\method};

         \addplot[ very thick, dotted, color=\ourscolor]  table {figures/main_figure_data/ns2wy/ctd_2nd.dat};
         \addlegendentry{\method};

         \addplot[smooth,mark=x,mark size=3pt,red] 
      coordinates{
        (3,51.4)}; 
   
         \addplot[thick, mark=triangle*, color=\tentcolor, opacity=0.5, mark options ={fill=\tentcolor , fill opacity=0.5}]  table {figures/main_figure_data/ns2wy/tent.dat};
         \addlegendentry{TENT};

         \addplot [thick, solid, mark=oplus*, color=\shotcolor, opacity=0.5,  mark options ={fill=\shotcolor , fill opacity=0.5}] table {figures/main_figure_data/ns2wy/shot.dat};
         \addlegendentry{SHOT};

         \addplot [thick, solid, mark=asterisk, color=\urmacolor, opacity=0.5,  mark options ={fill=\urmacolor , fill opacity=0.5}] table {figures/main_figure_data/ns2wy/ur.dat};
         \addlegendentry{URMA};
         
         \legend{};
        \end{axis}
    \end{tikzpicture} \\

\end{tabular}

    \caption{Evolution of the performance of baselines without degradation prevention strategies as they train over 20k iterations. Our method (\methodcore{}) uses an unsupervised criterion to stop training. 
     The horizontal dotted line illustrates that we keep the model obtained at the stopping point (marked with a cross). 
    Models are
    trained on nuScenes (NS) and \emph{unsupervisedly} adapted to SemanticKITTI (SK$_{10}$) and Waymo Open (WO$_{10}$).}
    \label{fig:main_figure}
\end{figure*}

\section{Introduction}

The goal of domain adaptation (DA) is to transfer knowledge learned from a source domain, typically with abundant or cheap annotated data, into a model suited for a target domain, typically with less data or data more expensive to annotate, thus saving acquisition or annotation costs. Concretely, DA studies learning schemes to adapt networks to different forms of shifts between source and target data distributions. If no annotation is available for the target domain, the problem is referred to as unsupervised domain adaptation (UDA).

The traditional UDA setup requires the presence of both source and target data during training. However, this is less desirable in practical scenarios for two reasons: (i)~source and target data are not always accessible at the same time due to data development cycles or to data privacy constraints, and (ii)~many models have already been trained on existing source data, and retraining on both source and target data is suboptimal in terms of consumed resources.

In this work, we address \emph{source-free unsupervised domain adaptation} (SFUDA) for 3D semantic segmentation. In this setting, target adaptation is carried out using unlabeled target data and without any access to source data; a model trained on source data is however available. As opposed to vanilla UDA, SFUDA cannot rely on source supervision to prevent the training process from drifting towards collapse~\cite{huang2021model, yi2023source}.
It is illustrated in~\cref{fig:main_figure} for baseline methods, where training first benefits to the models before being detrimental. This phenomenon is often mitigated in papers by rules of thumb, such as qualitative assessment or early stopping based on ground-truth target labels, which are however supposed to be unavailable. Though widely used in existing work, such practices obscure quantitative comparisons and raise concerns about their actual applicability. Our method departs from these practices: it totally ignores any target ground truth.

While widely exploited on image datasets, domain adaptation has recently gained attraction regarding point clouds~\cite{yi2021complete}. This task is particularly challenging because domain shifts are multiple, including specific covariate shifts due to sensors, acquisition conditions heterogeneity, and differences of class proportions between domains \cite{yi2021complete}.
Techniques like self-training and mixing~\cite{saltori2022cosmix}, object size adaptation~\cite{wang2020train} and surface regularization~\cite{michele2024saluda} have been proven effective in UDA for 3D semantic segmentation. The SFUDA setup has also been studied for 3D object detection, leveraging the temporal consistency of objects~\cite{saltori2020sf}.

For SFUDA in 3D segmentation, we resort to a straightforward yet highly effective training scheme involving two losses: one is to encourage model certainty on target samples and the other is to regularize the divergence in class distribution between source and target.
To avoid the degradation issue, we propose an unsupervised criterion that indicates when to stop the training.
For this criterion, the agreement of the trained model with a reference model is measured.
The red curve in~\cref{fig:main_figure} visualizes the evolution of our model's performance during training; the red cross marks the point when training is halted using our criterion.
Furthermore, we repurpose the stopping criterion as an unsupervised \emph{validator}, in the sense of Musgrave et al.~\cite{musgrave2022benchmarking}. We thus can unsupervisedly tune all hyperparameters used in our base SFUDA framework, making it completely hyperparameter-free.
To summarize, our contributions are the following:
\begin{itemize}
    \item We propose an unsupervised stopping criterion targeting the degradation issue of 3D SFUDA. 
    \item To achieve hyperparameter-freedom, we repurpose the stopping criterion as an unsupervised model validator.
    \item We introduce a SFUDA training scheme that works for 3D lidar data semantic segmentation and show promising results for image semantic segmentation.
    \item Extensive experiments (real-to-real and synthetic-to-real) show that our method outperforms the SOTA of 3D SFUDA.

\end{itemize}

\section{Related work}
\label{sec:related}

\subsection{SFUDA in Computer Vision}

Traditional Unsupervised Domain Adaptation techniques rely on a variety of approaches to handle  potential discrepancies between source and target domains~\cite{wilson2020survey}. While some approaches look for \emph{Domain-invariant features} by minimizing statistical divergences between source and target feature representations (\eg,~\cite{ long2015learning,long2017deep, sun2016deep, wang2017deep, damodaran2018deepjdot, fatras2021jumbot}), or through adversarial training (\eg,~\cite{ganin2016domain,tzeng2017adversarial,long2018conditional}), another line of work considers finding a \emph{Mapping between domains}~\cite{hoffman2018cycada, choi2018stargan}. Based on the assumption that both domains are not too different, other strategies were proved efficient, such as  reducing prediction uncertainty on target samples in \emph{Self-supervised methods}~\cite{vu2019advent, wangtent}, relying on~\emph{Pseudo-labeling} \cite{saito2017asymmetric,zou2018unsupervised,zhang2021prototypical,corbiere2021confidence} or \emph{Self-ensembling}~\cite{laine2016temporal, tarvainen2017mean, tranheden2021dacs, hoyer2022daformer}, that maintains a teacher model using a temporal exponential moving average of the student to ensure training stability.

In SFUDA, also called unsupervised model adaptation, and contrary to previous methods, source data is no longer available at adaptation time~\cite{liang2020we,teja2021uncertainty}. Some abstract source information is however sometimes used, \eg, adapting the target statistics of batches to those of the source~\cite{ioffe2015batch,LI2018109,nado2020evaluating,wangtent,mirza2022dua,schneider2020improving}.
The seminal work SHOT~\cite{shotliang20a} freezes the classification layer of the source model and finetunes the remaining parameters by leveraging an information maximization loss, composed of entropy minimization at the sample level to enforce unambiguous predictions, while promoting global diversity by constraining predicted class proportions~\cite{krause2010discriminativeclustering, shi2012informationtheoretical, hu2017learningdiscrete}. Without any prior knowledge, diversity turns into an objective of producing a balanced class distribution. Also, SHOT uses pseudo-labeling based on prototypes obtained by clustering classes in the target domain.

TENT~\cite{wangtent} freezes the model trained on source data but learns affine transformations in each normalization layer, whose parameters are trained to minimize classification entropy. Benefits lie in the reduced complexity of the linear adapters, which enforce simple changes in the normalization layer.
However, as highlighted in~\cref{fig:main_figure}, it is not sufficient to prevent the model from drifting towards collapse. To prevent this behavior, a possibility is to freeze the trainable weights of the source network and work only on batch norm statistics.
AdaBN~\cite{LI2018109} replaces the running statistics (mean and variance) of the source dataset by the running statistics of the target dataset.
Rather than computing the running statistics on test data once and for all, PTBN~\cite{nado2020evaluating} solely relies on the batch statistics of test data at inference time. In a similar spirit,
MixedBN in~\cite{michele2024saluda}, which is not \emph{per se} a SFUDA method, mixes at training time both source and target statistics of the combined source-target dataset, but requires the source data. We showcase in the remainder a small adaptation of it to the SFUDA case.
 
Performing adaptation at test time, those methods do not show the pathological drift exhibited in~\cref{fig:main_figure}. However, their performances compare unfavorably to methods that train a model, \eg,~\cite{shotliang20a,wangtent}. 
In this work, we propose to use these non-learned models as guardrails for the optimization process.

\parag{Semantic segmentation.} URMDA~\cite{teja2021uncertainty} is one of the first methods tackling semantic segmentation in SFUDA, by minimizing an uncertainty loss to make the feature representation more robust to noise, and by exploiting class-balanced pseudo-labeling~\cite{zou2018unsupervised}. Self-training, especially with pseudo-labels is also a popular approach~\cite{corbiere2021confidence,ye2021source,kundu2021generalize,liu2021source, huang2021model, zhao2023towards}.
In \cite{huang2021model}, the self-training stability is enforced by constraining the current model using consistency with previous models.

\parag{3D-specific SFUDA.} 
Applying UDA to 3D data has recently received a lot of attention, with a focus on detection \cite{yang2021st3d,yang2021st3d++,Luo_2021_ICCV, saltori2020sf, you2022exploiting, wang2020train, zhang2021srdan, xu2021spg, peng2022cl3d} and segmentation \cite{yi2021complete, saltori2022cosmix, michele2024saluda}. But there are only a few works on SFUDA. Some are specifically focusing on object detection  \cite{saltori2020sf, hegde2021attentive}, leveraging the trackability of cars over several frames~\cite{saltori2020sf}, or improve the identification of regions-of-interest by using attentive class prototypes~\cite{hegde2021attentive}. Others target online SFUDA for semantic segmentation~\cite{saltori2022gipso}, relying on spatio-temporal sequential lidar data, as well as on an additional point cloud processing network to produce geometric features.

\subsection{Mitigating the drift in SFUDA}

Addressing model drift during adaptation is a significant challenge in SFUDA. It is typically done by parameter tuning or early stopping based on target scores. While it offers insight into the upper-bound performance of a method, it does not account for real-world scenarios where target performance is not readily available.

\parag{Using validators.}
Validators have been introduced in UDA as methods for selecting hyperparameters without any access to target labels~\cite{musgrave2022benchmarking, ericsson2023better}. In \cite{michele2024saluda}, target entropy, information maximization (IM), and source validation have been proven to be reliable in an UDA semantic segmentation task on 3D data. SND~\cite{saito2021tune} is used in \cite{zhao2023towards} as a criterion to guide the update rate of the EMA teacher. RankME \cite{garrido2023rankme} assesses the quality of self-supervised representations without labeled downstream data, and can thus also be used to select models.

\parag{Learning stabilization.}
Another approach is to improve the training stability, \eg, modulating the learning rate or the update rate of the EMA teacher for pseudo-labeling~\cite{zhao2023towards}. In DT-ST \cite{zhao2023towards}, the update interval of the EMA teacher is selected based on the evolution of the SND~\cite{saito2021tune} or entropy values. In \cite{yi2023source}, the degradation is explained for pseudo-labeling approaches with the impact of noisy-labels, and an early-learning regularization term is introduced, putting more weight on the early predictions of the network in the training process.
\section{Method}

Our approach is mostly model-agnostic. We consider a model $\model$, with trainable parameters $\param$, that takes as input a point cloud $\pc$ and that outputs, for each point $\pt \ssp\in \pc$, a probabilistic classification prediction $\pmodel\param(\pc)_\pt \ssp\in [0,1]^\nclasses$ among $\nclasses$ classes (generally after a softmax as final layer). Without loss of generality, we consider that $\pc$ can also be a batch of point clouds, that are processed in parallel.
We assume we are in the more usual white-box SFUDA setting~\cite{fang2022source}: we know the architecture and have access to the weights. We denote by
$\pmodel{\param^\src}$
the model trained on source data~$\dataset^\src$. ($\dataset^\src$ is unavailable at domain adaptation time.)
Finally, we assume we know the source class distribution $\distrib^\src \ssp= \distrib(\dataset^\src) \ssp\in [0,1]^\nclasses$. Our goal is to find, without any ground-truth knowledge of the target data~$\dataset^\tgt$, new parameters $\param^\tgt$ such that the model $\pmodel{\param^\tgt}$ performs well on~$\dataset^\tgt$.

The framework, coined as \method{}, is composed of three elements that can be used independently: (i)~a training scheme to regularize the  
adaptation of the source-only model to target data, (ii)~a stopping criterion (\methodstop{}) to halt training and prevent performance degradation, which is additionally repurposed as a \emph{validator} (\methodvalid{}) to unsupervisedly tune training hyperparameters, and (iii)~a self-training module using the initially-adapted model (i)+(ii) (\methodcore{}) as a starting point.

\subsection{Training scheme}
Rather than training a new model from scratch, we assume that the target domain is not widely different from the source domain and adapt the already-trained model $\pmodel{\param^\src}$ by fine-tuning it on target data~$\dataset^\tgt$, without any label supervision.

\parag{General idea.} To train on unlabeled target data, we need a guidance that does not require ground-truth knowledge. To that end, we consider two training objectives. First, and quite classically, the trained (adapted) model should be discriminative, i.e., points should be classified with a large margin, which is one way to promote certainty in the predictions. Second, and more originally, the predicted class distribution of the target data should not only be diverse but in fact similar enough to the source class distribution.

As already noted, previous SFUDA work only considers perfect class balancing~\cite{shotliang20a}, while autonomous driving data contains severe class imbalance, with factors of proportion up to three orders of magnitude~\cite{liu2022less}. Besides, blindly balancing the classes ignores information that is readily available in the distribution of the source data. Additionally, favoring the alignment of the predicated class distribution onto source data is consistent with the fine-tuning strategy, which consists in finding $\param^\tgt$ in the neighborhood of~$\param^\src$. Conversely, if target data is actually very different from source data, domain adaptation makes little sense in the first place. While the first objective (discriminability) is neither particular to the task nor to the target domain, the second one (distribution similarity with source data) is specific both to the task and to the target data.

\parag{Formal description.} Concretely, to perform the training on target data, we use a loss that does not require ground-truth knowledge. This new loss is composed of two terms, which correspond to the two objectives mentioned above.

The first term penalizes ambiguity in the probabilistic class predictions. To that end, we classically~\cite{wangtent,shotliang20a} measure the entropy of predictions:
\begin{align}
    \loss_\discrim(\pc) = \frac1{|\pc|} \sum_{\pt \in \pc} \entropy(\pmodel{\param^\tgt}(\pc)_\pt)
    \label{eq:lossdiscrim}
\end{align}
where $|\pc|$ is the number of points in~$\pc$, and $\entropy$ is the entropy function.

The second term penalizes the discrepancy between the known class distribution in the source data~$\distrib^\src$, which we assume is not widely different from the (unknown) class distribution in the target data~$\distrib^\tgt$, and the predicted class distribution of~$\distrib^\tgt$, estimated as the average on the current point cloud (or batch)~$\pc$:
\begin{equation}
\loss_\simsrc(\pc) = \KL(\distrib(\pc) || \distrib^\src)\,,\quad
\text{where }\distrib(\pc) = \frac1{|\pc|} \sum_{\pt \in \pc} \pmodel{\param^\tgt}(\pc)_\pt
\end{equation}
and $\KL$ is the Kullback–Leibler divergence.
Of note, our approach differs from prior work~\cite{shotliang20a}, which tries to enforce similarity with the uniform class distribution.
In urban scene segmentation, while the source's class distribution is not perfectly aligned with the target's, it still serves as a more accurate prior than uniform. While an explicit class distribution prior has already been used in UDA \cite{hoffman2016fcns,chen2017no}, we develop it here in the specific context of SFUDA: whereas source data is inaccessible, we assume the source class distribution remains available.

Our final loss is the sum of these two terms. We do not introduce any balancing factor as the two losses somehow have a similar nature and
range of values. Indeed, like $\loss_\discrim$, $\loss_\simsrc$ can also be expressed with \mbox{(cross-)entropies}:
\begin{equation}
\loss_\simsrc(\pc) \ssp= \KL(\distrib(\pc) || \distrib^\src) \ssp= H(\distrib(\pc),\distrib^\src) \ssp-H(\distrib(\pc)) \, .
\end{equation}
Yet, to prevent overconfidence in discriminability, we consider a hinge-loss-like variant of $\loss_\discrim$ that ignores samples with very low entropy. Similarly, to prevent the adapted model from following exactly the estimated distribution of classes in the target set, we clip $\loss_\simsrc$ under a certain threshold. Our actual loss is:
\begin{equation}
    \loss(\pc) = \max(0,\loss_\discrim(\pc) -\lambda) + \max(0,\loss_\simsrc(\pc) -\lambda)\,.
\end{equation}
We use the same margin $\lambda$ for both losses, which is set to $0.02$ in all experiments.

The 3D source model is trained from scratch. Once trained, the Batch Normalization (BN) layers~\cite{ioffe2015batch} within the 3D model profoundly embody the characteristics of the source domain.
It results in significant covariate shifts when the model is applied to the target domain. Competitive results in 3D UDA are reported~\cite{michele2024saluda}
by simply altering BN statistics, with AdaBN~\cite{LI2018109}, PTBN~\cite{nado2020evaluating} or MixedBN~\cite{michele2024saluda}.
Despite its low operational cost, the effectiveness of BN adaptation in 3D perception is intriguing. Here, we explore this idea for 3D SFUDA.

We conducted an extensive study to determine  which parameters (the entire network, the classifier, or the BN layers) are better to finetune. Our finding is that most parameter schemes yield similar results. (See supp.\,mat. for details.)

As it is sufficient to only alter few parameters, we keep the model $\pmodel{\param^\src}$ completely frozen and replace BN layers by optimizable linear layers initialized with BN statistics, scale and bias. A similar affine transformation is also used, but at inference time, in \cite{wangtent}.

\subsection{Unsupervised stopping criterion and model validator}

As discussed above, training in current SFUDA methods starts to provide gain over the source-only model, before degrading (\cref{fig:main_figure}). Workarounds include adding hyper-parameters that are hard to set without peeking at the inaccessible target ground truth, \eg, fixed number of iterations or learning-rate scheduling.

A rightful solution is to rely on a \emph{validator}, which scores adapted models to choose the best one~\cite{musgrave2022benchmarking}. Using such a validator to tune hyperparameters (including the number of training iterations), is a way to make domain adaptation methods truly unsupervised~\cite{michele2024saluda,zhao2023towards}. 
A constraint is to use a validator that is not based on the same principle as the validated domain adaptation method.
As an example, using the minimization of entropy both as a validator and as an objective model optimization would lead to an infinite training.
As validators tend to measure the same kind of aspects that DA methods try to optimize, i.e., class discriminability and class diversity, this situation is not uncommon.

In fact, as illustrated in the experiment~\cref{ssec:stopping_criterion_experiment}, existing validators are not well suited for our method and fail to select a good model. The reason is that, as a gauge of discriminability, a number of validators are also based or inspired by a measure of entropy, as is our method. Regarding diversity, as existing validators are designed to be general and target-set agnostic, they tend to measure how uniform the class distribution is, which is basically the only thing one can do without any prior on the target set. But as explained above, it is not appropriate for autonomous driving data, which features highly-imbalanced classes. A specific validator-like criterion is needed with our SFUDA training.

\parag{Objective.} Our goal here is to try to capture the best performance achieved by a model as it trains, without any label knowledge on target data. More precisely, we aim to identify the point when the unknown, underlying performance of a model being trained starts to degrade. 

\parag{General idea.} In a supervised setting, a validation dataset is used to stop training when the performance on this data starts to drop, thus reducing the risk of over-fitting up to a certain extent. In UDA, the source data can be used either during the training for stabilization purposes, or as a validator~\cite{musgrave2022benchmarking} to find an optimal hyperparameter setting or an optimal point to stop the training. But in SFUDA, source data is not available; we can thus only use a model trained using source data as a basis to construct a validator or a stopping criterion.

For this construction, rather than just using model $\pmodel{\param^\src}$, we propose to use an additional auxiliary model $\damodel$ that is already adapted to the target data in an SFUDA fashion, and which is thus better than $\pmodel{\param^\src}$.
The idea however remains to explore the space of domain adaptations using our SFUDA training, starting from $\pmodel{\param^\tgt_0} \ssp= \pmodel{\param^\src}$. The auxiliary model $\damodel$ only acts as a kind of anchor to detect when the model being trained strays too much and degrades. It does not alter the training of $\pmodel{\param^\tgt}$ in any way, and definitely does not act as an upper bound in terms of performance. It merely helps to identify when training $\pmodel{\param^\tgt}$ should stop.

\parag{Formal description.} Given two models $\model, \othermodel$ that classify (among $\nclasses$ classes) the points $\data$ of a dataset $\dataset$, we measure their \emph{class assignment agreement} $\caa(\model,\othermodel)$ by counting the number of times they make identical predictions:
\begin{align}
\hspace{-2mm}\caa(\model,\othermodel) \ssp= \frac1{|\dataset|} 
\!\sum_{\data\in\dataset}\! \mathds{1}(\argmax_{k \in \interval{1,K}} \model(\data)_{k} \ssp= \argmax_{k\in \interval{1,K}} \othermodel(\data)_{k})\,.
\end{align}
As alternatives to this hard counting, we  experimented with various divergences to measure the agreement (symmetric KL divergence, L1 and L2 norms). All options gave similar results (see supp.\,mat.) and we kept the simplest one.

The measure $\caa(\model,\othermodel)$, which is also a metric, can be used to define a model validator in the sense of Musgrave et al.~\cite{musgrave2022benchmarking}, i.e., as a way to select the best model among a set of choices. Given a reference model $\othermodel$ and a set of models $\Models$, the best model $\model^*$ is the one that agrees the most with $\othermodel$:
\begin{align}
\model^* = %
\argmax_{\model \in \Models} \caa(\model,\othermodel)\,.
\label{eqn:validator}
\end{align}

We now consider a model $\model$ being trained, with parameters $\param_i$ at iteration~$i$. In its training trajectory from $\param_0$, the \emph{closest agreement point} of $\model$ with another model $\othermodel$ is the smallest iteration~$i^*$ that maximizes the agreement, i.e., $i^* = \min \argmax_{i}  \caa(\pmodel{\param_i},\othermodel)$.
Given that most on-going trainings tend to improve the performance, before the performance starts to drop continuously (cf.\ \cref{fig:main_figure}), we consider as stopping point the first  reversal in the increasing agreement phase, i.e., the first iteration $\stopt\imath$ after which the agreement starts to decrease:
\begin{align}
\stopt\imath = \argmin_{i} \caa(\pmodel{\param_i},\othermodel) \geq \caa(\pmodel{\param_{i+1}},\othermodel)\,.
\end{align}
The advantage of this \emph{first disagreement trend} is that it does not have any parameter and is quick to compute, whereas the closest agreement requires a maximum training horizon. In theory, the stopping could be sub-optimal if there are local maxima in the evolution of the class assignment agreement. However, in practice, we do not check the agreement after processing each batch but after a significant number of iterations (typically 1000), which has a smoothing effect. In our experiments, even checking the agreement as often as every 100 iterations, which is practically useless on our context, yields similar results.

Empirically, the training of model $\model$ stops when reaching the maximum agreement of $60$-$80\%$ with $\damodel$, but at a performance much higher than $\damodel$ by a large margin.
Though using an auxiliary model $\damodel$ as anchor can be seen as a limitation in that it does not favor a disruptive improvement of $\model$, we argue, as shown in our experiments, that the remaining slack of $20$-$40\%$ is sufficient to provide substantial benefits, while preventing catastrophic outcomes in SFUDA.

Note that taking $\othermodel \ssp= \pmodel{\param^\src}$ would lead to a degenerate case because the closest agreement point for $\caa(\pmodel{\param^\tgt_i},\othermodel)$ is then reached with $i^* \ssp= 0$, i.e., $\param^\tgt_0 \ssp= \param^\src$, meaning that there is no adaptation on target data from the model trained on source data. Therefore, we have to take as training starting point $\pmodel{\param^\tgt_0}$ a model close to $\pmodel{\param^\src}$, but not equal to it. We have several possible choices, among the pure SFUDA methods, as discussed below.

\parag{Selection of a reference model $\damodel$.}
In theory, the reference model $\damodel$ can be any model at hand and reliable, provided it is not based on the same principles as 
the training scheme. However, as a practical guideline, low-cost, hyperparameter-free and training-free reference models are more favorable for SFUDA.

Recent 3D UDA SOTA~\cite{michele2024saluda} reveals the intriguing effectiveness of low-cost BN adaptation methods.
We revisit these methods in the SFUDA context and observe competitive performance. Interestingly, BN adaptation methods do not require training, hence they do not suffer from the degradation issue of training-based methods.
In addition, methods like AdaBN or PTBN are hyperparameter-free, which is ideal for the unsupervised setting of SFUDA.
BN-adapted models hence become our primary choices to select a reference model $\damodel$.

In the following,  
we use PTBN as our default reference model. It gives similar results as AdaBN but can be evaluated on the fly, thus requiring less computation.
And we denote by \methodstop{} the corresponding stopping criterion.

\parag{Model validator.} 
The stopping criterion \methodstop{}, can serve as a model validator, referred to as \methodvalid{}, whose score is simply defined as the agreement level at the stopping point, i.e., $\caa(\pmodel{\param_{\stopt\imath}},\othermodel)$.
The validator helps unsupervisedly tune the hyperparmeters to obtain the best model $\model^*$, thanks to~\cref{eqn:validator}.

 \subsection{Self-training module}
The proposed training scheme along with the criterion \methodstop{} and the validator \methodvalid{} allow us to adapt the pretrained source model to a target domain using only target data; more importantly the entire process is hyperparameter-free. As later demonstrated in~\cref{sec:benchmark}, this adapted model alone, referred to as \methodcore{}, performs better or is on par with SOTA baselines.

To obtain the final \method{} model, we conduct the second phase of self-training~\cite{corbiere2021confidence,ye2021source,kundu2021generalize, liu2021source, huang2021model, zhao2023towards}.
Specifically, starting from \methodcore{}, pseudo-labels are computed on the fly for unlabeled target data, and then are used to self-train the network with the standard cross-entropy loss. 
To stabilize training, the EMA teacher model is used for pseudo-labeling~\cite{araslanov2021self}.
Additionally, we employ the Dynamic Teacher Update (DTU)~\cite{zhao2023towards} to adjust the update rate of the teacher model dynamically, further stabilizing SFUDA self-training.

\section{Experiments}

\subsection{Experimental setup}

\parag{Datasets.} The datasets we use for evaluation are listed in \cref{tab:datasets}. It is worth noting the variety of (rotating) lidar sensors (in particular number of beams), labeled classes, and world scenes. Besides, one of the six datasets is synthetic.

\parag{Class mapping.} The SFUDA setting assumes that source and target domains share semantic classes. In practice, when comparing existing datasets with ground-truth data, not all classes are shared and there are sometimes partial class overlaps. For each source-target pair, we therefore have to select and aggregate common classes in the two datasets to evaluate the quality of the domain adaptation. However, we do not train source-only models based on class mappings; we use the official classes of each dataset. The class mapping (\cref{tab:datasets} and supp.\,mat.) is only used at evaluation time, to map source-domain classes inferred on target data onto common classes that can be compared based on target ground truth.

\parag{Adapted domains.} The different domain adaptation settings we experiment with are summarized in \cref{tab:datasets}. In the following, we write as subscript the number of aggregated common classes that we use to evaluate the quality of the adaptation. We address different types of domain shifts: real-to-real and sparse-to-dense (\nstosk, \nstoposs, \nstopd, \nstowy), as well as synthetic-to-real (\synthtosk) including dense-to-sparse (\synthtoposs).

\begin{table*}[t]
\caption{Datasets used in our domain adaptation experiments.}
\label{tab:datasets}
    \centering
    \tabcolsep 1mm
\resizebox{\linewidth}{!}{
\begin{tabular}{l@{}lc|l@{~}r|c|l|l}
\toprule
Dataset  & & & Lidar & \llap{beams} & cls. & Region of the world & {Adaptation pairs} 
\\
\midrule
nuScenes & \cite{caesar2020nuscenes} &  (NS) & HDL-32E & /~32 &  16 &  Boston, Singapore 
\\
SynLiDAR & \cite{xiao2022transfer} & (SL) &  \textit{synthetic} & /~64 &  22 &  Unreal Engine~4
\\
PandaSet & \cite{xiao2021pandaset} & (PD) & Pandar64 & /~64 &  37  & 2 US cities & \nstopd~\cite{Sanchez_2023_ICCV} 
\\
Waymo Open & \cite{Ettinger_2021_ICCV} & (WO) & L.B.H. & /~64 & 23   & 3 US cities & \nstowy~\cite{kim2023single} 
\\
SemanticPOSS & \cite{pan2020semanticposs} & (SP) & Pandora & /~40 & 14  &  Peking University & \nstoposs~\cite{Sanchez_2023_ICCV}, \synthtoposs~\cite{saltori2022cosmix} \\
SemanticKITTI & \cite{behley2019iccv} & (SK) & HDL-64E & /~64 & 19   & Karlsruhe & \nstosk~\cite{yi2021complete}, \synthtosk~\cite{saltori2022cosmix}
\\
\bottomrule
\multicolumn{8}{l}{In adaptation pairs, subscript number on target indicate the number of mapped classes (cls.).}
\end{tabular}}
\end{table*}

\parag{Network setting.} For all evaluated methods, we use the same sparse-voxel Minkowski U-Net~\cite{choy20194d} with 10\,cm voxel size. It is a commonly used model for automotive lidar semantic segmentation. 
The model contains 49 batch normalization layers, thus, adapted parameters represent $0.06\%$ of the model parameters.

As in~\cite{yi2021complete, saltori2022cosmix}, we do not use lidar intensity as input feature. Lidar intensities are difficult to synthesize in simulated datasets and, for real datasets, reflectance calibration may vary a lot from one sensor to another.

To train our method, we use AdamW with a learning rate of $10^{-5}$, a weight decay of 0.01, and a batch size of~4. We use $\lambda=0.02$ in all settings and train for at most 20k iterations on target data, creating checkpoints every 1k iterations to test our stopping criterion.
The source-only models are trained to achieve high performance on the source validation set, regardless of the target data and without considering class mapping.

We show in our ablation study and the application to image modality (both see supp.~mat.) 
that a wide range of models can serve as reference. However BN adaptation models are the most readily available and remain competitive in performance.

\parag{Evaluation.} We measure performance with the classwise intersection over union (IoU) and the mean IoU (mIoU) over all classes, as done in the official \sk\ benchmark~\cite{behley2019iccv}, i.e., computed over the whole evaluation dataset.

\begin{figure}[t!]

\def\widthvar{4.7}
\def\heigthvar{4.2}
\def\xtickvar{{2,6,10,14,18}}

\def\lt{very thick}

\begin{tabular}{c@{\hskip 10pt}|cc}
    \multicolumn{1}{c|}{
    \centering
    
    \begin{tikzpicture}
        
        \begin{customlegend}[legend columns=2,legend style={column sep=.3ex, nodes={scale=0.885, transform shape, text width=5em, align=left}, draw=white!80!black}, 
        legend entries={[align=left]\nstosk, [align=left]\small\synthtosk, [align=left]\small\synthtoposs, [align=left]\small\nstoposs, [align=left]\small\nstopd, [align=left]\small\nstowy},
        legend image code/.code={\draw[mark repeat=2,mark phase=2,##1]plot coordinates {(0cm,0cm)(0.3cm,0cm)(0.35cm,0cm)};}
        ]
        \addlegendimage{\lt, solid,red}
        \addlegendimage{\lt, smooth,blue}
        \addlegendimage{\lt, smooth,green} 
        \addlegendimage{\lt, smooth,orange} 
        \addlegendimage{\lt, smooth,black}

        \addlegendimage{\lt, smooth,brown}
        
        \end{customlegend}
    \end{tikzpicture}
    }

    & 

     \multicolumn{1}{r}{
    \centering
    
    \begin{tikzpicture}
        \begin{customlegend}[legend columns=1,legend style={align=center,column sep=.4ex, nodes={scale=0.8, transform shape}, draw=white!80!black}, 
        legend entries={ [font=\normalsize]Learn. rate $\eta$ \scriptsize{(\nstosk)} ,$10^{-1,-2,-3,-4, -6,-7}$, $10^{-5}$ },
        legend image code/.code={\draw[mark repeat=2,mark phase=2,##1]plot coordinates {(0cm,0cm)(0.3cm,0cm)(0.325cm,0cm)};}
        ]
        
        \addlegendimage{empty legend}
        \addlegendimage{thick, smooth,gray}
        \addlegendimage{very thick,smooth,red}

        \end{customlegend}
    \end{tikzpicture}
    }
    & 

    \multicolumn{1}{c}{
    \centering
    \begin{tikzpicture}
        \begin{customlegend}[legend columns=1,legend style={align=center,column sep=.4ex, nodes={scale=0.8, transform shape}, draw=white!80!black}, 
        legend entries={[font=\normalsize]Margin $\lambda$ \scriptsize{(\nstosk)}, [font=\large]$0{,\,} 0.04{,\,} 0.06 {,\,} 0.08$, [font=\large]$0.02$}, 
        legend image code/.code={\draw[mark repeat=2,mark phase=2,##1]plot coordinates {(0cm,0cm)(0.3cm,0cm)(0.35cm,0cm)};}
        ]
        \addlegendimage{empty legend}
        \addlegendimage{thick, solid,gray}
        \addlegendimage{very thick, smooth,red}
        
        \end{customlegend}
    \end{tikzpicture}
    }
     
    \\
    \begin{tikzpicture}[baseline={([yshift=\baselineskip]current bounding box.north)}]
        \tikzstyle{every node}=[font=\footnotesize]
        \begin{axis}[
            width=\widthvar cm,
            height=\heigthvar cm,
            font=\footnotesize,
            xtick=\xtickvar,
            ylabel=mIoU ($\%$),
            ylabel shift=-5 pt,
            xmin = 0,
            xmax = 19,
            ymin = 20,
            ymax = 70,
            label style={font=\footnotesize},
            tick label style={font=\footnotesize},
            title style={yshift=-1.ex,},
            grid=major,
            legend pos=north west,
            grid=major,
            legend style={at={(0.5,-0.5)},    
                anchor=north,legend columns=4},
                xticklabels={,,}
        ]
        
        \addplot[\lt, smooth,red] 
        table{figures/ckpt_selection/ns2sk/miou.dat};
        \addlegendentry{\nstosk};
        \addplot[\lt,smooth,mark=x,mark size=3pt,red] 
        coordinates{(4,44.5)};
        \addplot[\lt,dashed,red] table{figures/ckpt_selection/ns2sk/miou_2nd.dat};
            
        \addplot[\lt, smooth,blue] 
        table{figures/ckpt_selection/syn2sk/miou.dat};
        \addplot[\lt, smooth,mark=x,mark size=3pt,blue] 
        coordinates{(10, 28.237324953079217)};
        \addplot[\lt, dashed,blue] 
        table{figures/ckpt_selection/syn2sk/miou_2nd.dat};

        \addplot[\lt,smooth,green] 
        table{figures/ckpt_selection/syn2poss/miou.dat};
        \addplot[\lt,smooth,mark=x,mark size=3pt,green] 
        coordinates{(3, 35.91034710407257)};
        \addplot[\lt,dashed,green] 
        table{figures/ckpt_selection/syn2poss/miou_2nd.dat};

        \addplot[\lt,smooth,orange] 
        table{figures/ckpt_selection/ns2poss/miou.dat};
        \addplot[\lt,smooth,mark=x,mark size=3pt,orange] 
        coordinates{(2, 61.07349395751953)};
        \addplot[\lt,dashed,orange] 
        table{figures/ckpt_selection/ns2poss/miou_2nd.dat};

        \addplot[\lt,smooth,black] 
        table{figures/ckpt_selection/ns2pd/miou.dat};
        \addplot[\lt,smooth,mark=x,mark size=3pt,black] 
        coordinates{(1, 63.3091926574707)};
        \addplot[\lt,dashed,black] 
        table{figures/ckpt_selection/ns2pd/miou_2nd.dat};
        
        \addplot[\lt,smooth,brown] 
        table{figures/ckpt_selection/ns2wy/miou.dat};
        \addplot[\lt,smooth,mark=x,mark size=3pt,brown] 
        coordinates{(3, 51.38615965843201)};
        \addplot[\lt,dashed,brown] 
        table{figures/ckpt_selection/ns2wy/miou_2nd.dat};

         \legend{};
    
         \legend{};
        \end{axis}
    \end{tikzpicture} 
    &
\begin{tikzpicture}[baseline={([yshift=\baselineskip]current bounding box.north)}]
        \tikzstyle{every node}=[font=\footnotesize]
        \begin{axis}[
            width=\widthvar cm,
            height=\heigthvar cm,
            font=\footnotesize,
            xtick=\xtickvar,
            ylabel shift=-5 pt,
            xmin = 0,
            xmax = 19,
            ymin = 10,
            ymax = 49,
            label style={font=\footnotesize},
            tick label style={font=\footnotesize},
            title style={yshift=-1.ex,},
            grid=major,
            legend pos=north west,
            legend style={draw=none},
            grid=major,
            legend style={at={(0.5,-0.5)},    
                anchor=north,legend columns=4},
                xticklabels={,,}
        ]
        
     \addplot[thick, color=gray]  table {figures/ablation_lr/miou/lr_10_1.dat};
    \addlegendentry{\nstosk};
    \addplot[thick, color=gray]  table {figures/ablation_lr/miou/lr_10_2.dat};
    \addlegendentry{\nstosk};
    \addplot[thick, color=gray]  table {figures/ablation_lr/miou/lr_10_3.dat};
    \addlegendentry{\nstosk};
    \addplot[thick, color=gray]  table {figures/ablation_lr/miou/lr_10_4.dat};
    \addlegendentry{\nstosk};
    \addplot[thick, color=gray]  table {figures/ablation_lr/miou/lr_10_6.dat};
    \addlegendentry{\nstosk};
    \addplot[thick, color=gray]  table {figures/ablation_lr/miou/lr_10_7.dat};
    \addlegendentry{\nstosk};

    \addplot[very thick, color=red]  table {figures/ablation_lr/miou/lr_10_5.dat};
    \addlegendentry{\nstosk};
    \addplot[very thick, color=red, dashed]  table {figures/ablation_lr/miou/lr_10_5_2nd.dat};
    \addlegendentry{\nstosk};
    \addplot[very thick,smooth,mark=x,mark size=3pt,red] 
    coordinates{(4,44.5)};

         \legend{};

         \legend{};
        \end{axis}
    \end{tikzpicture} 

    & 
    \begin{tikzpicture}[baseline={([yshift=\baselineskip]current bounding box.north)}]
        \tikzstyle{every node}=[font=\footnotesize]
        \begin{axis}[
            width=\widthvar cm,
            height=\heigthvar cm,
            font=\footnotesize,
            xtick=\xtickvar,
            ylabel shift=-5 pt,
            xmin = 0,
            xmax = 19,
            ymin = 30,
            ymax = 49,
            label style={font=\footnotesize},
            tick label style={font=\footnotesize},
            title style={yshift=-1.ex,},
            grid=major,
            legend pos=north west,
            legend style={draw=none},
            grid=major,
            legend style={at={(0.5,-0.5)},    
                anchor=north,legend columns=4},
                xticklabels={,,}
        ]
        
    \addplot[thick, color=gray]  table {figures/ablation_hinge/miou/0_00.dat};
    \addlegendentry{\nstosk};
    \addplot[thick, color=gray]  table {figures/ablation_hinge/miou/0_04.dat};
    \addlegendentry{\nstosk};
    \addplot[thick, color=gray]  table {figures/ablation_hinge/miou/0_06.dat};
    \addlegendentry{\nstosk};
    \addplot[thick, color=gray]  table {figures/ablation_hinge/miou/0_08.dat};
    \addlegendentry{\nstosk};

     \addplot[very thick, color=red]  table {figures/ablation_hinge/miou/0_02.dat};
    \addplot[very thick, color=red, dashed]  table {figures/ablation_hinge/miou/0_02_2nd.dat};
    \addlegendentry{\nstosk};
    \addplot[very thick, smooth,mark=x,mark size=3pt,red] coordinates{(4,44.5)};
    \legend{};
    \legend{};
    \end{axis}
\end{tikzpicture}

\\
   \begin{tikzpicture}[baseline={([yshift=\baselineskip]current bounding box.north)}]
        \tikzstyle{every node}=[font=\footnotesize]
        \begin{axis}[
            width=\widthvar cm,
            height=\heigthvar cm,
            font=\footnotesize,
            xtick=\xtickvar,
            ylabel=\caa ($\%$),
            ylabel shift=-2.2 pt,
            ymin = 40,
            ymax = 95,
            xmin = 0,
            xmax = 19,
            label style={font=\footnotesize},
            tick label style={font=\footnotesize},
            title style={yshift=-1.ex,},
            grid=major,
            legend pos=north west,
            grid=major,
            legend style={at={(0.5,-0.5)},    
                anchor=north,legend columns=4},
        ]
    
    \addplot[\lt, smooth,red] 
    table{figures/ckpt_selection/ns2sk/criterion.dat};
    \addplot[\lt, smooth,mark=x,mark size=3pt,red] 
    coordinates{(4,81.17736577987671)};
    \addplot[\lt, dashed,red]table{figures/ckpt_selection/ns2sk/criterion_2nd.dat};

    \addplot[\lt, smooth,blue] 
    table{figures/ckpt_selection/syn2sk/criterion.dat};
    \addplot[\lt, smooth,mark=x,mark size=3pt,blue] 
    coordinates{(10, 66.72413945198059)};
    \addplot[\lt, dashed,blue] 
    table{figures/ckpt_selection/syn2sk/criterion_2nd.dat};

    \addplot[\lt, smooth,green]
    table{figures/ckpt_selection/syn2poss/criterion.dat};
    \addplot[\lt, smooth,mark=x,mark size=3pt,green] 
    coordinates{(3, 59.404247999191284)};
    \addplot[\lt, dashed,green]
    table{figures/ckpt_selection/syn2poss/criterion_2nd.dat};
    
    \addplot[\lt, smooth,orange] 
    table{figures/ckpt_selection/ns2poss/criterion.dat};
    \addplot[\lt, smooth,mark=x,mark size=3pt,orange] 
    coordinates{(2, 88.54910731315613)};
    \addplot[\lt, dashed,orange] 
    table{figures/ckpt_selection/ns2poss/criterion_2nd.dat};

    \addplot[\lt, smooth,black] 
    table{figures/ckpt_selection/ns2pd/criterion.dat};
    \addplot[\lt, smooth,mark=x,mark size=3pt,black] 
    coordinates{(1, 92.29360818862916)};
    \addplot[\lt, dashed,black] 
    table{figures/ckpt_selection/ns2pd/criterion_2nd.dat};

    \addplot[\lt, smooth,brown] 
    table{figures/ckpt_selection/ns2wy/criterion.dat};
    \addplot[\lt, smooth,mark=x,mark size=3pt,brown] 
    coordinates{(3, 86.78926825523376)};
    \addplot[\lt, dashed,brown] table{figures/ckpt_selection/ns2wy/criterion_2nd.dat};

         \legend{};
         \legend{};
        \end{axis}
    \end{tikzpicture} 
& 
\hspace*{-0.005em}
\begin{tikzpicture}[baseline={([yshift=\baselineskip]current bounding box.north)}]
        \tikzstyle{every node}=[font=\footnotesize]
        \begin{axis}[
            width=\widthvar cm,
            height=\heigthvar cm,
            font=\footnotesize,
            xtick=\xtickvar,
            ylabel shift=-2.2 pt,
            ymin = 30,
            ymax = 90,
            xmin = 0,
            xmax = 19,
            label style={font=\footnotesize},
            tick label style={font=\footnotesize},
            title style={yshift=-1.ex,},
            legend pos=south east,
            grid=major,
            legend pos=north west,
            grid=major,
            legend style={at={(0.5,-0.5)},    
                anchor=north,legend columns=4},
        ]
        
     \addplot[thick, color=gray]  table {figures/ablation_lr/criterion/lr_10_1.dat};
    \addlegendentry{\nstosk};

    \addplot[thick, color=gray]  table {figures/ablation_lr/criterion/lr_10_2.dat};
    \addlegendentry{\nstosk};

    \addplot[thick, color=gray]  table {figures/ablation_lr/criterion/lr_10_3.dat};
    \addlegendentry{\nstosk};

    \addplot[thick, color=gray]  table {figures/ablation_lr/criterion/lr_10_4.dat};
    \addlegendentry{\nstosk};

    \addplot[thick, color=gray]  table {figures/ablation_lr/criterion/lr_10_6.dat};
    \addlegendentry{\nstosk};

    \addplot[thick, color=gray]  table {figures/ablation_lr/criterion/lr_10_7.dat};
    \addlegendentry{\nstosk};

    \addplot[very thick, color=red]  table {figures/ablation_lr/criterion/lr_10_5.dat};
    \addlegendentry{\nstosk};
    \addplot[very thick,smooth,mark=x,mark size=3pt,red] coordinates{(4, 81.23)};
     \addplot[very thick, color=red, dashed]  table {figures/ablation_lr/criterion/lr_10_5_2nd.dat};

         \legend{};
         \legend{};
        \end{axis}
    \end{tikzpicture} 
& 
\hspace*{-0.005em}

{\begin{tikzpicture}[baseline={([yshift=\baselineskip]current bounding box.north)}]
        \tikzstyle{every node}=[font=\footnotesize]
        \begin{axis}[
            width=\widthvar cm,
            height=\heigthvar cm,
            font=\footnotesize,
            xtick=\xtickvar,
            ylabel shift=-5 pt,
            ymin = 70,
            ymax = 84,
            xmin = 0,
            xmax = 19,
            label style={font=\footnotesize},
            tick label style={font=\footnotesize},
            title style={yshift=-1.ex,},
            legend pos=south east,
            grid=major,
            legend pos=north west,
            grid=major,
            legend style={at={(0.5,-0.5)},    
                anchor=north,legend columns=4},
        ]
        %
     \addplot[thick, color=gray]  table {figures/ablation_hinge/criterion/0_00.dat};
    \addlegendentry{\nstosk};
    \addplot[very thick, color=red]  table {figures/ablation_hinge/criterion/0_02.dat};
    \addlegendentry{\nstosk};
    \addplot[very thick, color=red, dashed]  table {figures/ablation_hinge/criterion/0_02_2nd.dat};
    \addplot[very thick, smooth,mark=x,mark size=3pt,red] 
      coordinates{(4, 81.23)};
    \addplot[thick, color=gray]  table {figures/ablation_hinge/criterion/0_04.dat};
    \addlegendentry{\nstosk};
    \addplot[thick, color=gray]  table {figures/ablation_hinge/criterion/0_06.dat};
    \addlegendentry{\nstosk};
    \addplot[thick, color=gray]  table {figures/ablation_hinge/criterion/0_08.dat};
    \addlegendentry{\nstosk};
         \legend{};
         \legend{};
        \end{axis}
    \end{tikzpicture}}

\end{tabular}
    \caption{Performance \%mIoU (top), as reference, and class agreement in \% (bottom), for training over 20k iterations. 
    \textbf{(1st column)} the crosses indicate when \methodstop{} stops the training in different SFUDA setups. Dashed lines after the crosses just illustrate the expected degradation issue. In reality, we do not continue training once the criterion is triggered. \textbf{(2nd and 3rd columns)} the red curves correspond to the hyperparameters $\eta$ and $\lambda$ selected using \methodvalid{} in \nstosk, showing we pick the best ones.}
    
\label{fig:checkpoint_selection_lr_selection}
\end{figure}
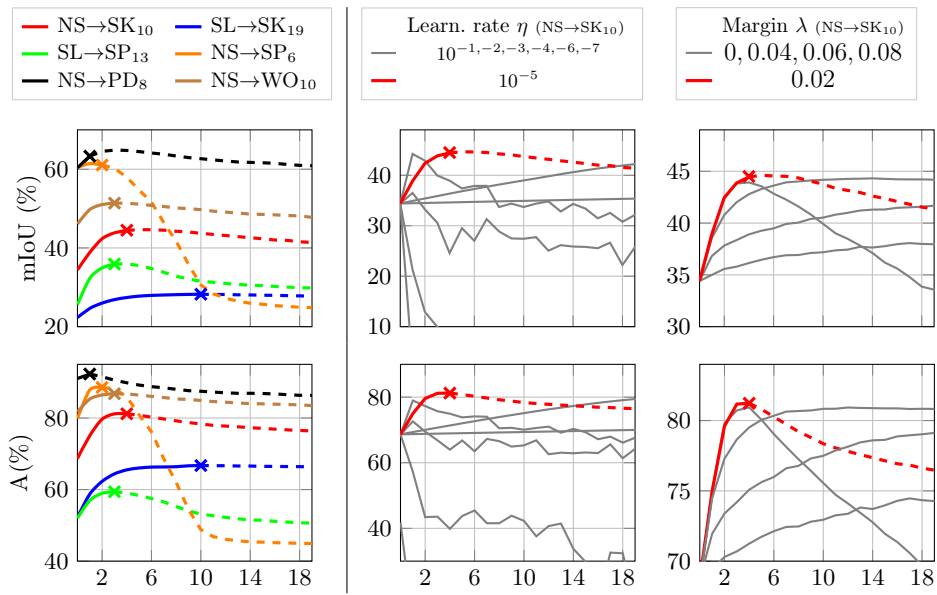

\subsection{Stopping criterion \methodstop{}}
\label{ssec:stopping_criterion_experiment}

In this section, we evaluate the quality of  our stopping criterion \methodstop{}.

First, \cref{fig:checkpoint_selection_lr_selection}(left) shows that our training scheme, while being relatively stable (little performance gap between the last and maximal mIoUs) on the majority of adaptation scenarios, can still suffer from a sharp drop of performance: -38.0 pp.\ on \nstoposs. This highlights the need for using a good stopping criterion.

Second, on the six practical domain adaptation cases we study, our stop criterion \methodstop{} is able to identify a model reaching a performance close to the best achievable one. This highlights the effectiveness of our method. In none of the observed runs was \methodstop{} misled by a local maxima of $\caa$. Computation is thus saved without giving up performance.

Third, we benchmark  \methodstop{} 
in~\cref{tab:experiments_validators}.
It outperforms other validators used as stopping criteria by a significant margin. 
RankMe, which is designed to score feature quality, always chooses a model close to the source-only model.
As expected, the `Entropy' validator selects suboptimal models as it relies on one of the ingredients that we also use for our actual domain adaptation (cf.\ \cref{eq:lossdiscrim}.

\begin{table}[t!]
    \caption{Unsupervised stopping criteria to select the best checkpoint in 20k training iterations (one checkpoint every 1k iterations).
    \textit{Oracle w/ GT} gives the upper bound.}
\label{tab:experiments_validators}
    \small
    \setlength{\tabcolsep}{2pt}
    \centering
        \begin{tabular}{l@{}r|c|c|c|c|c|c}
            \toprule
             \multicolumn{2}{l|}{Stop. Criterion} &
             \multicolumn{1}{l|}{{\ns}$\rightarrow$\skns}  & 
             \multicolumn{1}{l|}{{\synth}$\rightarrow$\sksyn}  & 
             \multicolumn{1}{l|}{{\synth}$\rightarrow$\sposssyn} & 
             \multicolumn{1}{l|}{{\ns}$\rightarrow$\spossns} & 
             \multicolumn{1}{l|}{{\ns}$\rightarrow$\wyns} & 
             \multicolumn{1}{l}{{\ns}$\rightarrow$\pdns}  \\
            \midrule
            Entropy & \cite{musgrave2022benchmarking} & 41.4 &	27.8  &	29.7 &	23.7 &  47.8 & 60.9 \\
            SND & \cite{saito2021tune} & 41.4	& 22.3 & 30.5 & 23.7  & 47.8 & 60.9  \\
            IM & \cite{musgrave2022benchmarking} & 42.4 &	27.8 & \underline{34.8} & 57.5  & \underline{51.1} & 63.0 \\
            BNM & \cite{cui2020towards} & \underline{43.9} &	27.8  & 	32.1 &	\underline{59.9}  & \underline{51.1} & 63.0\\
            RankME & \cite{garrido2023rankme} & 42.4 & \textbf{28.2}  &	32.1 &	57.5 & 51.0 & \textbf{63.3} \\
            \rowcolor{blue!10}
            \multicolumn{2}{l|} {\methodstop{}}
            & \textbf{44.5} & \textbf{28.2} & \textbf{35.9} & \textbf{61.1} & \textbf{51.4} &    \textbf{63.3} \\
            \midrule
            \rowcolor{gray!10}\multicolumn{2}{l|}{\textit{Oracle w/ GT}} & {44.7} & {28.2} & {36.0} & {61.4} & {51.4} & {64.9}  \\
            \bottomrule
        \end{tabular}
\end{table}

In conclusion, we see that our stopping criterion \methodstop{} systematically selects high-performing models. We however do not claim it is applicable beyond SFUDA, but that it is well suited for that problem.

\subsection{Model validator \methodvalid{}}

\cref{fig:checkpoint_selection_lr_selection}(right) shows performance curves on \nstosk{}, for a range of learning rates~$\eta$ and margins $\lambda$, aligned with their agreement score~$\caa$.
We observe that the agreement, which we can easily compute, is a good proxy for the actual mIoU, which cannot be known for selecting the highest one as the ground truth is not accessible. The weighted Spearman correlation (as in \cite{musgrave2022benchmarking}) between performance and agreement is 0.95 for the learning rates and 0.75 for the margins. Selecting the highest agreement thus is close to selecting the highest mIoU. 
In fact, \methodvalid{} selects $\eta \ssp= 10^{-5}$ and $\lambda \ssp= 0.02$. 

\subsection{3D-SFUDA benchmark}
\label{sec:benchmark}

\begin{table}[t!]
    \caption{
    Performance (mIoU\%) on target validation sets in two SFUDA settings: strict (without hyperparameters, or with hyperparameters tuned with a validator) and vanilla (with hyperparameters set using target ground truth). For additional comparison, we provide UDA results (using source data at adaptation time).}
\label{tab:experiments_main}
    \small
    \setlength{\tabcolsep}{1.9pt}
        \begin{tabular}{l|l|c|c|c|c|c|c|c|c}
            \toprule
              &\multicolumn{1}{r|}{Domains} & 
              Src.&H.P. &  
              \multicolumn{1}{l|}{{\ns}$\rightarrow$}  & 
              \multicolumn{1}{l|}{{\synth}$\rightarrow$} & 
              \multicolumn{1}{l|}{{\synth}$\rightarrow$} & 
              \multicolumn{1}{l|}{{\ns}$\rightarrow$} & 
              \multicolumn{1}{l|}{{\ns}$\rightarrow$} & 
              \multicolumn{1}{l}{{\ns}$\rightarrow$} \\
             & Method & 
             free & free& 
             \multicolumn{1}{r|}{~\quad\skns} & 
             \multicolumn{1}{r|}{~\quad\sksyn} & 
             \multicolumn{1}{r|}{~\quad\sposssyn} & 
             \multicolumn{1}{r|}{~\quad\spossns} & 
             \multicolumn{1}{r|}{~\quad\wyns} & 
             \multicolumn{1}{r}{~\quad\pdns} \\
            \midrule
            & Source-only  & \cmark & \cmark &  34.4 &   22.3  & 25.6 &   60.4 &   46.1 &  60.4 \\
            & AdaBN~\cite{LI2018109}          & \cmark& \cmark & 39.9 &   24.6  &  25.4 &  57.7   &47.7   & 59.6 \\ 
            & PTBN~\cite{nado2020evaluating}  & \cmark& \cmark &  39.4 &  22.4 &   23.7 &   54.7  &  42.3  & 60.2\\
            & MeanBN~\cite{michele2024saluda} & \cmark & \cmark & \underline{41.7} &   \underline{26.9} &  \underline{27.7} &  \underline{60.9} & \underline{50.3} &  \underline{61.3}  \\
            \rowcolor{blue!10}
            \cellcolor{white}\multirow{-5}{*}{\rotatebox[origin=c]{90}{\scriptsize{strict} SFUDA}}
            & \methodcore{}
            & \cmark &  \cmark & \textbf{44.5} & \textbf{28.2} & \textbf{35.9} & \textbf{61.1} & \textbf{51.4} & \textbf{63.3} \\
            
            \midrule
            
            &SHOT~\cite{shotliang20a} 
            & \cmark & \xmark & 34.9 & 18.4 & 21.7 & 42.4 & 37.3 & 43.7\\ 
            &TENT~\cite{wangtent} & \cmark & \xmark & 37.9 & 24.5 & 28.3 & 45.1 & 40.4 & 59.1\\ 
            &URMDA~\cite{teja2021uncertainty} & \cmark & \xmark & 29.4 & 25.4 & 24.5 & 30.8 & 42.7 & 56.9  \\

            &SHOT\,+\,ELR \cite{yi2023source} & \cmark & \xmark & \underline{40.5} & \underline{27.1} & \underline{36.9}  & 59.4 & 49.5 & 60.9 \\
            &DT-ST \cite{zhao2023towards} & \cmark & \oxmark & 35.6  & 23.5 & 36.8 & \underline{63.1} & \underline{51.8} & \underline{62.5}  \\ 

            \rowcolor{blue!10}
            \cellcolor{white}\multirow{-6}{*}{\rotatebox[origin=c]{90}{\scriptsize{(loose)} SFUDA}} &
            {\method{}}& \cmark & \oxmark & \textbf{45.4} & \textbf{32.4}  & \textbf{39.1} & \textbf{64.5} &  \textbf{55.5} & \textbf{65.7} \\ 
            
            \midrule

            \rowcolor{black!10}
   
            \rowcolor{black!10}&
            CoSMix~~\cite{saltori2022cosmix} & \xmark &\xmark & 38.3 & 28.0 & 40.8 & 65.2 & -- & --\\
            \rowcolor{black!10}
            \multirow{-2}{*}{\rotatebox[origin=c]{90}{UDA}} &
            SALUDA~~\cite{michele2024saluda} & \xmark &\xmark & {46.2} & 31.2 &  42.9 & 65.8 & -- & -- \\
            \bottomrule
        \end{tabular}

        \scriptsize
        H.P.\,free (no hyperparameter or selected with validator): \cmark\,{=}\,Yes; \xmark\,{=}\,No and parameter sets specific to each setting either reported in literature~\cite{saltori2022cosmix,michele2024saluda} or re-run by ourselves when default parameter do not perform correctly~\cite{shotliang20a,wangtent,teja2021uncertainty}; \oxmark\,{=}\,No but using one single set of parameters for all settings taken from image SFUDA literature~\cite{zhao2023towards,yi2023source}). Src.\,free: not using any source data at adaptation time.
    
\end{table}

\parag{Strict SFUDA setting (hyperparameter free).}
We consider here a strict SFUDA setting: any hyperparameter, if it exists, 
must be tuned without any access to target scores, thus, \eg, using to a SFUDA validator.

We compare \methodcore{} to methods that do not have any hyperparameter and that can thus be used in a pure SFUDA setting:
\emph{Source-only}, which is the model $\pmodel{\param^\src}$ trained on source data without any adaptation, \emph{AdaBN}~\cite{LI2018109}, \emph{PTBN}~\cite{nado2020evaluating} and \emph{MeanBN}, which is a simple source-free adaptation of MixedBN~\cite{michele2024saluda} (see supp.~mat. for a detailed description).

The strict SFUDA setting is presented in the upper part of\cref{tab:experiments_main}.
\methodcore{} systematically outperforms all other parameterless approaches, sometimes with a large margin (up to +8.2 pp.\ on \synthtoposs).

\parag{Loose SFUDA setting.}
In this setting, we allow the use of hyperparameters tuned by looking at the target performances. 
These hyperparameters may be specific to each adaptation pair (indicated by \xmark\ in \cref{tab:experiments_main}) or tuned once and for all (indicated by \oxmark\ in \cref{tab:experiments_main}), possibly on other modalities, \eg, images.

As the default hyperparameters of SHOT~\cite{shotliang20a}, TENT~\cite{wangtent}, and URMA~\cite{teja2021uncertainty} do not transfer to 3D SFUDA, we retrained these approaches with various sets of hyperparameters and selected the best performing ones for each adaptation pair.

Regarding SHOT\,+\,ELR \cite{yi2023source}, we used a grid-searched hyperparameter for SHOT and the two default hyperparameters for ELR, which are described as robust~\cite{yi2023source}.
As DT-ST~\cite{zhao2023towards} is designed for stability and robustness in the SFUDA setting, we used its default set of hyperparameters (experimented on images), which we also use for the DTU self-training module of \method{}.
Last, we report UDA scores (use of source data at adaptation time) for CosMix~\cite{saltori2022cosmix} and SALUDA~\cite{michele2024saluda}, as expected upper-bounds exploiting extra information.

The results obtained in the common ``vanilla'' SFUDA setting are presented in the middle part of~\cref{tab:experiments_main}. First, we observe that \method{} reaches state-of-the-art performance on all adaptation scenarios. Second, comparing the results of \methodcore{} and \method{} highlights the interest of using a self-training scheme for SFUDA.
Third, if not for \method{}, \methodcore{} ranks first or second in the vanilla benchmark, which shows that hyperparameter-less or hyperparameter-validated approaches are competitive.
Last, \method{} closes the gap between SFUDA methods and UDA approaches with an average gap of 1.2 mIoU point on four adaptation pairs.

\subsection{Application to image modality}
\label{sec:application_to_image_modality}
The formulation of \method{} appears to be general enough to be used for other modalities than 3D lidar data. To study this aspect, we conducted experiments on image segmentation and obtained promising results. Please refer to the supp. material for more details.

\subsection{Ablations}
\label{sec:app_ablations_losses}

\begin{wraptable}{r}{4cm}
    \vspace{-1.2cm}
    \caption{Loss and distribution study (\nstosk).}
\label{tab:experiments_ablation_sota_direct_im_wo_premap}
    \small
    \setlength{\tabcolsep}{2pt}
    \centering
        \begin{tabular}{c|cc|c}
            \toprule
             $\loss_\discrim$ &\multicolumn{2}{c|}{$\loss_\simsrc$} & max \\
              & {unif.} & {src} & mIoU\% \\ 
            \midrule
             & & & 34.4 \\ 
             \cmark & & &  34.4\\
              & \cmark & & 34.4  \\
              &  & \cmark & 40.9  \\
             \cmark & & \cmark & \textbf{44.7} \\ 
            \bottomrule
        \end{tabular}
    \vspace{-0.6cm}
\end{wraptable} 

\textbf{Loss terms.} Ablation of the two loss terms are presented in \cref{tab:experiments_ablation_sota_direct_im_wo_premap}, showing the relevance of each ingredient.\\
\textbf{Prior class distribution.} In \cref{tab:experiments_ablation_sota_direct_im_wo_premap}, we compare the performance obtained with a uniform prior, to the one obtained using the source class statistics. 
It clearly shows the advantage of taking into account the strong class imbalances in the data, even though they are approximated by the source statistics.

\section{Conclusion}
In this work, we propose simple and effective strategies to stabilize the performance of Source-Free Unsupervised Domain Adaptation in 3D semantic segmentation. Our contributions include a novel stopping criterion that measures an agreement with a reference model, and prevents catastrophic drifting of performance due to the under-constrained nature of the optimization problem. We also provide an easy to apply, yet efficient training scheme, that is well suited for the task of semantic segmentation in autonomous driving scenarios. We demonstrate the effectiveness of our proposal through extensive comparisons with state-of-the-art methods in 3D semantic segmentation, which is a challenging SFUDA instance, and we show its applicability in the image domain.


\section*{Acknowledgements}
We also acknowledge the support of the French Agence Nationale de la Recherche (ANR), under grants ANR-21-CE23-0032 (project MultiTrans), ANR-20-CHIA-0030 (OTTOPIA AI chair), and the European Lighthouse on Secure and Safe AI funded by the European Union under grant agreement No.~101070617. This work was performed using HPC resources from GENCI–IDRIS (2022-AD011013839, 2023-AD011013839R1).

%
%
\bibliographystyle{splncs04}
\bibliography{egbib}

\clearpage
\secondarytitle{\titletext\\
\textit{--- Supplementary Material ---}}
\normalsize
\appendix
\setcounter{table}{4}
\setcounter{figure}{2}
\makeatletter
\renewcommand\paragraph{\@startsection{paragraph}{4}{\z@}%
    {1ex \@plus0.2ex \@minus0.4ex}%
    {-0.5em}%
    {\normalfont\normalsize\bfseries}}
\makeatother

\renewcommand{\topfraction}{1}	
\renewcommand{\bottomfraction}{1}	
\setcounter{topnumber}{6}
\setcounter{bottomnumber}{6}
\setcounter{totalnumber}{8}     
\setcounter{dbltopnumber}{6}    
\renewcommand{\dbltopfraction}{1}	
\renewcommand{\textfraction}{0.00}	
\renewcommand{\floatpagefraction}{0.7}	
\renewcommand{\dblfloatpagefraction}{0.7}	

\subsection*{Overview}

In this document, we provide: experiments on the application of \method{} to the image modality (Sec.~\ref{sec:app:application_to_image_modality}), additional implementation details (Sec.~\ref{sec:app:implementation_details}), a guarantee of the soundness (Sec.~\ref{sec:app:soundness_guarantee}), and additional ablations: on the parameters to adapt (Sec.~\ref{sec:app:model_parameters_to_adapt}), on alternative distances for the consistency validator \methodstop\ (Sec.~\ref{sec:app:other_distances}) and on other reference models (Sec.~\ref{sec:app:other_reference_models}). We also report the performance of \methodstop\  with other training schemes (Sec.~\ref{sec:app:other_training_schemes}) and discuss the SFUDA hypothesis for our training scheme (Sec.~\ref{sec:app:sfuda_hypothesis}).
Additionally, we also provide the per-class results and comparison to non-SF UDA approaches (Sec.~\ref{sec:per_class_results}), qualitative results (Sec.~\ref{sec:qualitative_results}), and more details on the datasets and class mappings (Sec.~\ref{sec:datasets_overview}).

\section{Application to image modality}
\label{sec:app:application_to_image_modality}

While developed for 3D SFUDA, the formulation of \method{} appears to be general enough to be used for other modalities.
To study this aspect, we conducted experiments on image segmentation. We used the GTA5 dataset
\cite{richter2016playing} as source, and the Cityscapes (City) dataset \cite{cordts2016cityscapes} as target.

\begin{wraptable}{r}{5.5cm}
    \vspace{-1.cm}
    \caption{SFUDA for image modality.}
\label{tab:experiments_image}
    \small
    \setlength{\tabcolsep}{2pt}
    \centering
        \begin{tabular}{l|l|c}
            \toprule
              & Valid. &GTA5 $\rightarrow$ \\
             Method & ref. model & City \\
            \midrule
            Source-only   & & 36.8\\
            URMDA \cite{teja2021uncertainty} & &45.1 \\
            SFDA \cite{liu2021source} & &45.8 \\
            SDF \cite{ye2021source}& &49.4 \\
            HCL\cite{huang2021model} & & 48.1 \\
            DT-ST \cite{zhao2023towards} & &52.1\\
            \rowcolor{blue!10}
            \method{} & PTBN  & \textbf{53.4} \\  
            \rowcolor{blue!10}
            \method{} & DT-ST & \underline{53.2} \\
            \bottomrule
        \end{tabular}
        \vspace{-0.7cm}
\end{wraptable} 

This is also an opportunity to evaluate if different models can be used as reference models for the validation.
We remark, nevertheless, that it is common practice for image semantic segmentation to keep the ImageNet-pretrained batchnorm frozen during training on the source dataset. We cannot directly use a PTBN version of such source-only models as reference for \methodstop{}, in particular because the ImageNet-pretrained batchnorm statistics differ too much from those we would have obtained on the source training set. Therefore, we use a PTBN model built using a source-only model trained \emph{without} freezing the BN layers \cite{Chen_2019_ICCV}. We also test the DT-ST model from \cite{zhao2023towards}.

Our results are presented in \cref{tab:experiments_image}.
We also reach SOTA performances for the GT5$\rightarrow$City adaptation pair. As we use the self-training module of DT-ST, we can conclude that, as for 3D SFUDA, the final performance relies on the quality of the self-training starting point, which is provided here by \methodcore{}.

\section{Additional implementation details}
\label{sec:app:implementation_details}

We use PyTorch for our implementation \cite{paszke2019pytorch}. The models for \nstosk, \synthtosk{}, and \nstopd{} are trained on a single NVIDIA GeForce RTX 2080 Ti (11 GB) GPU. For \synthtoposs, \nstoposs{}, and \nstopd{},  we use a split NVIDIA A100-40GB GPU with 20 GB memory.

\paragraph{Code.}
AdaBN~\cite{LI2018109} and PTBN~\cite{nado2020evaluating} were not designed specifically for 3D point clouds; we implemented them. MeanBN is derived from the idea of MixedBN~\cite{michele2024saluda} (rather than the code of MixedBN, which requires source data, see just below); we implemented it ourselves. AdaBN, PTBN and MeanBN are hyperparameter-free. For DT-ST~\cite{zhao2023towards}, we used the official code repository and default parameters, as recommended. Code for \relax{SHOT}~\cite{shotliang20a}, \relax{TENT}~\cite{wangtent} and \relax{URMDA}~\cite{teja2021uncertainty} was taken from their official repository, with parameters set as described below.

\paragraph{Note on MixedBN and MeanBN.}
\label{sec:app:note_on_mixedbn}
In the main paper, we introduce MeanBN as a SFUDA version of MixedBN\cite{michele2024saluda}. Indeed, MixedBN computes the average running statistics of the source and target datasets by mixing them during the training, which cannot be done in an SFUDA setting. MeanBN just averages (with equal weight) the running statistics from source training and from passing the target data through the source-trained network: it is the average of the running statistics of Source-only and AdaBN.

\paragraph{Parameters selected for SHOT, TENT and URMDA.}

For \relax{SHOT}~\cite{shotliang20a}, we obtained the best results on the target validation set with a learning rate of $10^{-6}$ and a balancing hyperparameter of $\beta\ssp=10^{-5}$. For \relax{TENT}~\cite{wangtent} and \relax{URMDA}~\cite{teja2021uncertainty}, we used a learning rate of~$10^{-5}$. Additionally, as URMDA relies on \cite{zou2018unsupervised}
for setting per-class confidence thresholds, we achieved optimal results with significantly different values for the target portion~$p$, depending on source and target domains: $p\ssp=0.01$ (\nstosk, \nstowy), $p\ssp=0.9$ (\synthtosk, \synthtoposs, \nstopd), $p\ssp=0.1$ (\nstoposs). 

\paragraph{Self-training (ST)}
propagates and somehow denoises uncertain pseudo-labels. It has been successfully used in UDA \cite{saltori2022cosmix,michele2024saluda} and SFUDA \cite{zhao2023towards}. Table~3 in the main paper shows the benefits of adding self-training in our context (line \methodcore{} vs line \method{}).

We used the self-training from \cite{zhao2023towards}, which we adapted for point clouds, e.g., regarding augmentations. This self-training handles a confidence level for each class, making sure to also promote rare classes. This allows us to train on the target data, selecting a mostly-correct set of labels while keeping a sufficient balance of rare classes, also preventing collapse, which may occur when focusing mainly on most frequent classes.

\paragraph{Training time.}

Our stopping criterion \methodstop{} saves a lot of time and computation at the training stage. For example, we stop the training for \nstosk{} after 1.1~hr, compared to 6~hrs for a full 20k-iteration training. 

The design of our training scheme itself makes it also faster, as there is no costly centroid generation after each epoch like in SHOT~\cite{shotliang20a}, where 20k iterations require 30~hrs, or time-consuming surface reconstruction regularization like in SALUDA~\cite{michele2024saluda}, which is reported to run in 120 hrs. The self-training step then takes about 10~hrs.

\paragraph{GPU memory footprint.}

Our training scheme is also memory efficient at training time, as only one semantic segmentation network is needed. This is in contrast, \eg, to DT-ST~\cite{zhao2023towards}, where an additional teacher network is used, or to SALUDA~\cite{michele2024saluda}, which uses an additional geometric regularization head during training.

\section{Soundness guarantee}
\label{sec:app:soundness_guarantee}
We can show that \methodstop{} is \emph{sound} because the agreement $\caa(\model,\othermodel)$ (cf.\ Eq.\,(5)), which is bounded by~$1$, can only take at most $|\dataset|\,{+}\,1$ different values. Hence, the number of iterations, as defined by Eq.\,(7), is bounded by $|\dataset|$. Also, to check the stopping criterion efficiently, we actually only evaluate Eq.\,(7) after a fixed number $N$ of iterations (typically, $N\,{=}\,1000$). Even so, the number of iterations remains bounded, by $N|\dataset|$. In our experiments, the number of iterations at the stopping point is however much smaller than $N|\dataset|$, typically between 5 and 10k.

However, it is to be noted that we have no \emph{performance} guarantees, as most UDA and SFUDA methods, including validators \cite{musgrave2022benchmarking, saito2021tune}, whose performance results are generally empirical.

\section{Ablation: Model parameters to adapt}
\label{sec:app:model_parameters_to_adapt}

In \cref{tab:app:experiments_ablation}, we explore a wide range of possible options concerning the parameters to adapt, some of which are already proposed in the literature \cite{LI2018109, michele2024saluda, shotliang20a, zhao2023towards, wangtent}. Please note that reported values represent the maximum performance over a training for 20k iterations; a stopping criterion is to be used on top of that.

Although they differ in terms of maximum performance, most adaptation strategies make sense, except adapting the classification layer only (\cref{tab:app:experiments_ablation}.a). On the contrary, adapting the features in the backbone, including before each layer, is key to the performance, to obtain linearly separable features. Adapting the running statistics online both at train and eval time also is detrimental (\cref{tab:app:experiments_ablation}.b), probably because it does not ``see'' enough target data. In the end, we adopt for our method the affine transformations before each batch normalization layer as it performs the best, although adapting the backbone is on average nearly as good. Besides, it reduces the memory footprint as fewer parameters have to be updated (although not reducing gradient computation) and it could facilitate investigations for a deeper understanding of the adaptation.
\newpage

\begin{table*}[t!]
    \small
    \setlength{\tabcolsep}{4pt}
    \centering
     \caption{Ablation study }
\label{tab:app:experiments_ablation}

     \textbf{(a) Parameters to adapt}. Assuming frozen statistics, parameters to update can be 
     replacement of BN by linear layer,
     or the backbone weights only (without the classification layer) for different learning rates, or the classification layer only, or the complete network (backbone + classification layer). 

        \begin{tabular}{l||c|c||c|c|c||c||c|c|c}
            \toprule
               & \multicolumn{2}{c||}{BN$\rightarrow$ Lin.}  &  \multicolumn{3}{c||}{Backbone only} & \multicolumn{1}{c||}{Classif.} & \multicolumn{3}{c}{Backbone+classif.} \\
             Adaptation & w/o  & w/ & $10^{-5}$ & $10^{-6}$ & $10^{-7}$ & layer & $10^{-5}$  & $10^{-6}$ & $10^{-7}$ \\
              & bias  & bias &  &  &  &  &   &  &  \\
            \midrule
            \nstosk & 44.0\hphantom{$^\dagger$} & \textbf{44.7}  & 40.3 & 42.1\hphantom{$^\dagger$} & 42.0$^\dagger$ & 34.4 &   41.5 & 41.4$^\dagger$ & 35.7$^\dagger$  
            \\
            \synthtosk & 27.9$^\dagger$ & 28.2    & 27.9 & \textbf{28.5}\hphantom{$^\dagger$} & 26.7$^\dagger$ & 22.4 &   28.1 & 28.0$^\dagger$ & 23.3$^\dagger$ 
            \\
            \synthtoposs & 36.1\hphantom{$^\dagger$} & 36.0   & 31.3 & 36.6\hphantom{$^\dagger$} &  \textbf{36.9}$^\dagger$ & 34.1 &  36.6 & 36.9\hphantom{$^\dagger$} & 30.0$^\dagger$  
            \\
            \nstoposs & {61.5}\hphantom{$^\dagger$} & {61.4} & 60.5  & \textbf{61.5}$^\dagger$ & 60.9$^\dagger$ & 60.4 &  \textbf{61.5} & 61.4\hphantom{$^\dagger$} & 61.0$^\dagger$  
            \\
            \bottomrule
        \end{tabular}

    ~\\~\\

    \setlength{\tabcolsep}{2pt}

    \begin{minipage}{0.60\linewidth}
      \textbf{(b) Choice of running statistics for BN layers}, either fixed 
      or variable (per-instance norm. at train and eval time, or only at train and fixed at eval). 
    \end{minipage}~~
    \begin{minipage}{0.35\linewidth}
     \textbf{(c) Class distribution to target}, uniform or obtained from source data. 
    \end{minipage}
   
    \begin{minipage}{0.60\linewidth}

        \begin{tabular}{l||c|c|c|c|c}
            \toprule
              & \multicolumn{3}{c|}{Fixed statistics}  &  \multicolumn{2}{c}{Online statistics} \\
             Adaptation & source  & target &  mean & train  & train  \\
              &   &  &   & +eval  &  \\
            \midrule
            \nstosk & {44.7} & 43.4 & \textbf{45.9} & 39.1 & 43.7$^*$ \\
            \synthtosk & \textbf{28.2} & 26.2 & 27.4 & 22.2 & 26.7$^*$ \\
            \synthtoposs & \textbf{36.0} & 30.9 & 34.4 & 23.7 & 26.8$^*$ \\
            \nstoposs & \textbf{61.4} & 59.3 & 61.1 & 54.7 & 60.4$^*$ \\
            \bottomrule
        \end{tabular}
    \end{minipage}~~
    \begin{minipage}{0.35\linewidth}
        \begin{tabular}{l||c|c}
            \toprule
              &  \multicolumn{2}{c}{Distribution}  \\
            Adaptation  & uniform & source \\
            \midrule
            \nstosk & 35.0 & 44.7  
            \\
            \synthtosk &  23.8 & 28.2
            \\
            \synthtoposs &  25.6 & 36.0  
            \\
            \nstoposs &  60.7 & 61.4
            \\
            \bottomrule
        \end{tabular}
    \end{minipage}

~\\
\raggedright
Maximum mIoU\% over 20k iterations, learning rate $10^{-5}$ unless otherwise stated. \\
$^*$: performance strongly fluctuating. \quad\quad $^\dagger$: maximum reached at 20k iterations.
\end{table*}

\section{Ablation: Other distances for consistency validator}
\label{sec:app:other_distances}
\begin{table}[t!]
    \small
    \setlength{\tabcolsep}{2pt}
    \centering
     \caption{Performance of our criterion \methodstop\ and other using soft measurements to select a model being trained over 20k iterations (one model for each 1k iteration increment).}
\label{tab:app:experiments_validators_measure}
        \begin{tabular}{l@{}r|c|c|c|c|c|c}
            \toprule
             \multicolumn{2}{r|}{Adaptation} & {\ns}$\rightarrow$   \\
             Validator && {\skns} \\
            \midrule
            \multicolumn{2}{l|}{\methodstop\ (i.e., hard choice $\caa$)} & \textbf{44.5} \\
            \midrule
            \multicolumn{2}{l|}{\methodstop\ L2} & \textbf{44.5} \\
            \multicolumn{2}{l|}{\methodstop\ L1} & \textbf{44.5} \\
            \multicolumn{2}{l|}{\methodstop\ Symmetric KL} & \textbf{44.5} \\
            
            \bottomrule
        \end{tabular}
   
\end{table}
We show in \cref{tab:app:experiments_validators_measure} the results of our stopping criterion using various divergences to measure the agreement (symmetric KL divergence, L1 and L2 norms), instead of the default hard counting of identical predictions. As all different options give the same results we keep the simplest one, the hard counting of identical predictions.

\section{Ablation: Other reference models}
\label{sec:app:other_reference_models}
\begin{table}[t!]
    \caption{Performance of our \methodstop{} with different reference models to select a model being trained over 20k iterations (one model for each 1k iteration increment).}
\label{tab:app:experiments_other_reference}
    \small
    \setlength{\tabcolsep}{2pt}
    \centering
        \begin{tabular}{l@{}r|c|c|c|c|c|c}
            \toprule
             \multicolumn{2}{r|}{Adaptation} & 
             \multicolumn{1}{l|}{{\ns}$\rightarrow$}  & 
             \multicolumn{1}{l|}{{\synth}$\rightarrow$}  & 
             \multicolumn{1}{l|}{{\synth}$\rightarrow$} & 
             \multicolumn{1}{l|}{{\ns}$\rightarrow$} & 
             \multicolumn{1}{l|}{{\ns}$\rightarrow$} & 
             \multicolumn{1}{l}{{\ns}$\rightarrow$}  \\
             Validator && \multicolumn{1}{l|}{~~\skns} & 
             \multicolumn{1}{r|}{~~\sksyn} & 
             \multicolumn{1}{r|}{~~\sposssyn} & 
             \multicolumn{1}{r|}{~~\spossns} & 
             \multicolumn{1}{r|}{~~\wyns} & 
             \multicolumn{1}{r}{~~\pdns} \\
            \midrule
            Source-only  &  &  34.4 &   22.3  & 25.6 &   60.4 &   46.1 &  60.4 \\
            \rowcolor{black!20}
            {TTYD-train \small{(last iter.)}} &  & 39.2 & 27.8 & 28.1 & 23.4 & 47.7 &  60.8 \\
             \rowcolor{black!20}
             {TTYD-train \small{(max.\ value)~}} & & {44.7} & {28.2} & {36.0} & {61.4} & {51.4} & {64.9} \\ 
            \midrule
            \multicolumn{2}{l|}{\methodstop\ (i.e., w/ PTBN)} & {44.5} & \textbf{28.2} & {35.9} & {61.1} & \textbf{51.4} &    {63.3} \\
            \multicolumn{2}{l|}{\methodstop\ w/ AdaBn} &  {44.5} & \textbf{28.2} &  \textbf{36.0} & {61.1}  &  \textbf{51.4} & {63.3}  \\
            \multicolumn{2}{l|}{\methodstop\ w/ MeanBN} &   39.0 &  26.9 &  32.3 &  61.1 & 49.8 & 60.4 \\
            \midrule
            \methodstop\ w/ SHOT & \cite{shotliang20a} & 43.8 & 22.3& 29.8 & 60.4 & 46.1 & 63.3  \\
             \methodstop\ w/ TENT & \cite{wangtent} &  43.0 & 27.4& 35.9 & \textbf{61.4} & 50.2 & \textbf{64.5} \\
              \methodstop\ w/ URMDA & \cite{teja2021uncertainty} & 39.0  &24.7 & 25.6 & 60.4 & 46.1 & 60.4 \\
            \methodstop\ w/ SHOT\,+\,ELR &\cite{yi2023source}&  \textbf{44.6} & 28.1 & 32.3 & 60.4 & 51.0 & {63.3} \\
            \methodstop\ w/ DT-ST &  \cite{zhao2023towards} &  42.4 & 26.9 & 32.3 & 60.4 & 49.8 & {63.3}  \\

            \bottomrule
        \end{tabular}
\end{table}

In \cref{tab:app:experiments_other_reference}, we compare the performance of the model selected by \methodstop\ using PTBN as a reference model, against the selection of models using AdaBN and MeanBN as reference models. 
It can be seen that using PTBN or AdaBN as reference model are mostly equivalent. Using MeanBN is clearly inferior, probably because it is too close to the source-only model: it always selects a model trained for less iterations than our proposed alternatives.

We also tested other models as potential reference models: DT-ST, SHOT+ELR, SHOT, TENT and URMDA. We use the model obtained after 20k iterations as reference model for all these methods. DT-ST and and SHOT-ELR are able to select competitive checkpoints, improving performance over the source-only one in 5 out of the 6 domain adaptation scenarios. Although SHOT suffered from a strong performance degradation during training, and therefore would not be a natural choice as reference model, SHOT allows selection a better performing model that the source-only model in half of the domain adaptation settings, and never select a model performing worse than the source-only one. It is to be noted that PTBN, AdaBN, MeanBN are hyperparameter-free. We use default hyperparameters for DT-ST. For SHOT, TENT, URMDA, we use target-validated hyperparameters to study their potential.

\section{\methodstop\ for other training schemes}
\label{sec:app:other_training_schemes}
\begin{table}[t!]
    \caption{Performance of our criterion \methodstop\ to select a SHOT or URMDA model being trained over 20k iterations (one model for each 1k iteration increment).}
\label{tab:app:experiments_ttyd_shot}
    \small
    \setlength{\tabcolsep}{2pt}
    \centering
        \begin{tabular}{l@{}r|c|c|c|c|c|c}
            \toprule
             \multicolumn{2}{r|}{Adaptation} & {\ns}$\rightarrow$  & {\synth}$\rightarrow$  & {\synth}$\rightarrow$ & {\ns}$\rightarrow$ & {\ns}$\rightarrow$ & {\ns}$\rightarrow$  \\
             Validator && {\skns} & {\sksyn} & {\sposssyn} & {\spossns} & {\wyns} & {\pdns} \\
            \midrule
             
            Source-only  &  &  34.4 &   22.3  & 25.6 &   60.4 &   46.1 &  60.4 \\
            \midrule
            \rowcolor{black!20}
            {TTYD-train \small{(last iter.)}} &  & 39.2 & 27.8 & 28.1 & 23.4 & 47.7 &  60.8 \\
             \rowcolor{black!20}
             {TTYD-train \small{(max.\ value)~}} & & {44.7} & {28.2} & {36.0} & {61.4} & {51.4} & {64.9} \\ 
             \rowcolor{blue!20}
             {\methodcore}& 
            & \textbf{44.5} & \textbf{28.2} & \textbf{35.9} & \textbf{61.1} & \textbf{51.4} & \textbf{63.3} \\
\midrule
            \rowcolor{black!20}
            \multicolumn{2}{l|}{SHOT last iter.} & 34.9 & 18.4 & 21.7 & 42.4 & 37.3 & 43.7 \\
            \rowcolor{black!20}
            \multicolumn{2}{l|}{SHOT max.} & 42.7 & 27.9 & 36.7 & 61.2 & 50.1  & 62.9 \\
            \multicolumn{2}{l|}{SHOT w/ \methodstop} & 40.7 & 27.9 & 35.9 & 61.2 & 50.1 & 62.9 \\
            
            \midrule
            \rowcolor{black!20}
            \multicolumn{2}{l|}{URMDA last iter.} & 29.4 & 25.4 & 24.5 & 30.8 & 42.7 & 56.9\\
            \rowcolor{black!20}
            \multicolumn{2}{l|}{URMDA max.} & 37.5 &25.5 &  33.4 & 63.0 & 48.4 & 60.4   \\
            \multicolumn{2}{l|}{URMDA w/ \methodstop} & 37.2 & 25.6  &  25.6 & 60.4   & 46.1  & 60.4 \\
            
            \bottomrule
        \end{tabular}
\end{table}
In \cref{tab:app:experiments_ttyd_shot} we also apply \methodstop\ to SHOT and URMDA, as both methods are facing strong model degradation during training. We report the maximal achieved performance during training (max.), the performance reached after 20k iterations (last iter.), and the performance reached using our stopping criterion (\methodstop). We see that our stopping criterion is able to pick a model whose performance is close to the best achieved performance during training (max.).

The application of our stop criterion on TENT does not make sense as the starting point for the TENT method is identical to the reference model.

\section{SFUDA hypothesis}
\label{sec:app:sfuda_hypothesis}

For our training scheme, we use no source data. Besides a source-only trained model $\pmodel{\param^\src}$, we only use global statistics $\distrib^\src \,{=}\, \distrib(\dataset^\src)$ on source data, i.e., a few class frequencies. 
These class-wise point ratios are in fact often already provided on dataset datasheets, \eg, SemanticKITTI\cite{behley2019iccv}, nuScenes\cite{caesar2020nuscenes}.
This very minor requirement complies with motivations of source-free approaches, \eg, privacy, lost access or computation saving. As it can be seen in \cref{tab:app:alternative_prior}:  alternatives to our prior ($D^S$) in Eq.\,(2) (main paper) do not perform well on \nstosk. However, the correct target class data distribution ($D^T$), which of course is not available, but could be seen of a kind of oracle, helps to further improve the performance.

\begin{table}[]
\caption{Comparison of different priors in Eq.\,(2) on \nstosk. For easier comparison we report the maximal obtained performance with our training scheme without the selection of \methodstop.}
    \label{tab:app:alternative_prior}
    \centering
    \tabcolsep3pt
    \begin{tabular}{l|c|c|c|l}
    \toprule
    $\KL(\distrib(\pc) || ~?~)$ & unif. & $\distrib(\pmodel{\param^\src}(\dataset^\tgt))$ & $\distrib^\src $(ours) & $\distrib^\tgt$ (oracle) \\
    \midrule
    Ours (mIoU\%) & 35.0 &34.4 & 44.7 & 47.0 \\
    \bottomrule
\end{tabular}

\end{table}

\begin{center}

\end{center}
\vspace*{-3mm}

\section{Classwise results and related approaches}
\label{sec:per_class_results}

In this section, we detail classwise results of semantic segmentation after domain adaptation.  We also compare to UDA methods.

\paragraph{Per-class results.}

We provide in~\crefrange{tab:experiments_per_class_ns_sp}{tab:experiments_per_class_ns_pd} the classwise results for methods and domain adaptation settings reported in Tab.~2 of the main paper. It can be seen that the gain in performance (mIoU) achieved by our \methodcore\ originates, on all dataset settings, from a consistent improvement over a broad range of classes, not just a few of them.

\paragraph{UDA (with source data) as a kind of SFUDA upper bound.}

General UDA is privileged over the SFUDA setting because it has access to the source data at training time. UDA resutls thus represents a kind of upper bound to SFUDA's. To analyze this aspect, we compare to two state-of-the-art UDA methods, namely CoSMix \cite{saltori2022cosmix} and SALUDA \cite{michele2024saluda}, on the domain adaptation settings we experimented with and for which UDA results are available, i.e., \nstosk, \synthtosk, \synthtoposs\ and \nstoposs. 

Please note that CoSMix has hyperparameters, which have to be (and are) optimized for each setting on the ground-truth target validation set (which somewhat detracts from the lack of supervision). On the contrary, SALUDA uses an unsupervised validator (Entropy~\cite{musgrave2022benchmarking}), like we do with our own unsupervised stopping criterion and validator.

As can be seen in \crefrange{tab:experiments_per_class_ns_sp}{tab:appendix_experiments_per_class_ns_sk}, although CoSMiX and SALUDA do have a better mIoU on average, our method \methodcore\ still outperforms CoSMix on 2/4 domain adaptations and is only 1.8 to 4.7 percentage points behind SALUDA, except on \synthtoposs, where SALUDA remains 7.0 p.p.\ ahead. 
\method\  reduces the gaps with SALUDA down to 0.8 to 3.8 p.p., and even outperforms SALUDA by 1.2 p.p.\ on \synthtosk.

Please note that we compare to values reported in the SALUDA paper~\cite{michele2024saluda}, including for CoSMix \cite{saltori2022cosmix}, as the evaluation protocol in \cite{saltori2022cosmix} for mIoU calculation differs from the official evaluation metric \cite{behley2019iccv}, which we use instead. Furthermore, \cite{michele2024saluda} report results as an average over 3 runs, whereas we provide here only the results of a single run.

\begin{table}[b]
\small
\centering
\caption{\textbf{Classwise results for \nstoposs{}.} $^\dagger$ from \cite{michele2024saluda}.}
\label{tab:experiments_per_class_ns_sp}
\newcommand*\rotext{\multicolumn{1}{R{45}{1em}}}
\setlength{\tabcolsep}{2.2pt}
\begin{tabular}{lr|c|cccccc}
\toprule
 \rlap{\raisebox{8mm}{\nstoposs}}%
 \rlap{\raisebox{0mm}{\hspace{4.5mm}(\%\,IoU)}}%
 && \rotext{\%\,mIoU} & \rotext{Person} &	\rotext{Bike} & \rotext{Car} & \rotext{Ground} &	\rotext{Vegetation} &	\rotext{Manmade}\\
 
\midrule
\rowcolor{green!10}
\multicolumn{9}{l}{\emph{{Strict SFUDA}}}\\

Source-only && 60.4&  56.1&	7.5&	\textbf{65.0}&	\textbf{79.4}&	79.0&	\textbf{75.7} \\

AdaBN~\cite{LI2018109} & & 57.7 & \textbf{58.8} & \textbf{14.9} &	42.8 &	76.8 &	79.2 &	73.7  \\ 
PTBN~\cite{nado2020evaluating}& & 54.7 & 	55.2 &	10.5 &	 41.0 &	75.7 &	74.8 &	70.9 \\
MeanBN~\cite{michele2024saluda} & & 60.9 & 	58.6 &	12.4 &	60.7 &	78.0 &	80.0 &	75.5 \\
\rowcolor{blue!10}
{\methodcore~(ours)} && \textbf{61.1}& 57.0&	11.3&	{64.2}&	79.0&	\textbf{80.6} &	74.4  \\ 
\midrule
\rowcolor{orange!10}
\multicolumn{9}{l}{\emph{Loose SFUDA}}\\
SHOT~\cite{shotliang20a}&& 42.4 &	19.0&	0.0&	13.3&	78.7&	71.6&	72.1\\
TENT~\cite{wangtent}&&45.1	&36.0&	0.1&	35.9&	76.1&	62.0&	60.5\\
URMDA~\cite{teja2021uncertainty}&& 30.8&	36.2&	7.7&	2.6&	71.1&	26.2&	41.1\\
SHOT+ELR~\cite{yi2023source} && 59.4	& 54.0&	1.2&	67.0&	79.9&	78.3&	75.9\\
DT-ST~\cite{zhao2023towards} & & 63.1 & 	59.8 &	7.6 &	72.9 &	\textbf{81.0} &	79.2 &	78.2 \\
 \rowcolor{blue!10}
\method~(ours)& & \textbf{64.5} &	\textbf{61.0} &	\textbf{10.4} &	\textbf{74.5} &	80.9 &	\textbf{81.6} &	\textbf{78.8} \\ 
\midrule
\rowcolor{red!10}
\multicolumn{9}{l}{\emph{UDA methods with src data and (for CoSMix) parameters}} \\
\rowcolor{red!10}
 CoSMix$^\dagger$~\cite{saltori2022cosmix}&& 65.2  
 & {60.3} &	{24.1}	&	66.4	&	80.4&	{81.4}	&	78.3  \\
\rowcolor{red!10}
 SALUDA$^\dagger$~\cite{michele2024saluda} && {65.8}
 & 59.0		& 20.5	&	{70.6} & {82.6}	&	{81.4}	 &	{81.0}	  \\
 \bottomrule

\end{tabular}

\end{table}
\begin{table*}[t!]
\small
\centering
\caption{\textbf{Classwise results for \DAsetting{\ns}{\skns}.} $^\dagger$ from \cite{michele2024saluda}.}
\label{tab:appendix_experiments_per_class_ns_sk}
\newcommand*\rotext{\multicolumn{1}{R{45}{1em}}}
\setlength{\tabcolsep}{3.5pt}

\resizebox{\textwidth}{!}{
\begin{tabular}{lr|c|cccccccccc}
\toprule
 \rlap{\raisebox{11mm}{\DAsetting{\ns}{\skns}}}%
 \rlap{\raisebox{0mm}{\hspace{4.5mm}(\%\,IoU)}}%
 &&  \rotext{\%\,mIoU}  & \rotext{Car} & \rotext{Bicycle}& \rotext{Motorcycle} &\rotext{Truck} &	\rotext{Other vehicle}	 & \rotext{Pedestrian}	 & \rotext{Driveable surf.} &	\rotext{Sidewalk} &	\rotext{Terrain} &\rotext{Vegetation} \\

\midrule

\midrule
\rowcolor{green!10}
\multicolumn{13}{l}{\emph{{Strict SFUDA}}}\\
Source-only  &&  34.4 & 77.5 & 	8.8	 & 18.3	 &5.7 &	4.6 &	\textbf{52.0} &	38.8 &	25.6 &	29.7	 & 83.2 \\ 
AdaBN~~\cite{LI2018109} && 39.9 & 80.8 & 14.5 &	16.7 &	8.6 & 	3.8 & 	23.8 & 	\textbf{75.0} & 	\textbf{38.9} & 	\textbf{52.9}  & 84.0   \\ 
PTBN~~\cite{nado2020evaluating} && 39.4 & 	80.0 &	14.7 &	27.0 &	7.3 &	5.5 &	23.2	 &71.3 &	35.4 &	48.8& 	80.6 \\

MeanBN~~\cite{michele2024saluda} && 41.7 & 	87.0 & 	 \textbf{17.6} &	29.6 & 	12.1 &	4.4 &	43.8 & 	61.3 &	33.3 &	40.2 & 	\textbf{87.5}  \\

\rowcolor{blue!10}

\methodcore~(ours) & & \textbf{44.5} & \textbf{87.4} & 7.8 &	\textbf{30.1} &	\textbf{16.6} &	\textbf{8.3} &	{50.1} &	71.9 &	33.2 &	51.9 &	87.3 \\

\midrule 
\rowcolor{orange!10}
\multicolumn{13}{l}{\emph{Loose SFUDA}}\\
SHOT~\cite{shotliang20a} && 34.9 & 	90.2 &	1.2 &	8.6 &	20.9 &	6.2 &	1.2 &	68.9 &	19.0 &	\textbf{60.4} &	72.3 \\
TENT~\cite{wangtent} && 37.9 & 	58.4 &	0.1 &	4.6 &	\textbf{43.1} &	10.2 &	41.6 &	66.1 &	20.3 &	57.8 &	76.4\\
URMDA~\cite{teja2021uncertainty} && 29.4 & 	72.0 &	1.4	 &3.4 &	3.3 &	3.1 &	18.3 &	36.4 &	\textbf{36.8} &	 41.4 &	78.0\\
SHOT+ELR~\cite{yi2023source} && 40.5 & 	90.1 &	\textbf{2.8} &	18.2 &	16.2 &	\textbf{10.6} &	44.9 &	69.3 &	15.8 &	51.2 &	86.1\\
DT-ST~~\cite{zhao2023towards} && 35.6 &	88.6 &	0.0	 & 26.3 &	9.1 & 	{4.1} &	\textbf{54.9} & 	39.9 &	17.2 &	29.2 &	87.2 \\
\rowcolor{blue!10}
 \method~(ours) &  & \textbf{45.4} & 	\textbf{92.4} &	0.0	& \textbf{37.0}	 & {26.9}	 & 2.1 &	49.0 &	\textbf{72.8} &	{27.7} &	{56.3} &	\textbf{89.7}  \\
 
\midrule 
\rowcolor{red!10}
\multicolumn{13}{l}{\emph{UDA methods with source data and (for CoSMix) hyperparameters}} \\
\rowcolor{red!10}
CoSMix$^\dagger$~~\cite{saltori2022cosmix} & & 38.3
&77.1	&	10.4& 20.0	&{15.2}	&	6.6	&	{51.0}&	52.1&	31.8&	34.5&84.8 \\  
\rowcolor{red!10}
SALUDA$^\dagger$~~\cite{michele2024saluda}&&  {46.2}
&{89.8} &	13.2&	26.2&	{15.3}	&	7.0	&	37.6&	{79.0}&	{50.4}&	{55.0}&{88.3} \\

\bottomrule
\end{tabular}
}

\end{table*}
\begin{table*}[!ht]

\caption{\textbf{Classwise results for \synthtosk.} $^\dagger$ from \cite{michele2024saluda}.}
\label{tab:experiments_per_class_syn_sk}

\newcommand*\rotext{\multicolumn{1}{R{45}{1em}}}
\setlength{\tabcolsep}{2.2pt}
\resizebox{\textwidth}{!}{
\begin{tabular}{lr|c|ccccccccccccccccccc}
\toprule
 \rlap{\raisebox{10mm}{\synthtosk}}%
 \rlap{\raisebox{0mm}{\hspace{4.5mm}(\%\,IoU)}}%
 && \rotext{\%\,mIoU}&  \rotext{Car} &	\rotext{Bicycle} & \rotext{Motorcycle} & \rotext{Truck} &	\rotext{Other vehicle} &	\rotext{Pedestrian} &	\rotext{Bicyclist} &	\rotext{Motorcyclist} & \rotext{Road} & \rotext{Parking} & \rotext{Sidewalk} &	\rotext{Other ground} &	\rotext{Building} &	\rotext{Fence} & \rotext{Vegetation} &	\rotext{Trunk} & \rotext{Terrain} &	\rotext{Pole} &\rotext{Traffic sign} \\ \midrule

\rowcolor{green!10}
\multicolumn{22}{l}{\emph{{Strict SFUDA}}}\\
Source-only &&22.3 & 40.7 &	7.6 &	9.6 &	1.5 &	1.7	&21.0&	\textbf{47.1} &	1.6 &	21.9 &	4.7 &	34.0&	0.0	&36.3&	22.2 &	62.3 &	28.3 &	\textbf{48.5} &	28.8&	5.6\\
AdaBN~\cite{LI2018109} &&24.6 & \textbf{64.2} & 	8.5 &	9.1 &	2.9 &	3.3 &	20.8 &	27.0 &	0.4 &	56.5 &	\textbf{6.8} &	30.5 &	0.0 &	64.9 &	17.8 &	59.2 &	19.2 &	36.6	&28.0 &	11.5    \\ 
PTBN~\cite{nado2020evaluating} && 22.4 & 	53.5 &	6.5	 & \textbf{11.2} &	\textbf{4.7} &	\textbf{3.5} &	18.8 &	30.4 &	0.3 &	52.4 &	3.9 & 	33.2 & 	0.0 &	58.5 &	14.4 &	45.3 &	20.2 &	32.7 &	25.7 &	10.4  \\
MeanBN~\cite{michele2024saluda} && 26.9 &	59.6 & 	9.1 &	9.8 & 	2.4 & 	3.1 & 	23.6  & 	{37.3} &	1.2 &	42.5 &	\textbf{6.8} &	\textbf{34.0} &	0.1 &	60.2 &	\textbf{28.8} &	68.9 &	29.3 & 	42.3 &	38.0 & 	14.5 \\
\rowcolor{blue!10}
{\methodcore~(ours)} && \textbf{28.2}& 63.9	& \textbf{11.1} &	11.0 &	3.6 &	3.0	&\textbf{26.5}&	33.0 &	\textbf{1.7} &	\textbf{63.2} &	5.9 &	32.3 &	\textbf{0.2} &	\textbf{67.4} &	19.1 &	\textbf{72.6} &	\textbf{30.5}&	 35.4 &	\textbf{40.9} &	\textbf{15.2}  \\
\midrule

\rowcolor{orange!10}
\multicolumn{22}{l}{\emph{Loose SFUDA}}\\
SHOT~\cite{shotliang20a} && 18.4 &	49.5 &	1.0 &	2.1	& 4.5 &	4.2 &	13.7 & 	8.0 &	{0.5} &	60.0 &	4.2	 &24.0 &	\textbf{0.5}	& 46.5 &	16.7 &	38.0 &	 22.8 &	15.1 &	37.4 &	0.9\\
TENT~\cite{wangtent} && 24.5 &	57.8 & 	3.3	& 9.5 &	\textbf{12.4} &	2.5 &	11.7 &	20.3 &	0.0 &	52.0 &	0.3 &	34.2 &	0.0 &	60.8 &	15.6 &	66.9 &	29.9&	44.4 &	40.6 &	3.5\\
URMDA~\cite{teja2021uncertainty} && 25.4 & 	52.0 &	3.3 &	6.3	& 1.3	& 1.1 &	14.7 &	52.0 & 	1.2 &	26.2 &	\textbf{5.6} &	\textbf{37.0} &	0.1 &	46.3 &	\textbf{32.3} &	65.3 &	35.8 &	51.6 &	45.8 &	4.7\\
SHOT+ELR~\cite{yi2023source} && 27.1 & 	56.7	 & 4.1	 &10.0 &	3.3 &	1.7 &	31.4 &	32.7 &	 1.0 &	62.1 &	2.8	& 33.7 &	0.1 &	64.9 &	7.6 &	71.9 &	32.3 &	40.0 &	46.2 &	12.2\\
DT-ST~\cite{zhao2023towards} && 23.5	 & 34.9	& 2.1 &	10.9 &	2.3 & 	2.0 & 	29.2 &	\textbf{66.7} &	1.0 &	20.6 &	3.2 &	{35.1} &	0.0 &	27.8 &	5.4 &	60.4 &	30.7 &	 \textbf{52.9} &	48.8 &	12.6  \\
\rowcolor{blue!10}
\method~(ours) &  &  \textbf{32.4} & 	\textbf{77.0} &	\textbf{5.0} &	\textbf{12.8} &	{8.7}	 & \textbf{2.9} & 	\textbf{40.0} &	43.6 &	\textbf{1.2} &	\textbf{67.4} &	{5.5} &	34.8 &	0.0 &	\textbf{70.8} &	{8.4} &	\textbf{77.5} &	\textbf{40.4} &	38.6 &	\textbf{52.8} &	\textbf{28.1} \\

 \midrule 

\rowcolor{red!10}
\multicolumn{22}{l}{\emph{UDA methods with source data and (for CoSMix) hyperparameters}} \\
\rowcolor{red!10}

 CoSMix$^\dagger$~\cite{saltori2022cosmix}&& {\perf{28.0}} 
& {63.9} &	5.6 &	11.4 &	{5.7} &	{7.9}	 & 20.0 &	40.3 &	{3.8}	& {56.4} &	{13.2} &	{37.9} &	0.1 &	42.6 &	{29.5} &	66.9 &	27.9 &	29.6 &	{46.0} &	{22.5}  \\

\rowcolor{red!10}
 SALUDA$^\dagger$~\cite{michele2024saluda} && {31.2} 
 & {65.4}& 7.5 &13.6&3.2 &5.9&{23.9}&43.7&	1.7&52.9&{11.6}&{39.8}&{0.3}&{67.8}&28.2&{74.2}&{37.6}&43.6&{47.5}&{22.7} \\
 \bottomrule

\end{tabular}
}

\end{table*}
\begin{table*}[!ht]
\small
\centering

\caption{\textbf{Classwise results for \synthtoposs{}.} $^\dagger$ from \cite{michele2024saluda} and uses a voxel size of 5\,cm.}
\label{tab:experiments_per_class_syn_sp}
\newcommand*\rotext{\multicolumn{1}{R{45}{1em}}}
\setlength{\tabcolsep}{2.2pt}

\resizebox{\textwidth}{!}{
\begin{tabular}{lr|c|ccccccccccccc}
\toprule
 \rlap{\raisebox{9mm}{\synthtoposs}}%
 \rlap{\raisebox{0mm}{\hspace{4.5mm}(\%\,IoU)}}%
 & & \rotext{\%\,mIoU} & \rotext{Person} & \rotext{Rider} &	\rotext{Car}	& \rotext{Trunk}	& \rotext{Plants} &	\rotext{Traffic sign}	& \rotext{Pole} &	\rotext{Garbage can}& \rotext{Building}	& \rotext{Cone}	& \rotext{Fence} & \rotext{Bike}	& \rotext{Ground}  \\ \midrule

\rowcolor{green!10}
\multicolumn{16}{l}{\emph{{Strict SFUDA}}}\\
Source-only  && 25.6 & 43.2 & 	31.4 &	22.5 &	20.8 &	65.8 &	1.0 &	4.5 &	14.9 &	53.9 &	7.0 &	21.5 &	3.0	 & 43.4 \\ 
AdaBN~\cite{LI2018109}&& 25.4&  38.4 & 	17.8 & 	22.4 & 	23.6 & 	55.9 & 	\textbf{13.0} &	7.8	& 8.8 &	61.1 &	6.9 &	14.9 & 	\textbf{9.3} & 	50.9  \\ 
PTBN~\cite{nado2020evaluating} && 23.7 &	36.3	&20.4 &	27.0 &	19.9 &	43.4	 &10.6 &	6.8 &	8.2 &	58.8 &	5.2 &	15.3 &	8.5 &	47.7\\
MeanBN~\cite{michele2024saluda} && 27.7 & 	38.9 &	23.2 &	22.5 &	26.2 &	69.5 &	6.1 &	7.0 &	15.6 &	63.2 &	9.4 &	21.2 &	5.2 &	52.2\\
\rowcolor{blue!10}
{\methodcore~(ours)} && \textbf{35.9} &  \textbf{46.1} & \textbf{37.2} & 	\textbf{43.5} &	\textbf{31.3} &	\textbf{71.3} &	4.8 &	\textbf{20.5} &	\textbf{21.8} & 	\textbf{69.1} & 	\textbf{11.5} &	\textbf{25.4} &	4.3 & 	\textbf{79.9}   \\
\midrule 
\rowcolor{orange!10}
\multicolumn{16}{l}{\emph{Loose SFUDA}}\\
SHOT~\cite{shotliang20a} && 21.7 &	31.1 &	5.7&	11.8	& 32.9&	37.1&	8.0&	18.5&	4.6&	52.3&	6.2&	18.1&	0.1&	55.3 \\
TENT~\cite{wangtent} &&28.3 &	39.1&	30.0	&33.4	&20.0	&63.3&	0.0&	21.4&	3.0&	60.0&	16.8&	31.6&	\textbf{0.7}&	48.7 \\
URMDA~\cite{teja2021uncertainty}&& 24.5 &	42.0&	37.7&	\textbf{50.3}&	23.5&	46.1&	0.0&	21.5&	0.0&	41.9&	0.0&	\textbf{51.7}&	0.0&	3.4\\
SHOT+ELR~\cite{yi2023source} && 36.9 &	59.8&	29.1&	47.7&	\textbf{30.4}	&71.1&	1.3&	23.1&	12.1&	70.9&	\textbf{18.4}&	34.4&	0.4&	\textbf{81.9} \\
DT-ST~\cite{zhao2023towards} && 36.8 & 	\textbf{64.1} &	\textbf{57.1} &	47.3 &	21.5 &	65.3	 & 3.6 &	23.6 &	\textbf{28.3} &	58.5 &	{6.2} &	35.1 &	0.3 & 67.1 \\
\rowcolor{blue!10}
\method~(ours)&&  \textbf{39.1} & 	\textbf{64.1} &	54.8 &	{48.9} &	{27.8} &	\textbf{73.0} &	\textbf{8.8}	 &\textbf{29.4} &	14.1 &	\textbf{73.6} &	5.9 &	{36.8} &	{0.5} &	{70.7}\\
\midrule 
\rowcolor{red!10}
\multicolumn{16}{l}{\emph{UDA methods with source data and (for CoSMix) hyperparameters}} \\
\rowcolor{red!10}
 CoSMix$^\dagger$~\cite{saltori2022cosmix} && 40.8 & 50.9 & 54.5 &	34.9 &	33.6 &	71.1 & 19.4 &	35.6 &	26.8 &	65.2 &	30.4 &	24.0	&6.0 & 	78.5 \\
\rowcolor{red!10}
 SALUDA$^\dagger$~\cite{michele2024saluda} &  & 42.9 & 59.9 & {54.6}	&	59.2	&	{33.7}	&	69.8	&14.9&{40.9}&{30.8}&64.5&26.2&22.1&2.7&78.0	 \\
 \bottomrule

\end{tabular}
}

\end{table*}
\begin{table*}[!ht]
\small
\caption{\textbf{Classwise results for \nstowy{}.}} 
\label{tab:experiments_per_class_ns_wy}
\centering
\newcommand*\rotext{\multicolumn{1}{R{45}{1em}}}
\setlength{\tabcolsep}{2.2pt}
\begin{tabular}{lr|c|ccccccccccccc}
\toprule
 \rlap{\raisebox{11mm}{\nstowy}}%
 \rlap{\raisebox{0mm}{\hspace{4.5mm}(\%\,IoU)}}%
 && \rotext{\%\,mIoU} &  \rotext{Car} & \rotext{Bicycle}& \rotext{Motorcycle} & \rotext{Truck}& \rotext{Other vehicle} & \rotext{Pedestrian}& \rotext{Driveable surf.}& \rotext{Sidewalk}& \rotext{Walkable}& \rotext{Vegetation}\\ 
 \midrule

\midrule
\rowcolor{green!10}
\multicolumn{13}{l}{\emph{{Strict SFUDA}}}\\
Source-only && 46.1& 72.2 & 	6.2	 & 14.0 &	24.9 &	24.5 &	68.1 &	70.8 &	47.8 &	43.8 &	88.6  \\
AdaBN~\cite{LI2018109} && 47.7 & 70.5 & 	8.9	 & 9.1 &	27.6 &	 33.2 &	58.8 &	\textbf{82.2} &	51.5 &	46.4 & 	89.0  \\ 
PTBN~\cite{nado2020evaluating} && 42.3 & 	65.1 &	4.5 & 	7.7 & 	21.7 &	22.1 &	51.8 &	{80.3} &	46.4 &	40.4 &	83.3 \\
MeanBN~\cite{michele2024saluda} &&50.3 & 	75.2 &	\textbf{9.6} &	12.8 &	\textbf{30.0} &	\textbf{37.2} &	67.5 &	78.5 & 	52.2 &	\textbf{48.9} &	\textbf{91.5} \\
\rowcolor{blue!10}
{\methodcore~(ours)} && \textbf{51.4}&  \textbf{77.5} & 	7.6 &	\textbf{17.3} &	27.5 &	36.1 &	\textbf{74.2} &	{80.3} &	\textbf{53.8} &	48.4 &	91.1  \\ 
\midrule
\rowcolor{orange!10}
\multicolumn{13}{l}{\emph{Loose SFUDA}}\\
SHOT~\cite{shotliang20a}  && 37.3 &	56.2 &	0.8 &	7.6	& 15.2&	21.7&	36.9&	61.7&	45.9&	41.1&	85.7\\
TENT~\cite{wangtent} && 40.4	& 56.5 &	0.4 &	10.9 &	18.3 &	23.8 &	52.1 &	82.2 &	47.8 &	35.5 &	76.2\\
URMDA~\cite{teja2021uncertainty} && 42.7&	71.9&	1.7&	1.3&	26.2&	20.6&	60.2&	64.9&	52.1&	41.5&	86.5\\
SHOT+ELR~\cite{yi2023source} && 49.5 &	79.5&	2.2&	\textbf{24.0}&	26.2&	29.0&	67.6&	76.5&	51.9&	50.0 &	88.1\\
DT-ST~\cite{zhao2023towards} && 51.8 & 	81.0 &	6.8	 & 18.9 &	\textbf{33.1} &	42.9 &	77.6 &	72.1 &	47.5 &	45.7 &	92.7  \\
\rowcolor{blue!10}
\method~(ours) && \textbf{55.5} & 	\textbf{83.1} &	\textbf{8.4} &	{20.4} &\textbf{33.1} &	\textbf{46.0} &	\textbf{79.5} &	\textbf{82.2} &	\textbf{55.4} &	\textbf{53.0} &	\textbf{93.5} \\

 \midrule

\end{tabular}

\end{table*}
\begin{table}[!t]
\small
\centering
\caption{\textbf{Classwise results for \nstopd{}.}}
\label{tab:experiments_per_class_ns_pd}
\newcommand*\rotext{\multicolumn{1}{R{45}{1em}}}
\setlength{\tabcolsep}{2.2pt}
\vspace*{1mm}
\begin{tabular}{lr|c|ccccccccccccc|}
\toprule
 \rlap{\raisebox{13mm}{\nstopd}}%
 \rlap{\raisebox{0mm}{\hspace{4.5mm}(\%\,IoU)}}%
 && \rotext{\%\,mIoU} & \rotext{2-wheeled} & \rotext{Pedestrian}& \rotext{Driveable ground} &  \rotext{Sidewalk} & \rotext{Other ground} &  \rotext{Manmade} & \rotext{Vegetation} &  \rotext{4-wheeled} \\ 

\midrule
\rowcolor{green!10}
\multicolumn{11}{l}{\emph{{Strict SFUDA}}}\\
Source-only && 60.4 & 	27.6 &	64.2 &	71.6 &	45.1 &	24.2 &	88.1 &	75.0 &	87.2 \\
AdaBN~\cite{LI2018109} && 59.6 & 	31.3 &	51.6 &	77.3 &	44.5 &	28.5 &	86.0 &	73.1 &	84.3 \\ 
PTBN~\cite{nado2020evaluating} && 60.2 & 	\textbf{32.4} &	52.3 &	76.1 &	46.0 &	28.3 &	86.9 &	74.1 &	85.6 \\
MeanBN~\cite{michele2024saluda} && 61.3 & 	31.3 &	61.6 &	75.0 &	44.8 &	27.0 &	87.8 &	75.0 &	87.5 \\
\rowcolor{blue!10}
{\methodcore~(ours)} && \textbf{63.3} & 28.8 & 	\textbf{65.3} & 	\textbf{78.1} & 	\textbf{49.0} &	\textbf{30.5} &	\textbf{88.2} &	\textbf{76.2} &	\textbf{90.4}  \\ 
\midrule
\rowcolor{orange!10}
\multicolumn{11}{l}{\emph{Loose SFUDA}}\\
SHOT~\cite{shotliang20a} && 43.7&	0.7&	38.4&	27.7	&40.1&	17.1&	84.5&	67.8&	72.5\\
TENT~\cite{wangtent} && 59.1 &	14.8&	50.5&	\textbf{83.6}&	\textbf{50.8}&	25.8&	85.5&	72.7&	89.2 \\
URMDA~\cite{teja2021uncertainty} && 56.9&	17.0&	62.2	&68.9&	40.1&	22.6&	88.5&	71.9&	84.9\\
SHOT+ELR~\cite{yi2023source} && 60.9	& 15.2&	58.5&	78.1&	48.3&	30.0&	88.8&	77.4&	90.8\\
DT-ST~\cite{zhao2023towards} && 62.5 & 	32.7 &	\textbf{64.2} &	75.9 &	43.8 &	26.6 &	\textbf{89.1} &	{77.5} &	90.4 \\ 
\rowcolor{blue!10}
{\method}~(ours) && \textbf{65.7} & \textbf{35.2} &	\textbf{64.2} &	{81.7} &	{49.5} &	\textbf{35.9} &	88.4 &	\textbf{78.3} &	\textbf{92.9} \\
 \bottomrule

\end{tabular}

\end{table}

\clearpage

\section{Qualitative results}
\label{sec:qualitative_results}

\paragraph{Methods with no degradation prevention.}

\begin{figure*}
\def\widthimage{0.2}
\newcommand{\rotext}[1]{{\begin{turn}{90}{#1}\end{turn}}}
\setlength{\tabcolsep}{1pt}
\centering
\begin{tabular}{c@{~~~}cccc}

\rotatebox{90}{\enspace\enspace TENT }
&
\includegraphics[trim=40 0 20 10,clip,width=\widthimage\linewidth]{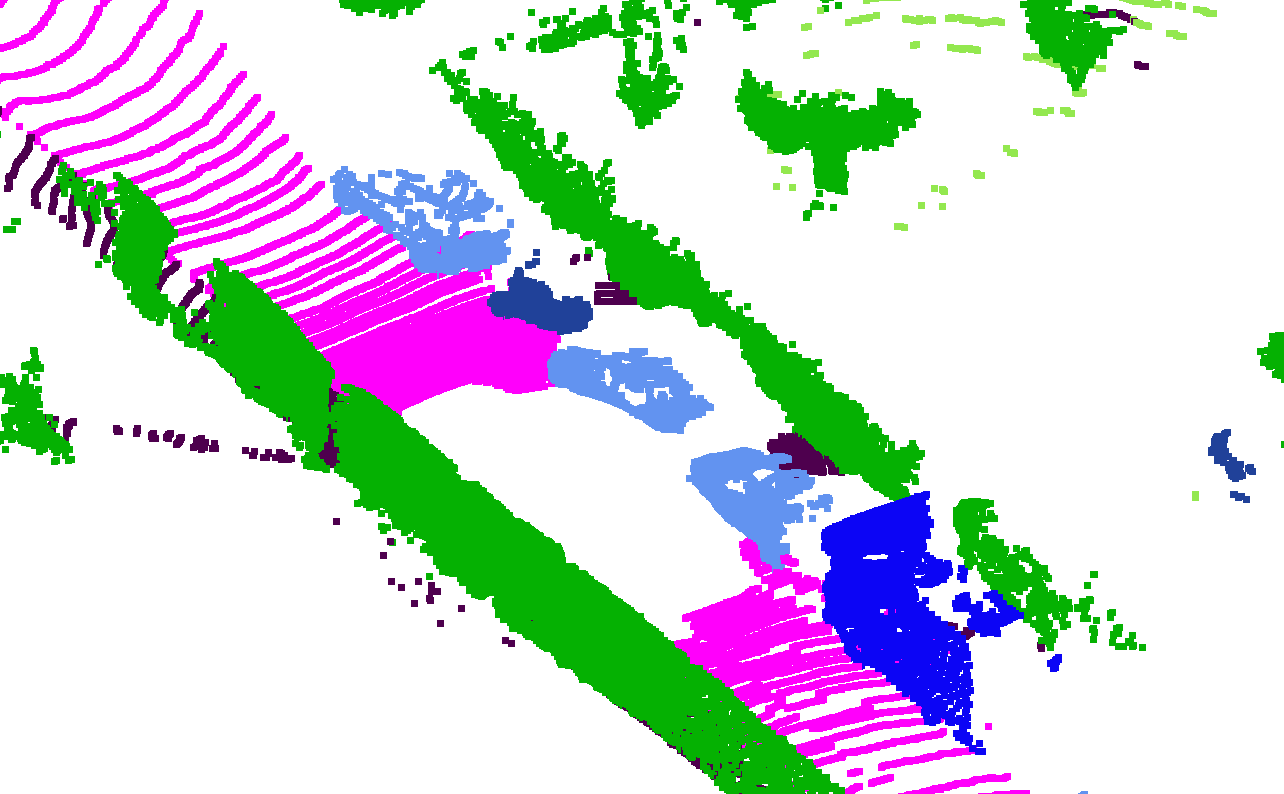}&
\includegraphics[trim=40 0 20 10,clip,width=\widthimage\linewidth]{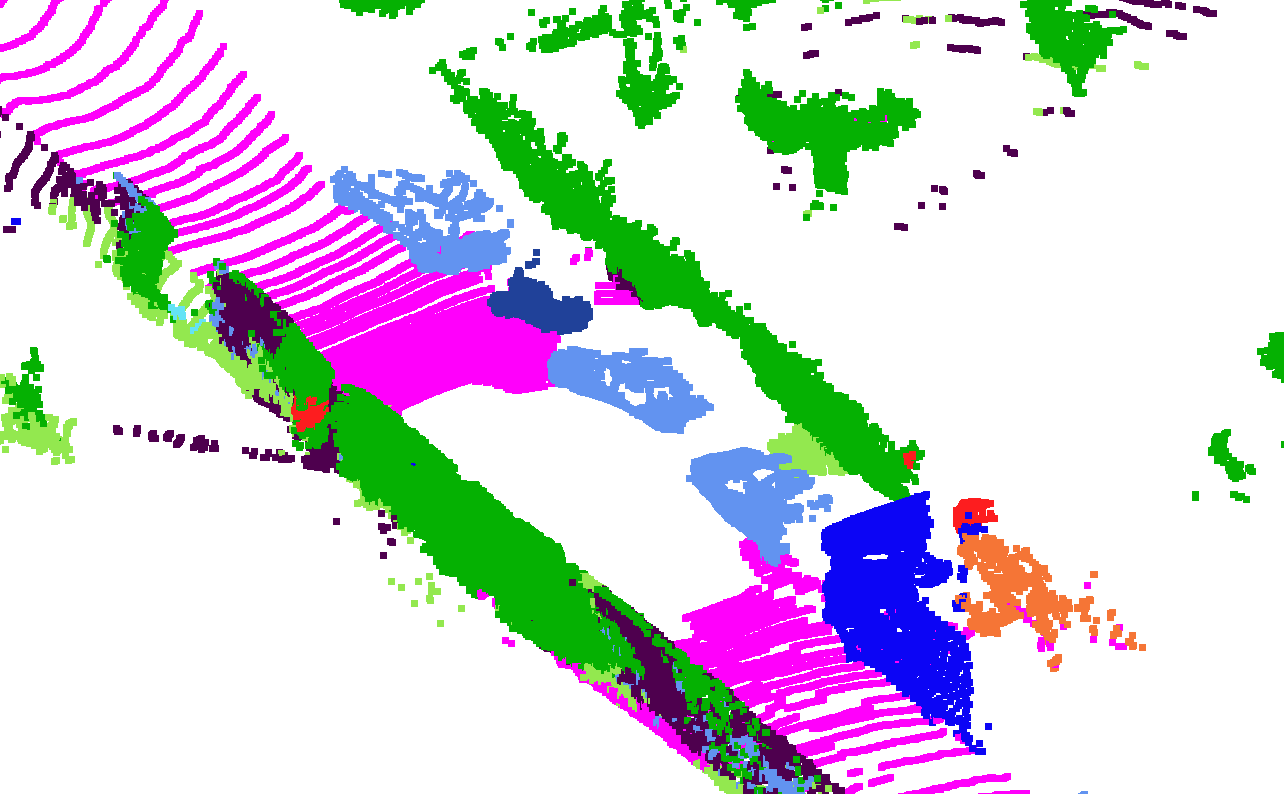}
 & 
\includegraphics[trim=40 0 20 10,clip,width=\widthimage\linewidth]{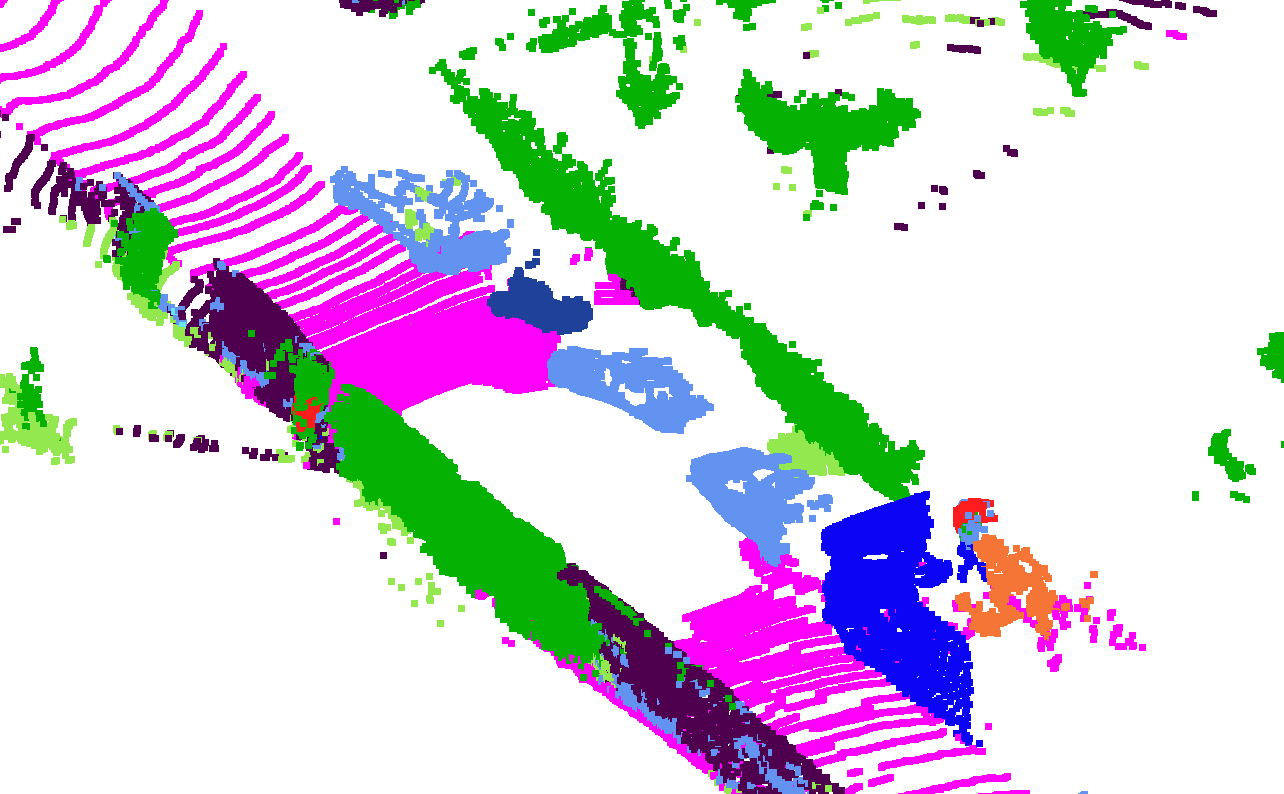}&

\begin{tikzpicture}
    \node[anchor=south west,inner sep=0] at (0,0) {\includegraphics[trim=40 0 20 10,clip,width=\widthimage\linewidth]{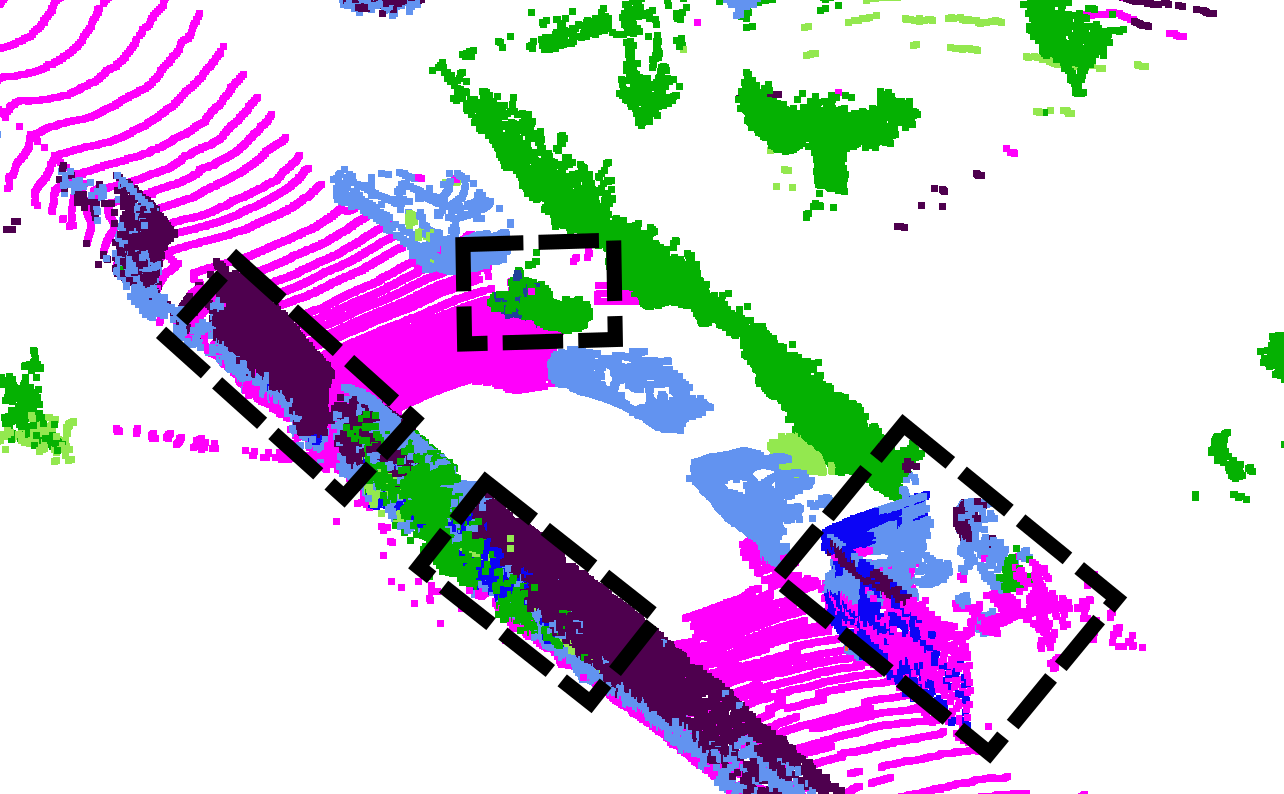}};






    


\end{tikzpicture}

\\
& GT &  Src.-only (start point) & Best model & After 20k iterations \\[2mm]

\rotatebox{90}{\enspace\enspace SHOT }
&
\includegraphics[trim=40 0 20 10,clip,width=\widthimage\linewidth]{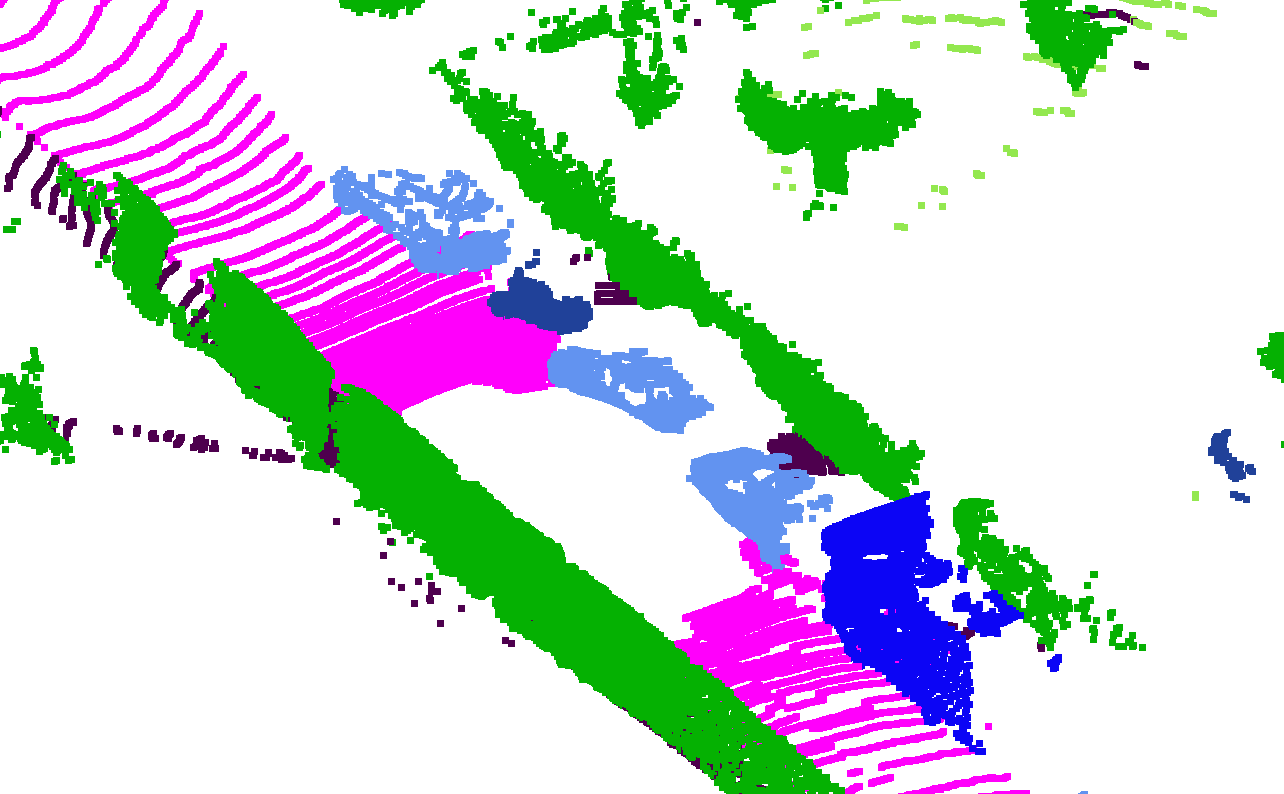}&
\includegraphics[trim=40 0 20 10,clip,width=\widthimage\linewidth]{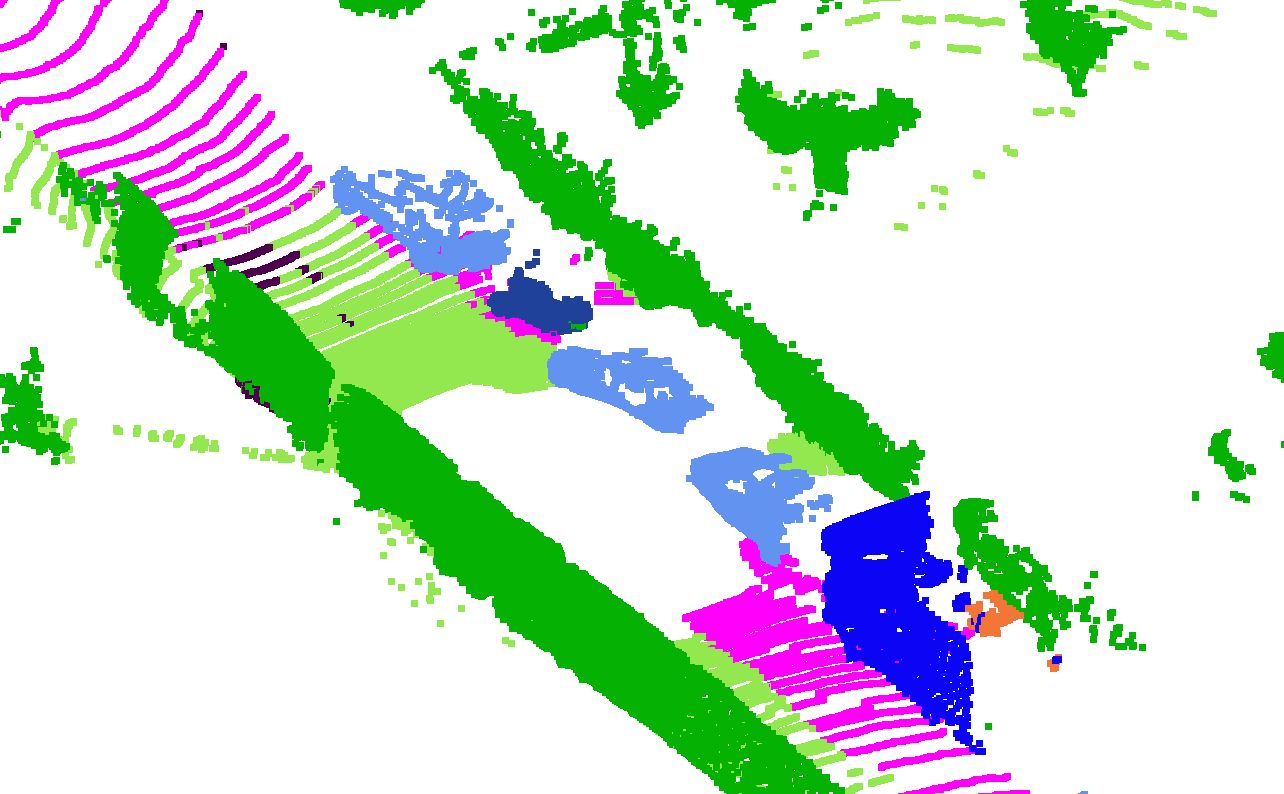}
 & 
\includegraphics[trim=40 0 20 10,clip,width=\widthimage\linewidth]{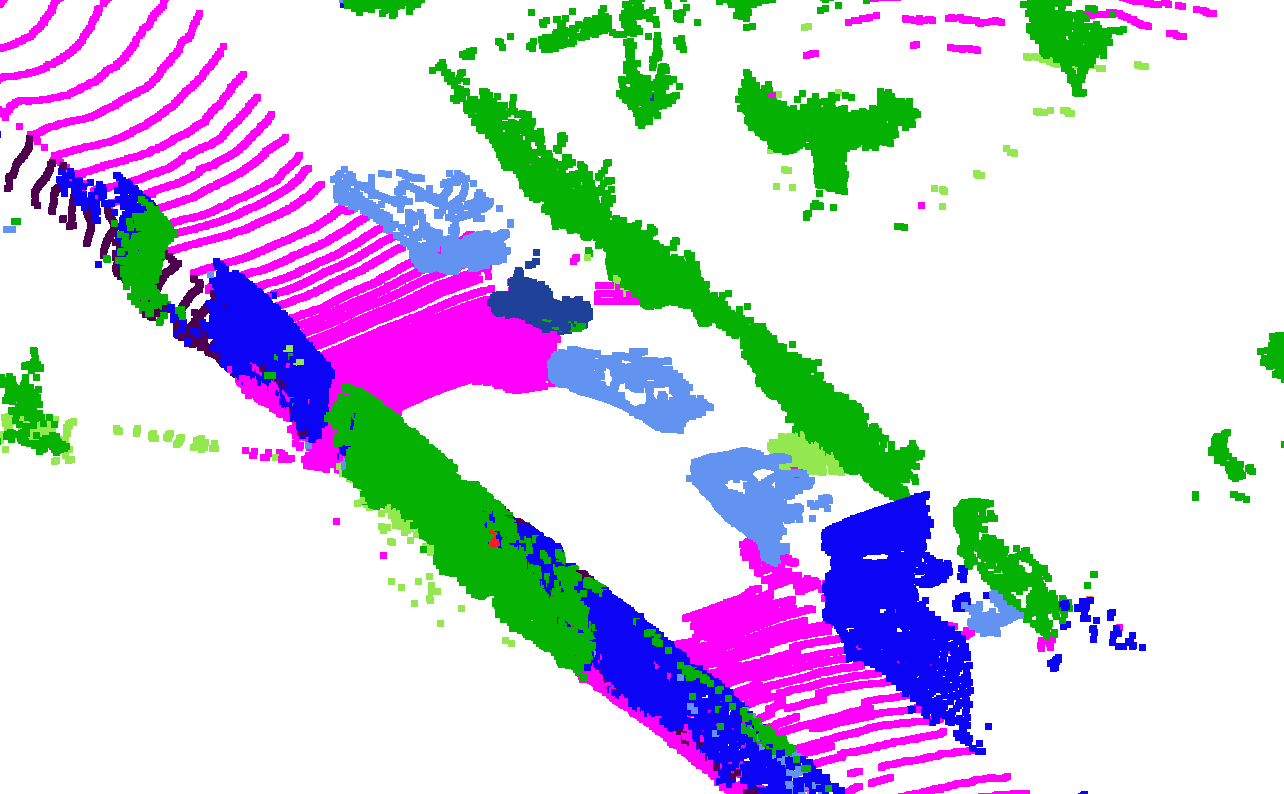}&
\begin{tikzpicture}
    \node[anchor=south west,inner sep=0] at (0,0) {\includegraphics[trim=40 0 20 10,clip,width=\widthimage\linewidth]{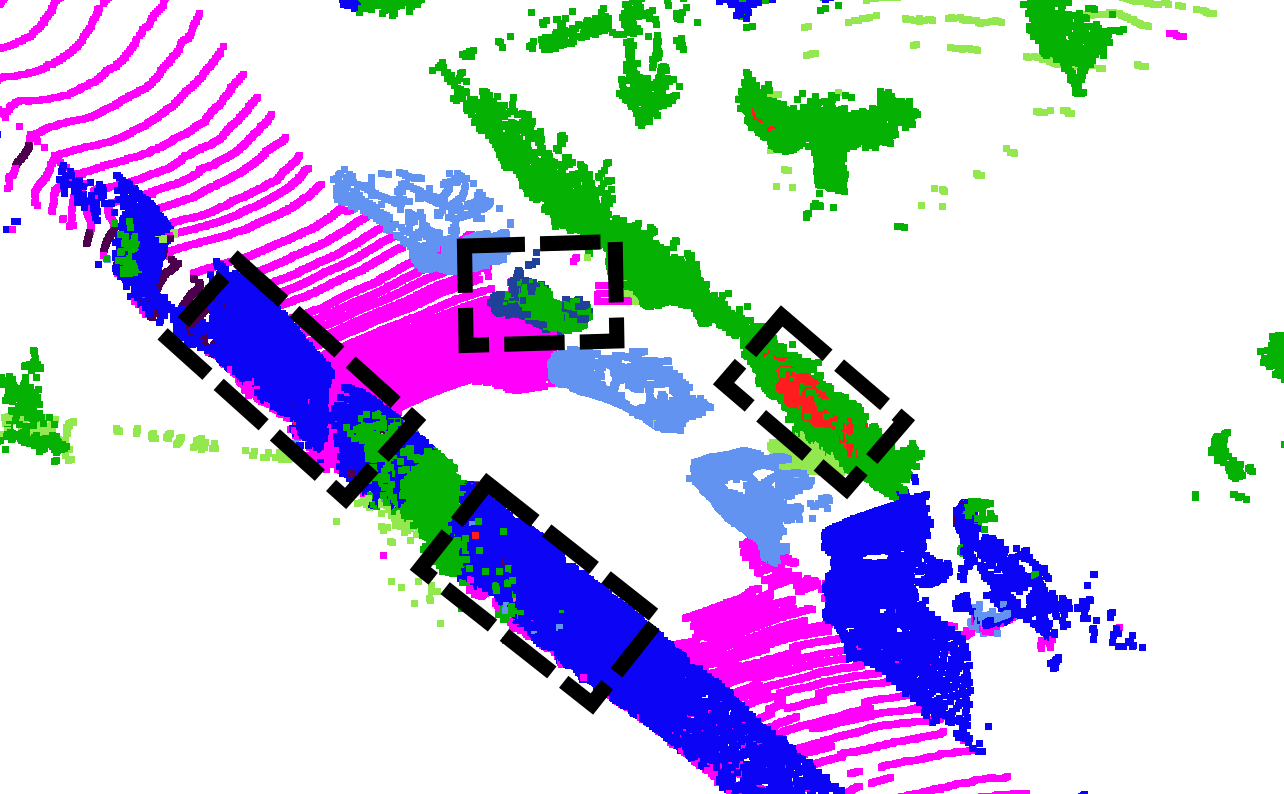}};

     
    
    
     

    

\end{tikzpicture}

\\
& GT &  Src.-only (start point) & Best model & After 20k iterations \\[2mm]

\rotatebox{90}{\enspace URMDA }
&
\includegraphics[trim=40 0 20 10,clip,width=\widthimage\linewidth]{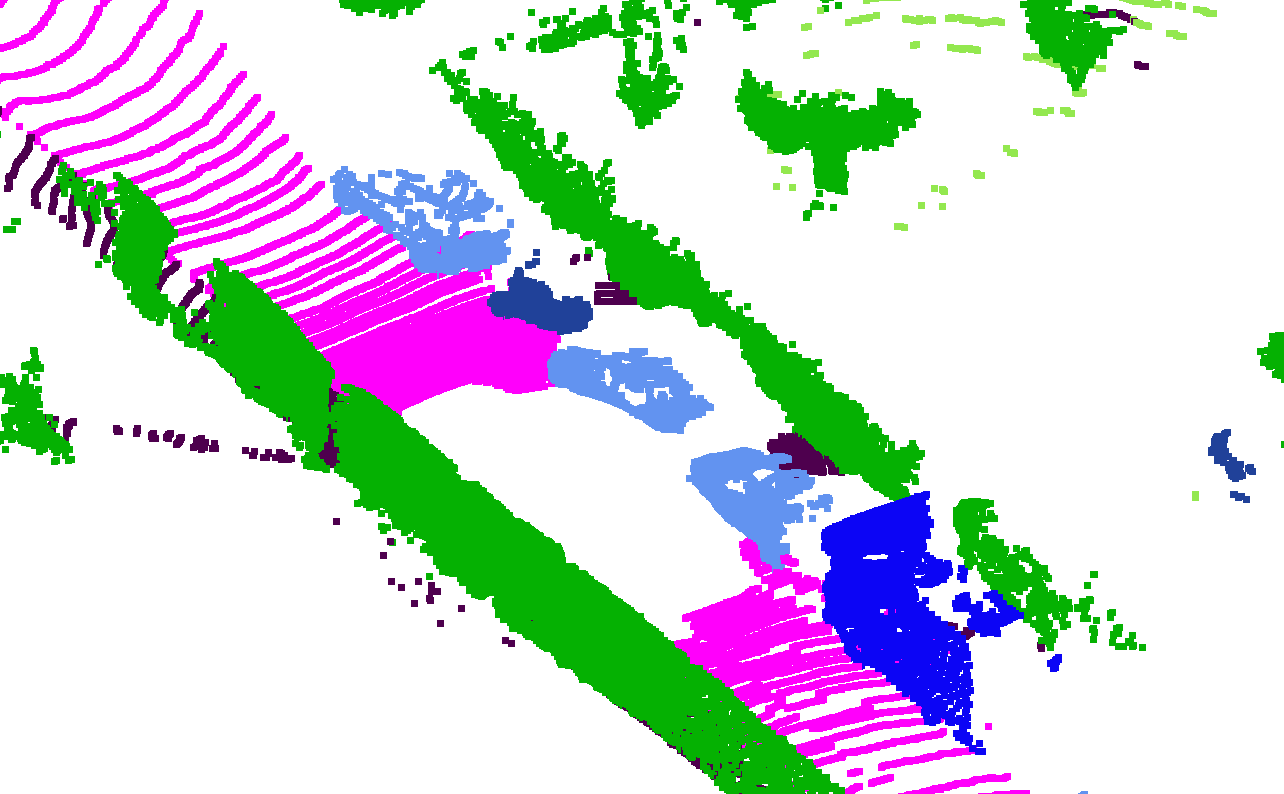}&
\includegraphics[trim=40 0 20 10,clip,width=\widthimage\linewidth]{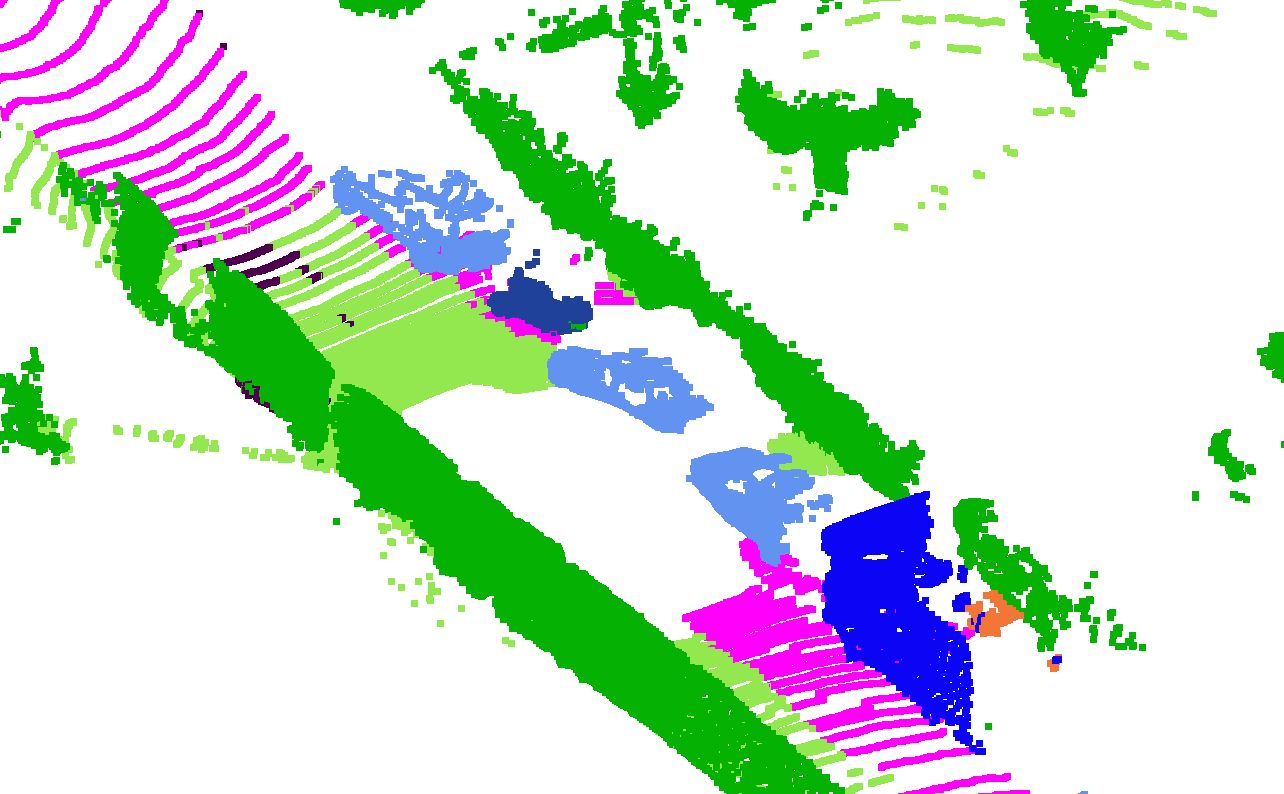}
 & 
\includegraphics[trim=40 0 20 10,clip,width=\widthimage\linewidth]{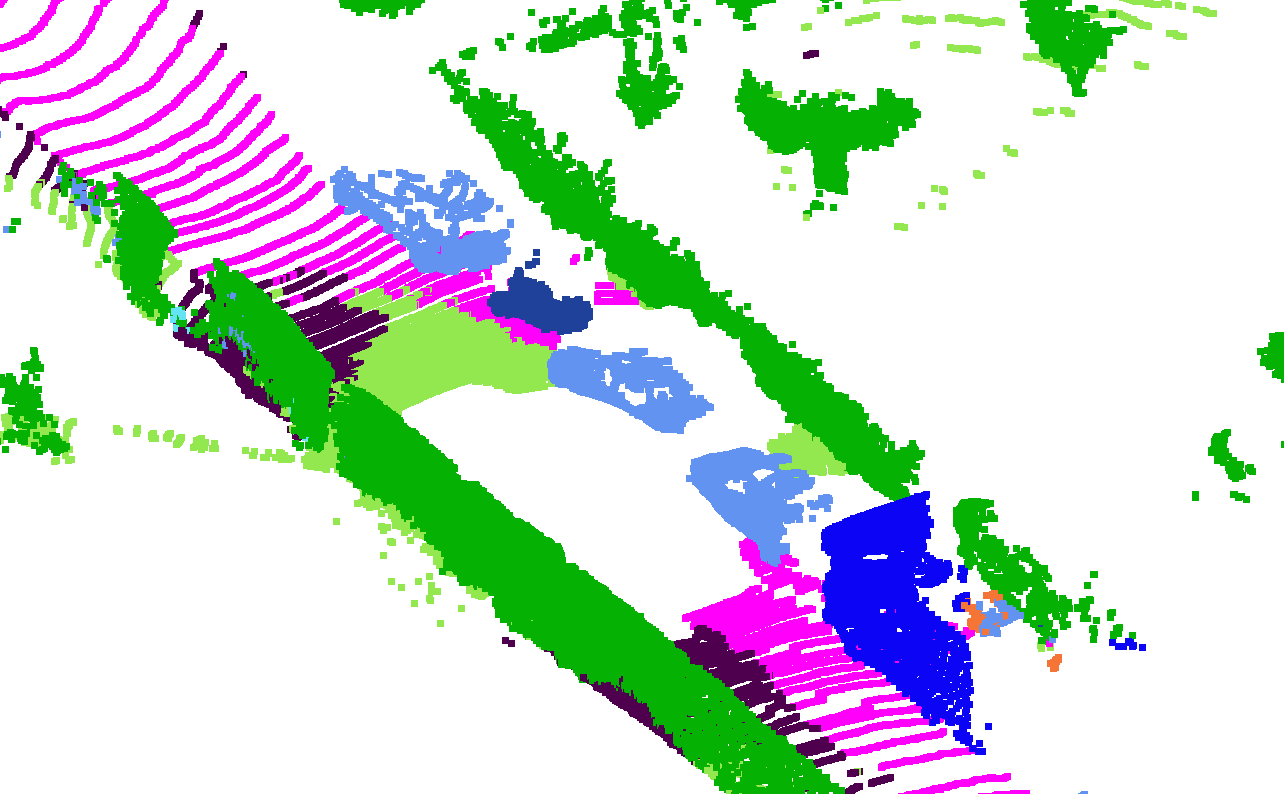}&

\begin{tikzpicture}
    \node[anchor=south west,inner sep=0] at (0,0) {\includegraphics[trim=40 0 20 10,clip,width=\widthimage\linewidth]{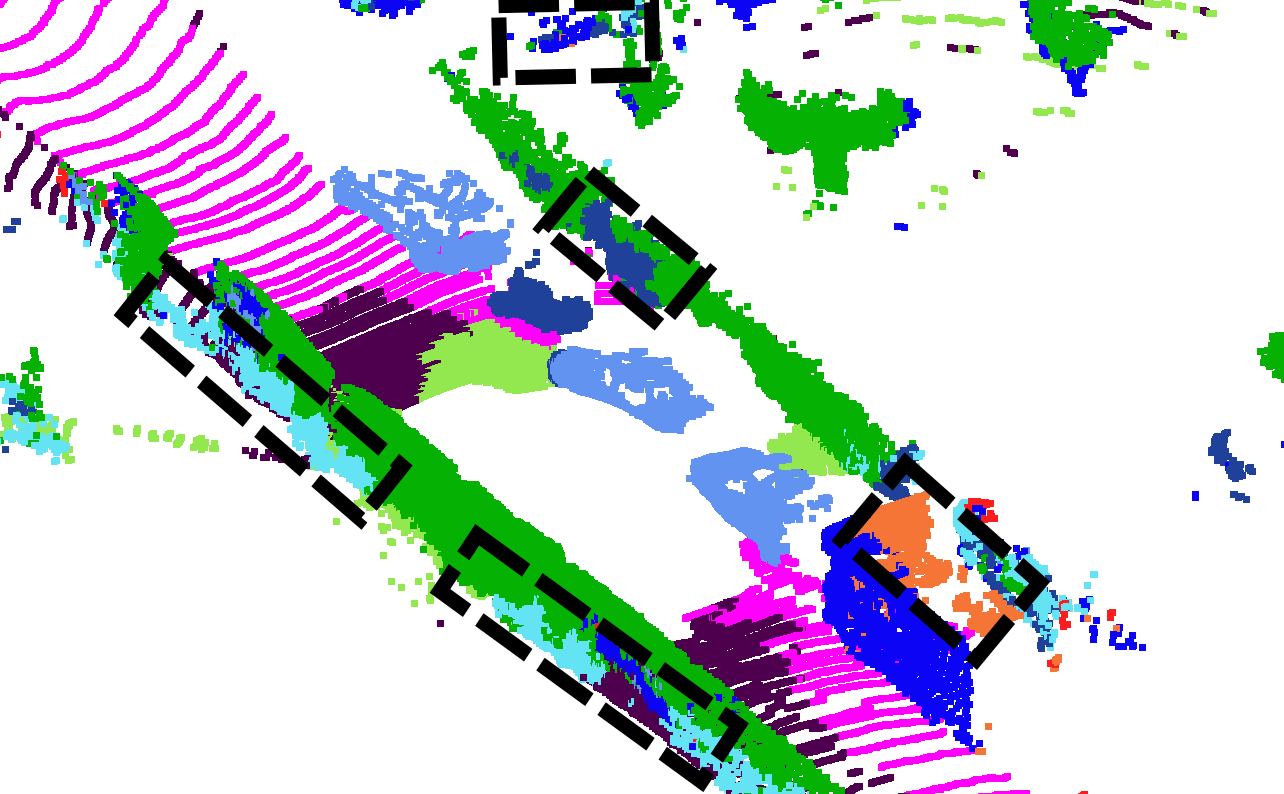}};


     


    

\end{tikzpicture}

\\
& GT & Src.-only (start point) & Best model & After 20k iterations \\

\end{tabular}

\caption{Examples of results with TENT~\cite{wangtent}, SHOT~\cite{shotliang20a} and URMDA~\cite{teja2021uncertainty} on \nstosk: ground truth (GT), initial model trained only on source data, best model as upper bound (using ground-truth knowledge of the target validation set), and ``full'' training for 20k iterations. ``Ignore'' points are removed for a better visualisation. Notable errors due to degradation are marked with a dashed rectangle.}
\label{fig:app:qualitative_others}
\end{figure*}

We illustrate in~\cref{fig:app:qualitative_others} the performance degradation when training is too long for TENT~\cite{wangtent}, SHOT~\cite{shotliang20a} and URMDA~\cite{teja2021uncertainty}. Note that, for these methods, we select the best trained model by looking at the ground-truth target validation set. It highlights the difference between what can be achieved in theory and what actually happens if training is not stopped with a criterion like ours.

One can observe that the TENT model, which estimates the normalization parameters of the batch norm layers on the target dataset, starts from a better source-only model, although it has not been trained on target data yet.  After 20k iterations, the motorcycle, the truck, and part of the vegetation are not correctly classified, although they were correctly classified in the source-only model. A similar degradation behavior can be seen for the SHOT method. The URMDA method does not perform as well as the others. After 20k iterations, it also shows a significant degradation with respect to both the source-only starting point and the best model: while the source-only model correctly segments the vegetation and the truck, the final model incorrectly labels part of the vegetation using various other classes, and wrongly predicts the class on the top of the truck.

\paragraph{Our stopping criterion.}

\begin{figure*}
\def\widthimage{0.21}
\newcommand{\rotext}[1]{{\begin{turn}{90}{#1}\end{turn}}}
\setlength{\tabcolsep}{1pt}
\centering
\begin{tabular}{c@{~}cccc}

\rotatebox{90}{\enspace\nstosk }
&
\includegraphics[trim=40 0 20 10,clip,width=\widthimage\linewidth]{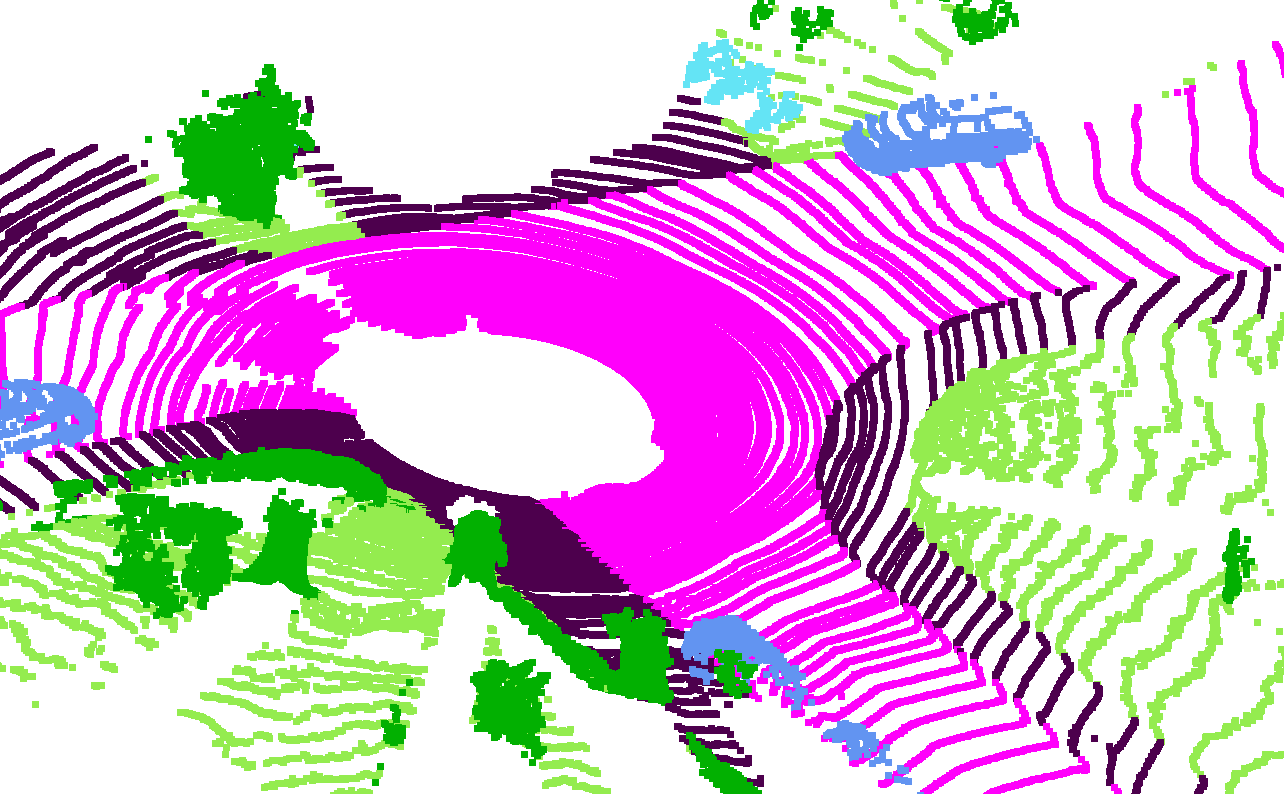}&
\includegraphics[trim=40 0 20 10,clip,width=\widthimage\linewidth]{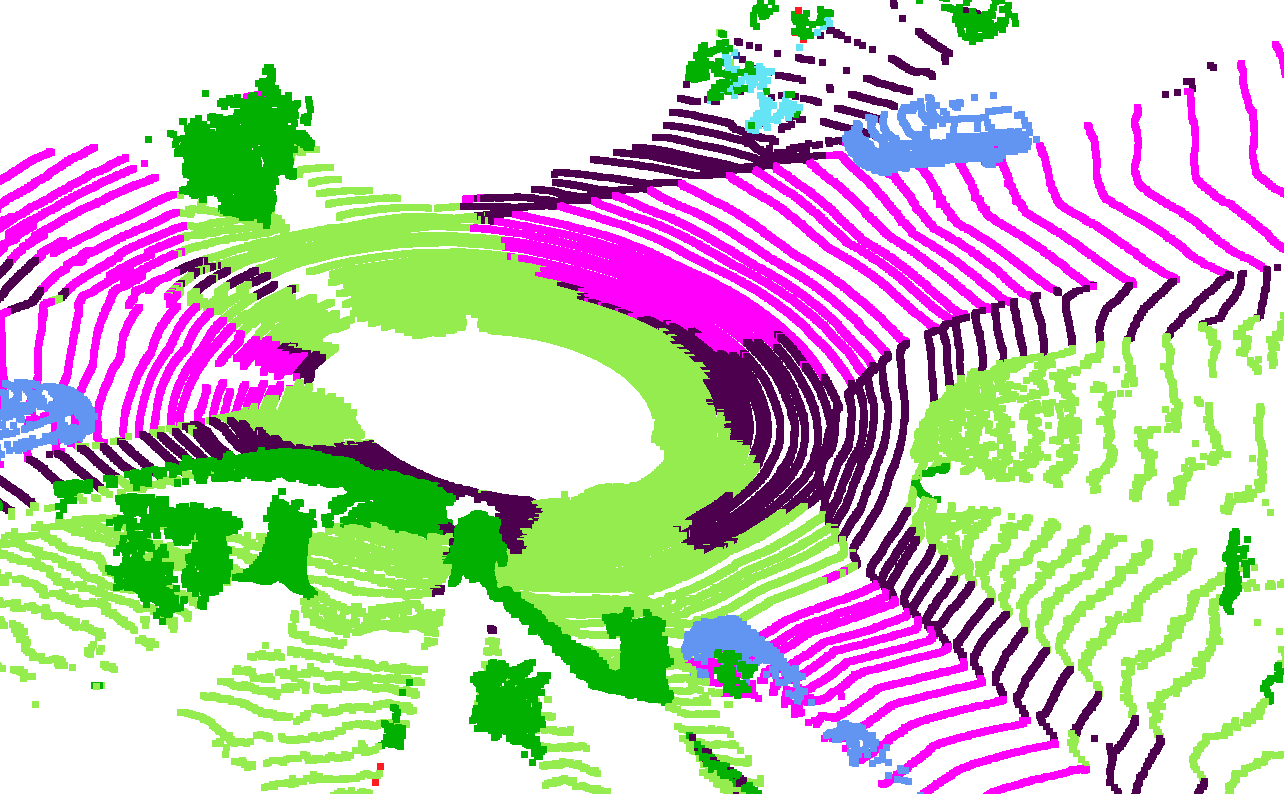}
 & 
\includegraphics[trim=40 0 20 10,clip,width=\widthimage\linewidth]{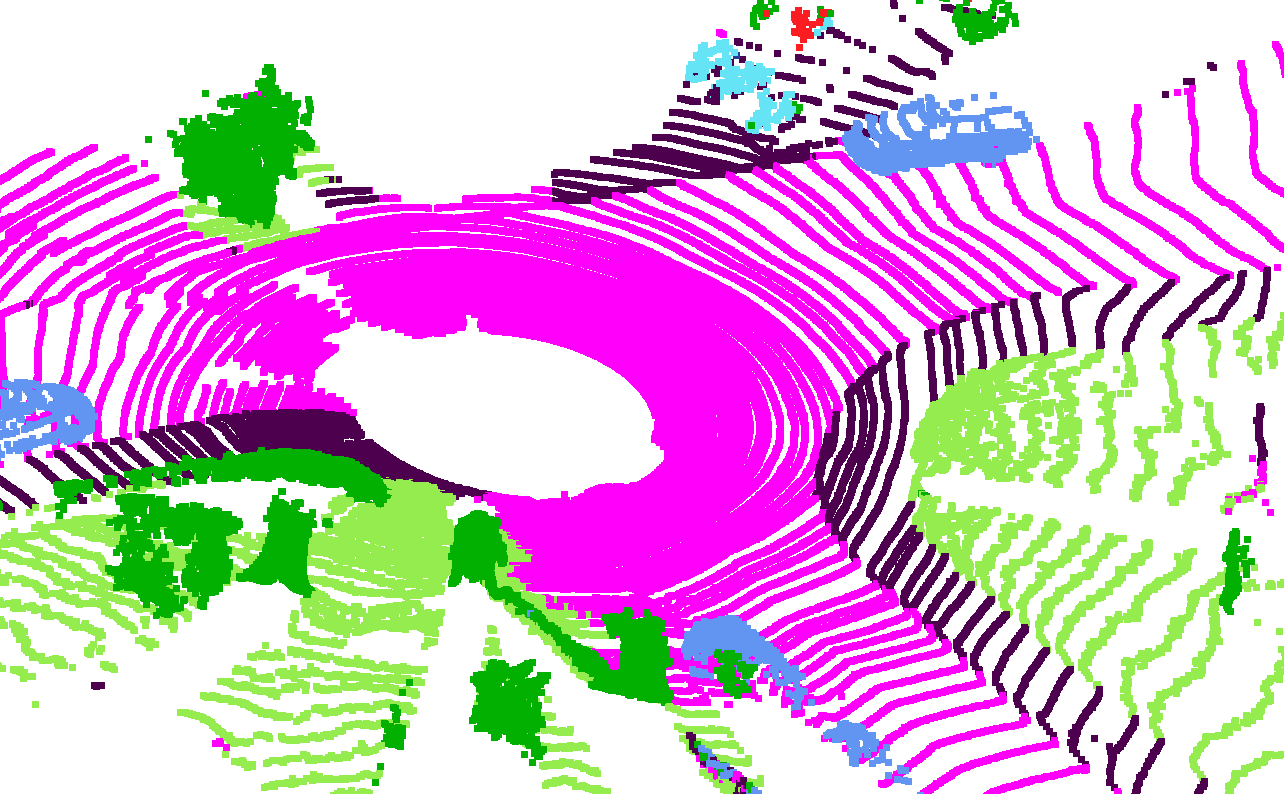}&
\begin{tikzpicture}
    \node[anchor=south west,inner sep=0] at (0,0) {\includegraphics[trim=40 0 20 10,clip,width=\widthimage\linewidth]{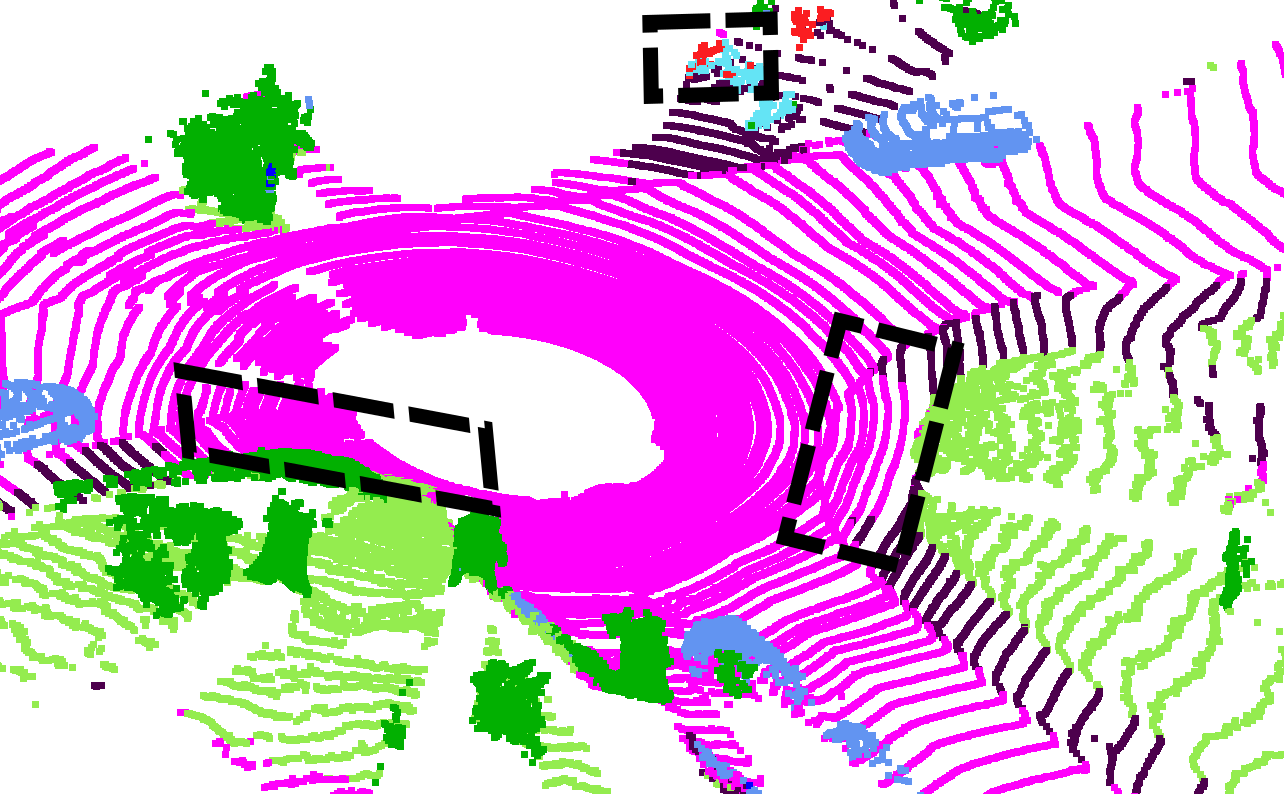}};
    

    

    

\end{tikzpicture}

\\
& GT & Src.-only (start point) & \methodstop & After 20k iterations \\[2mm]

\rotatebox{90}{\enspace \synthtosk }
&
\includegraphics[trim=40 0 20 10,clip,width=\widthimage\linewidth]{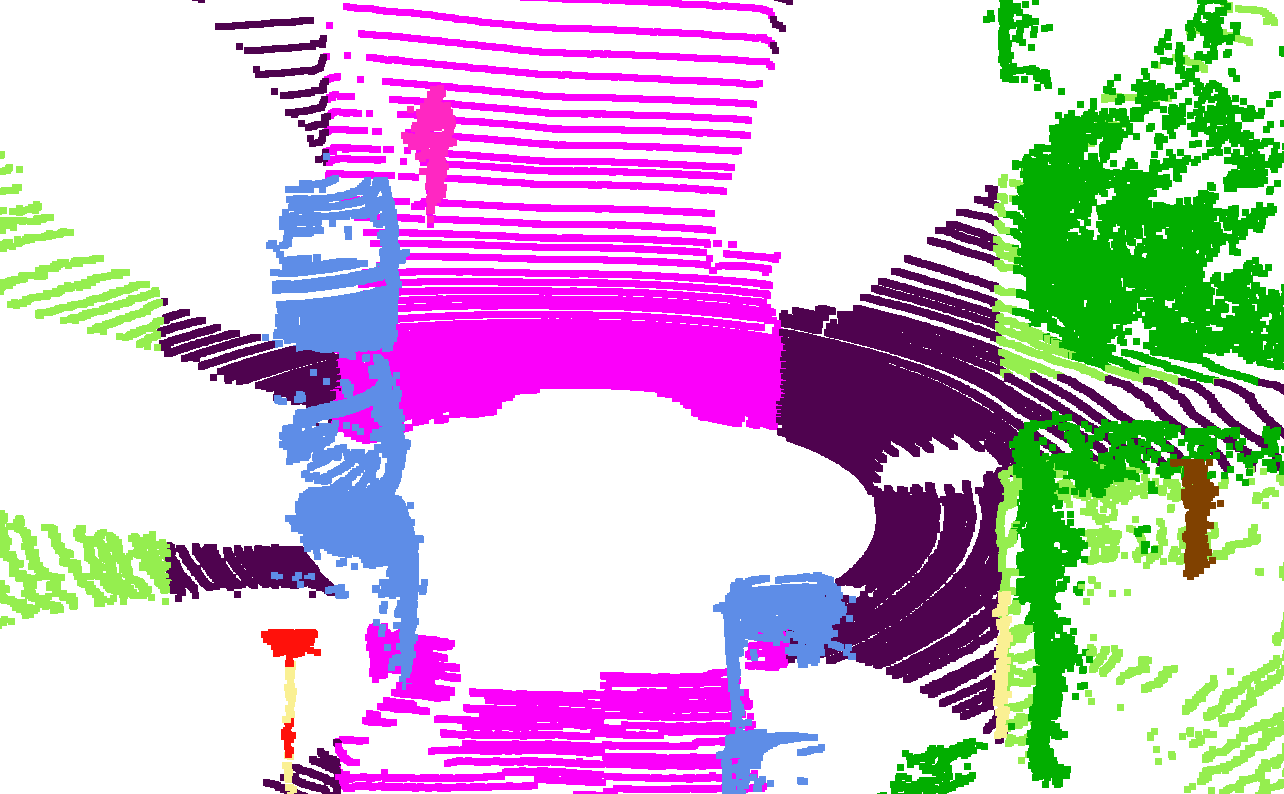}&
\includegraphics[trim=40 0 20 10,clip,width=\widthimage\linewidth]{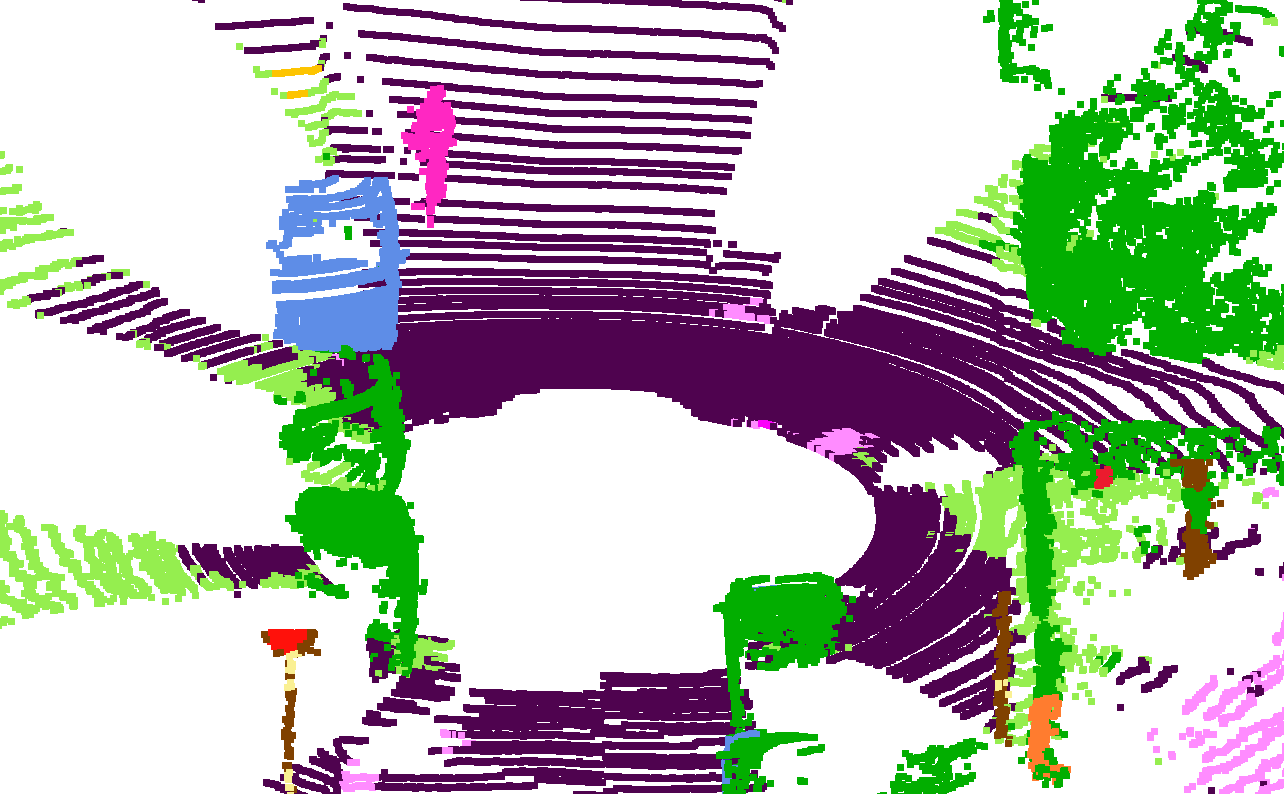}
 & 
\includegraphics[trim=40 0 20 10,clip,width=\widthimage\linewidth]{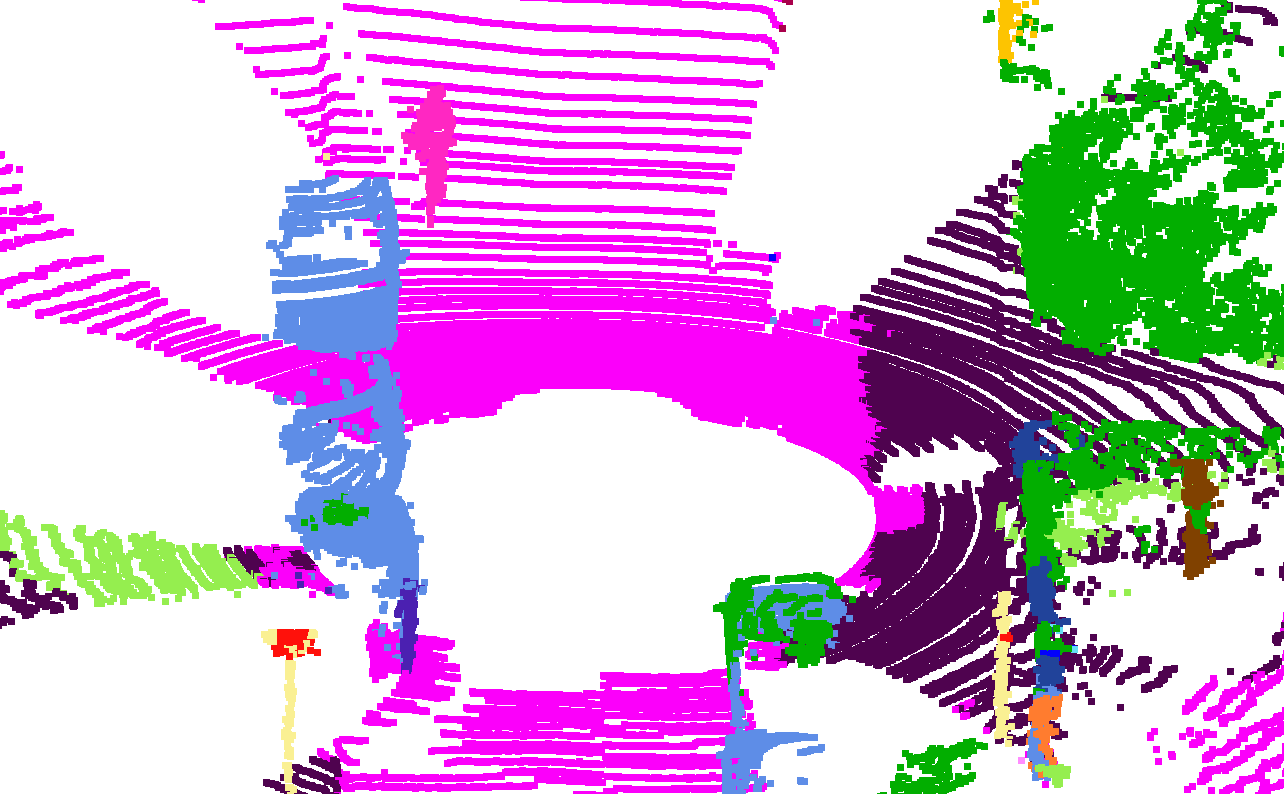}&
\begin{tikzpicture}

\node[anchor=south west,inner sep=0] at (0,0) {\includegraphics[trim=40 0 20 10,clip,width=\widthimage\linewidth]{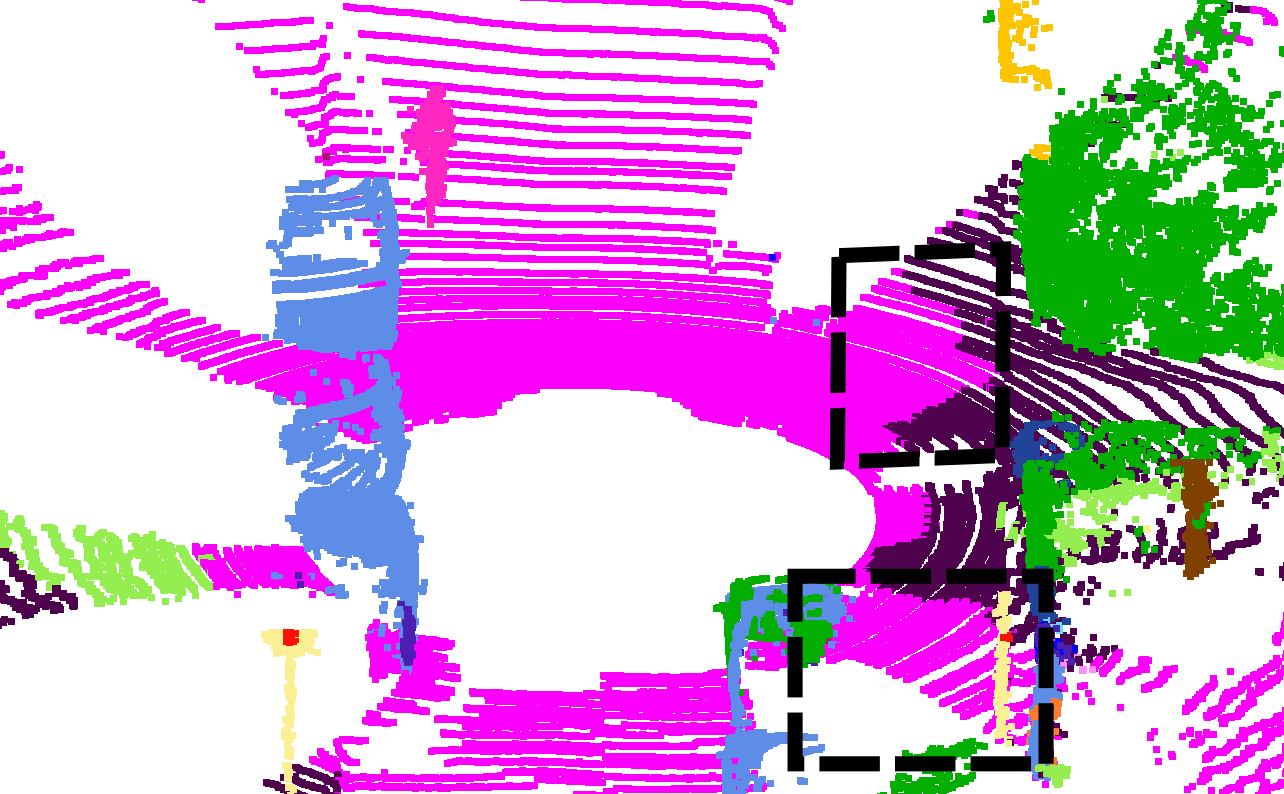}};





\end{tikzpicture}
    
\\
& GT & Src.-only (start point) & \methodstop & After 20k iterations \\[2mm]

\rotatebox{90}{\enspace \synthtoposs}
&
\includegraphics[trim=40 0 20 10,clip,width=\widthimage\linewidth]{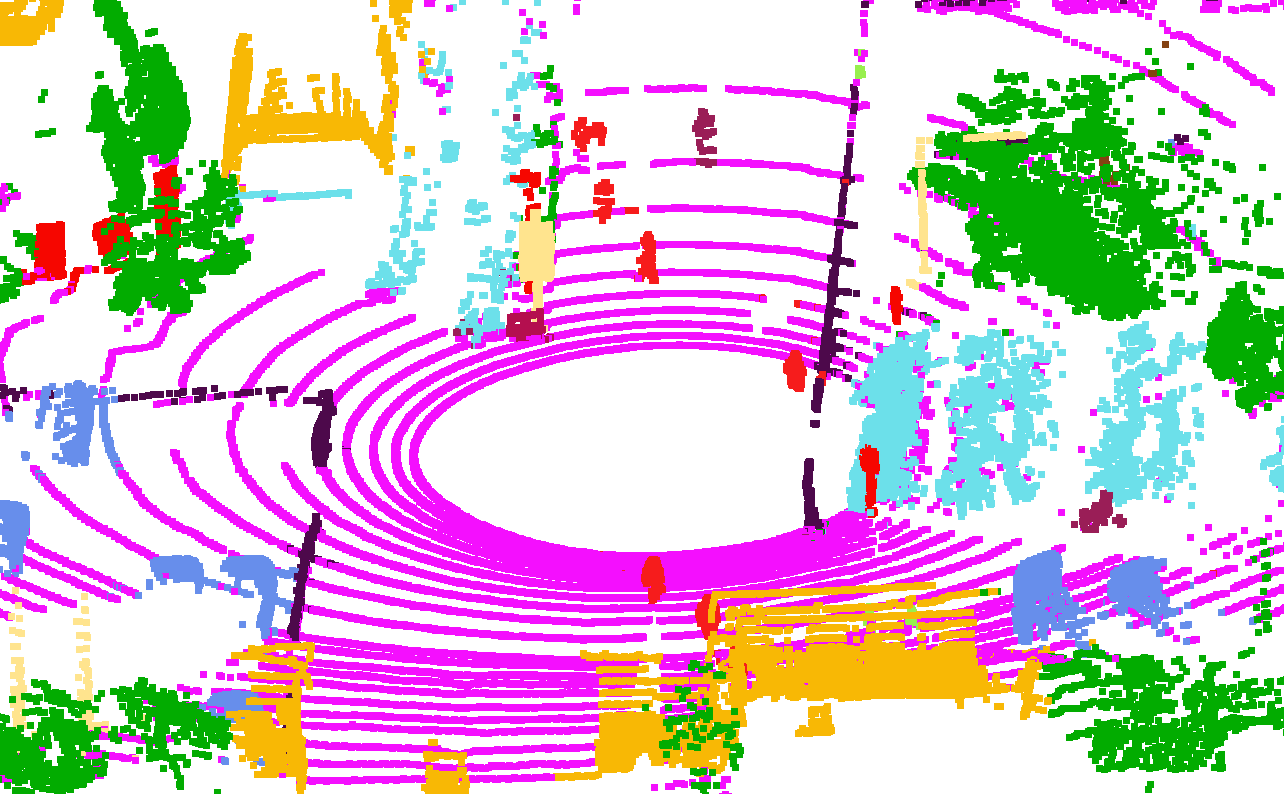}&
\includegraphics[trim=40 0 20 10,clip,width=\widthimage\linewidth]{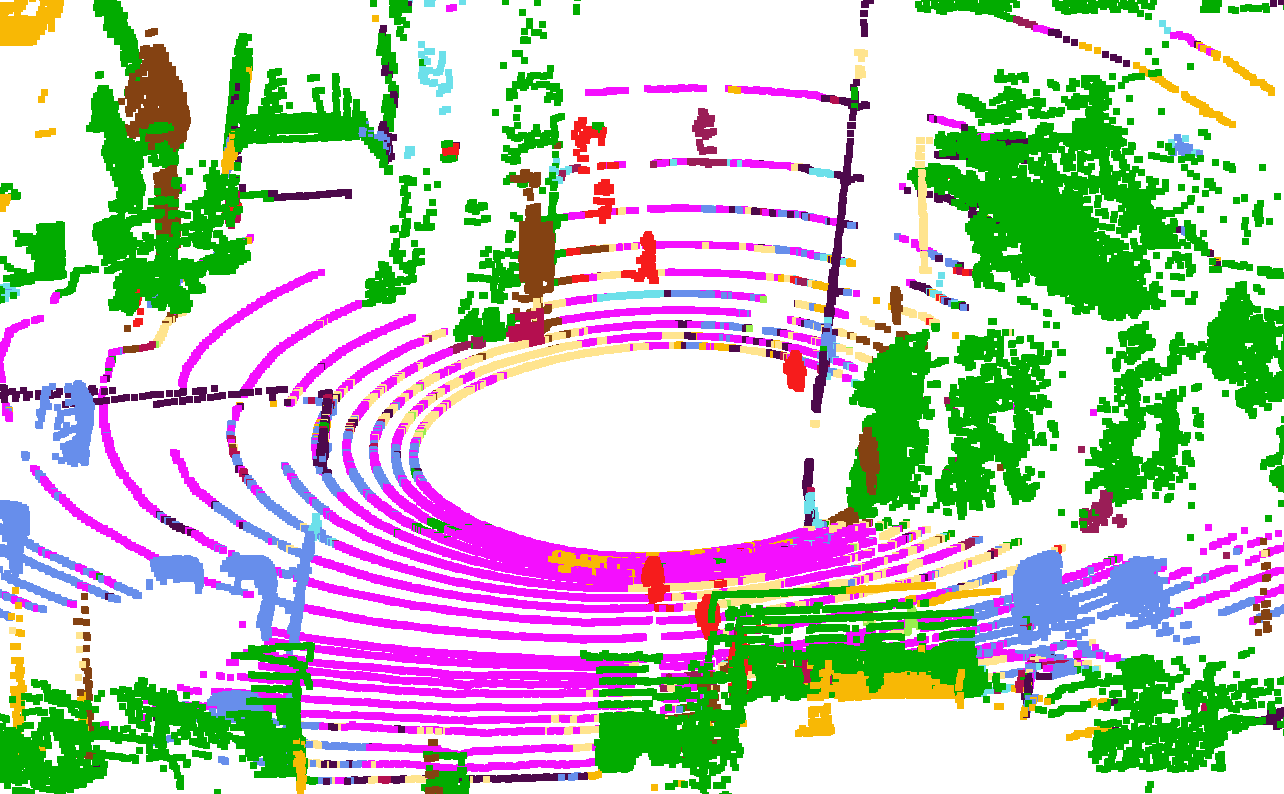}
 & 
\includegraphics[trim=40 0 20 10,clip,width=\widthimage\linewidth]{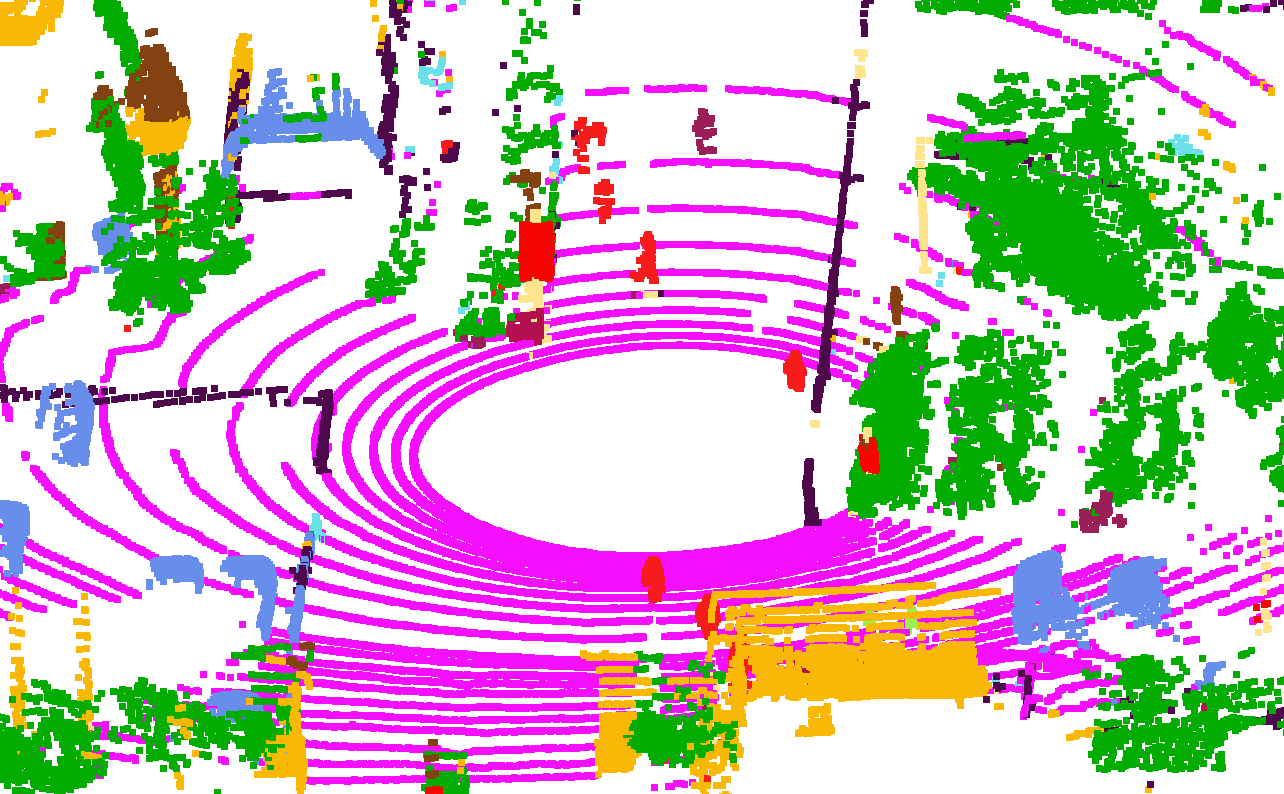}&

\begin{tikzpicture}
\node[anchor=south west,inner sep=0] at (0,0) {\includegraphics[trim=40 0 20 10,clip,width=\widthimage\linewidth]{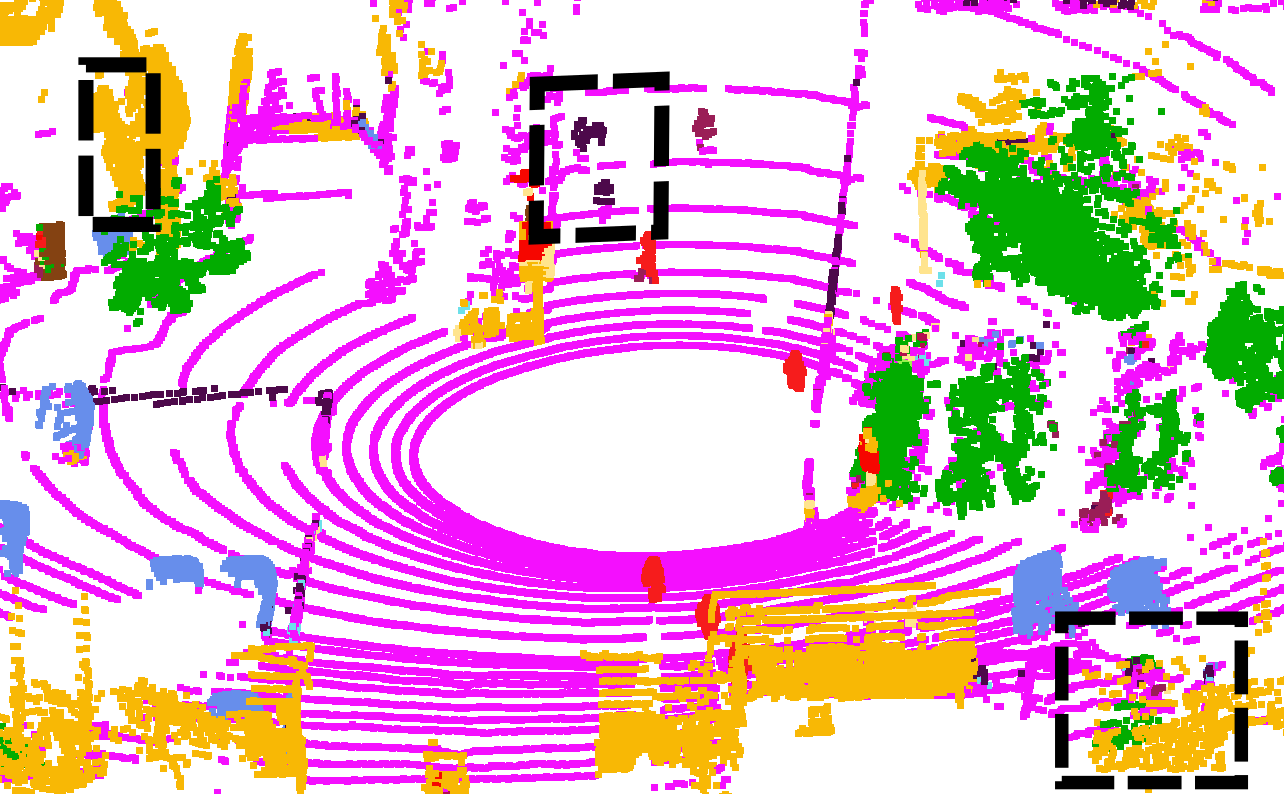}};







\end{tikzpicture}\\
& GT & Src.-only (start point) & \methodstop & After 20k iterations \\[2mm]

\rotatebox{90}{\enspace \nstoposs}
&
\includegraphics[trim=40 0 20 10,clip,width=\widthimage\linewidth]{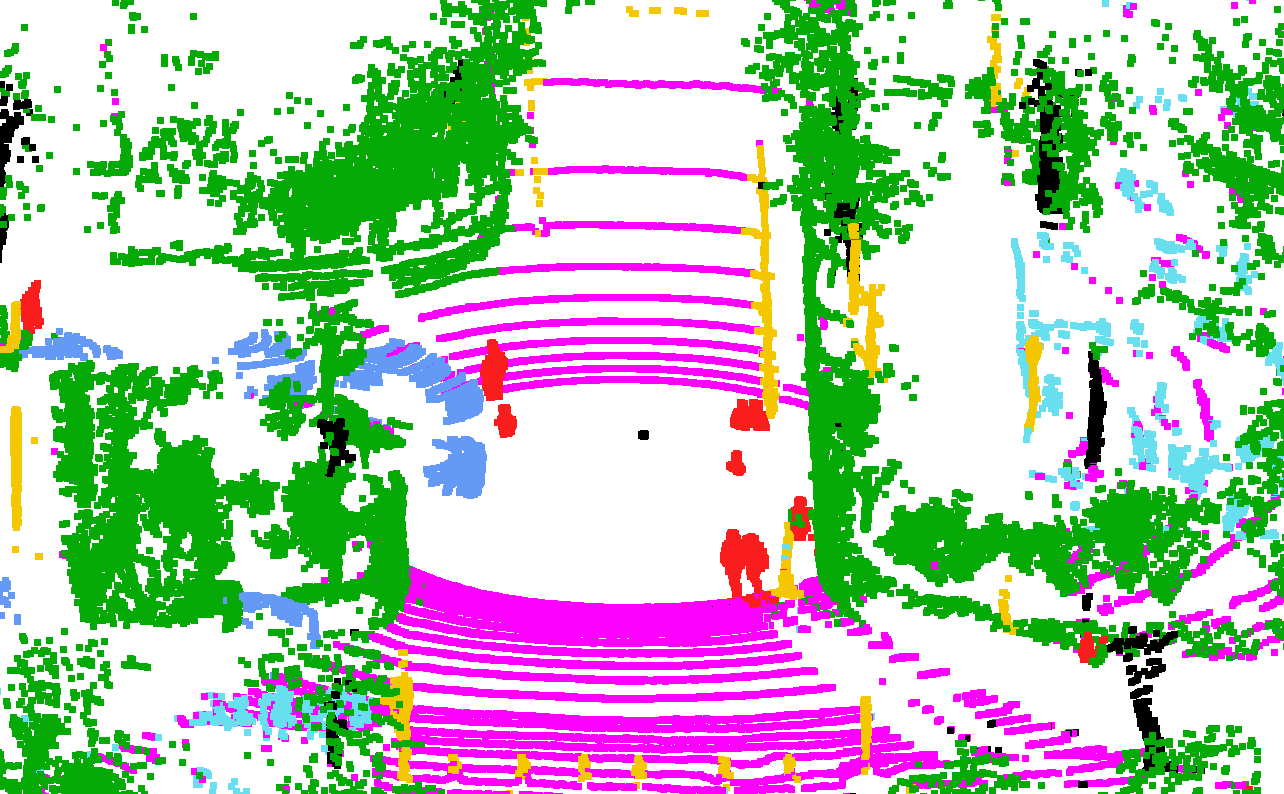}&
\includegraphics[trim=40 0 20 10,clip,width=\widthimage\linewidth]{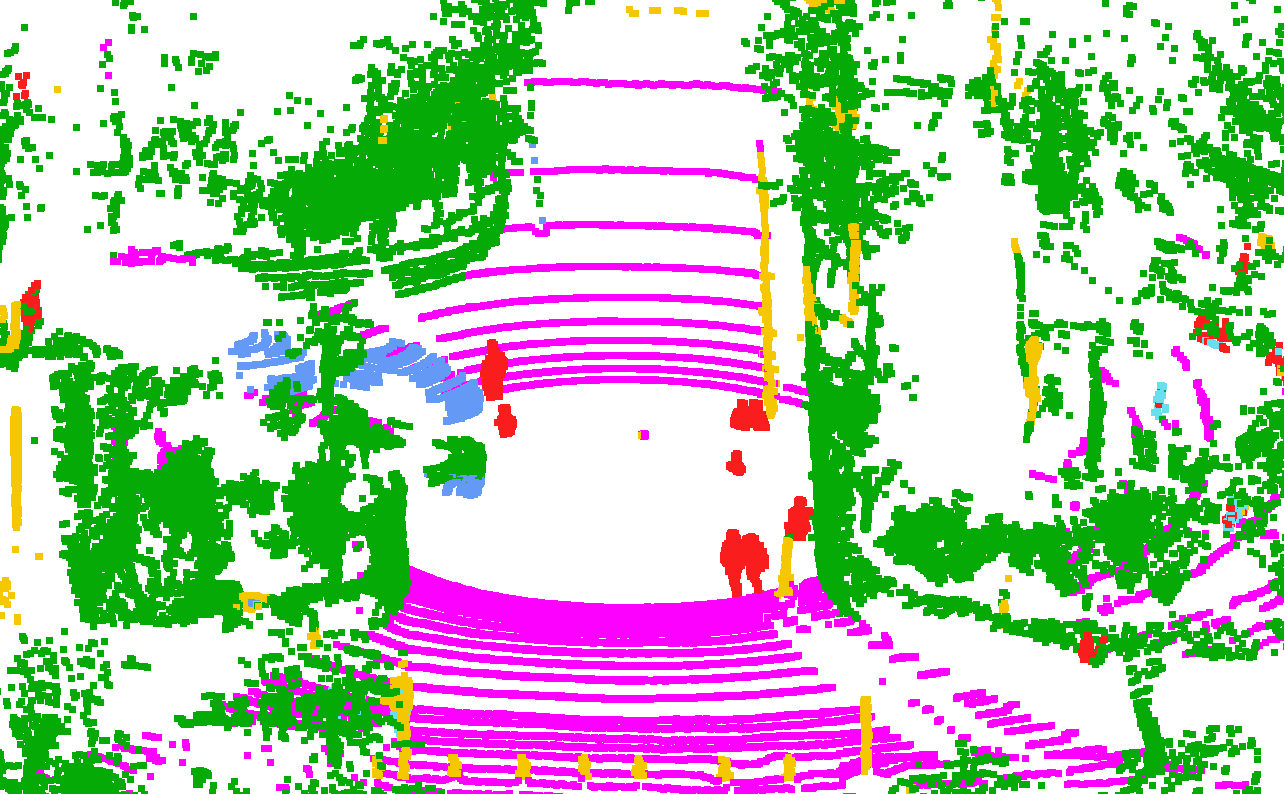}
 & 
\includegraphics[trim=40 0 20 10,clip,width=\widthimage\linewidth]{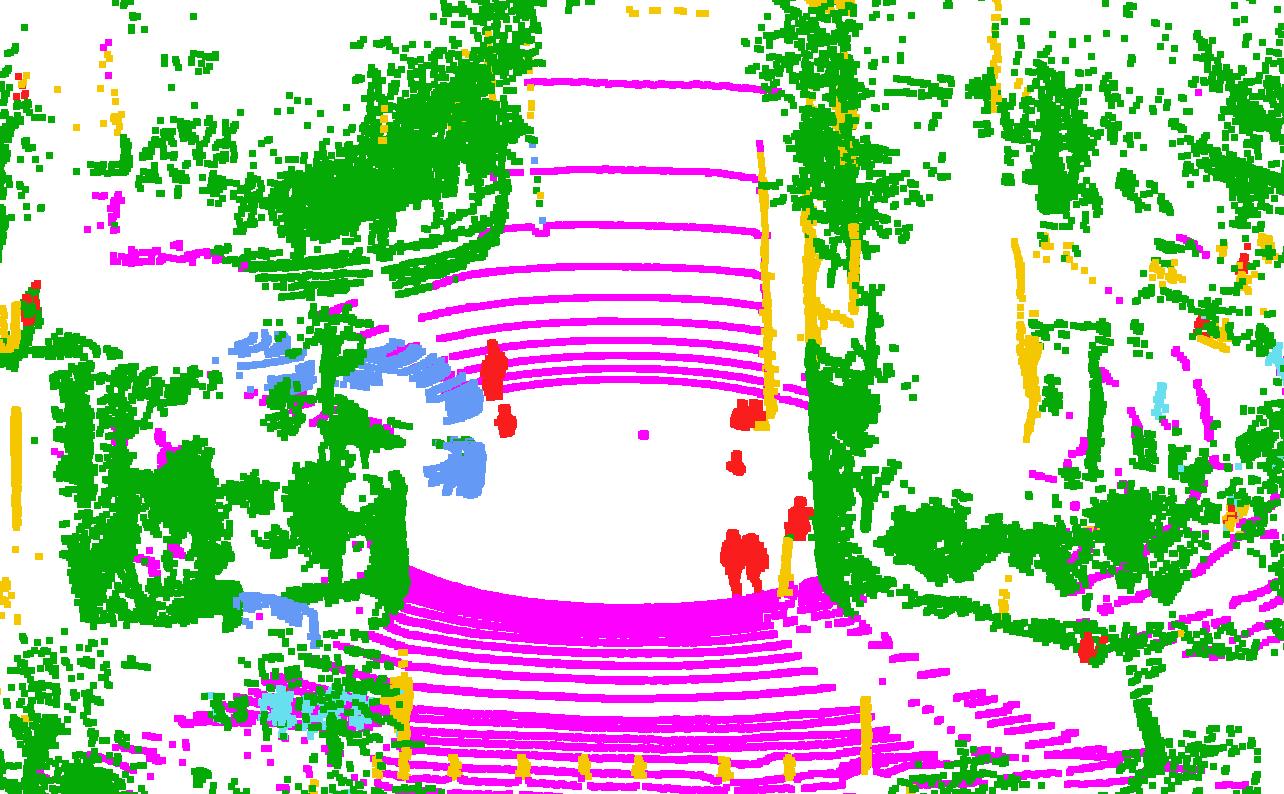}&
\begin{tikzpicture}
    \node[anchor=south west,inner sep=0] at (0,0) {\includegraphics[trim=40 0 20 10,clip,width=\widthimage\linewidth]{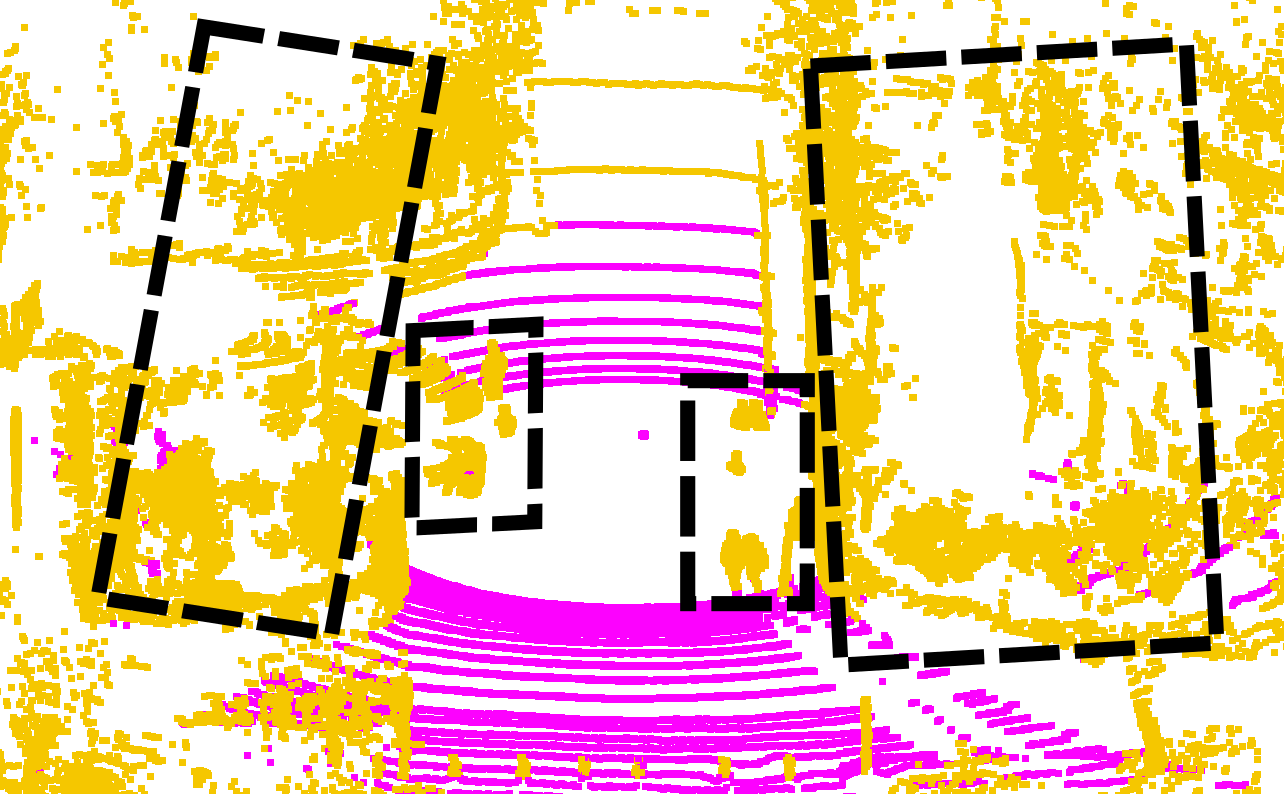}};
        

     




    
\end{tikzpicture}\\

& GT & Src.-only (start point) & \methodstop & After 20k iterations \\[2mm]


\rotatebox{90}{\enspace \nstowy }
&
\includegraphics[trim=40 0 20 10,clip,width=\widthimage\linewidth]{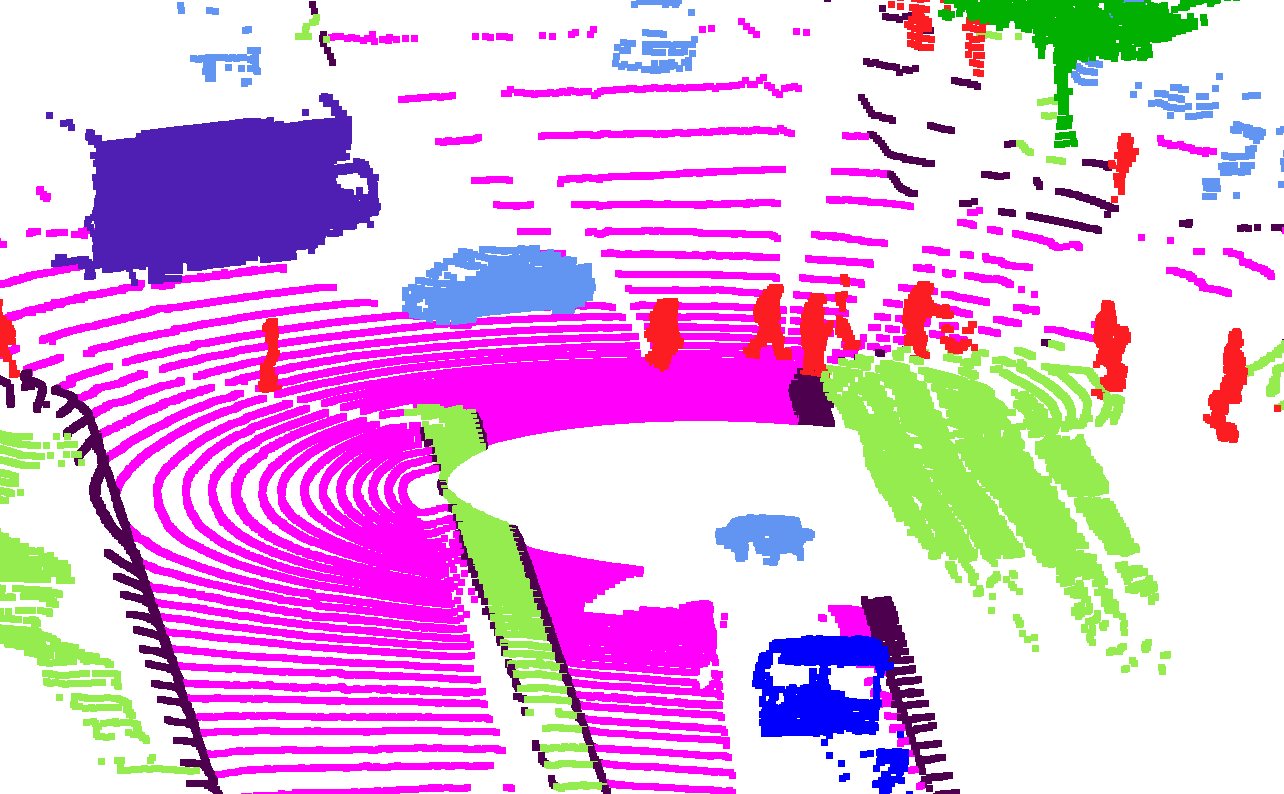}&
\includegraphics[trim=40 0 20 10,clip,width=\widthimage\linewidth]{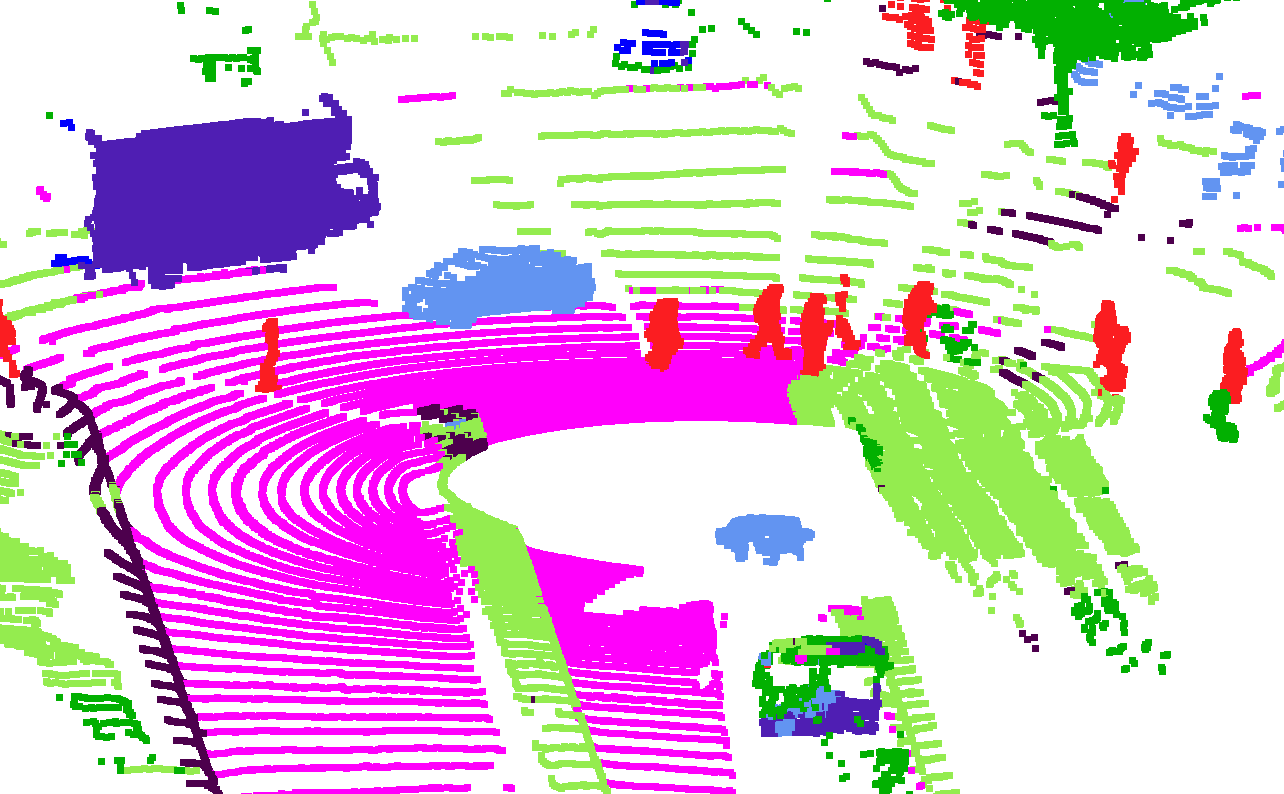}
 & 
\includegraphics[trim=40 0 20 10,clip,width=\widthimage\linewidth]{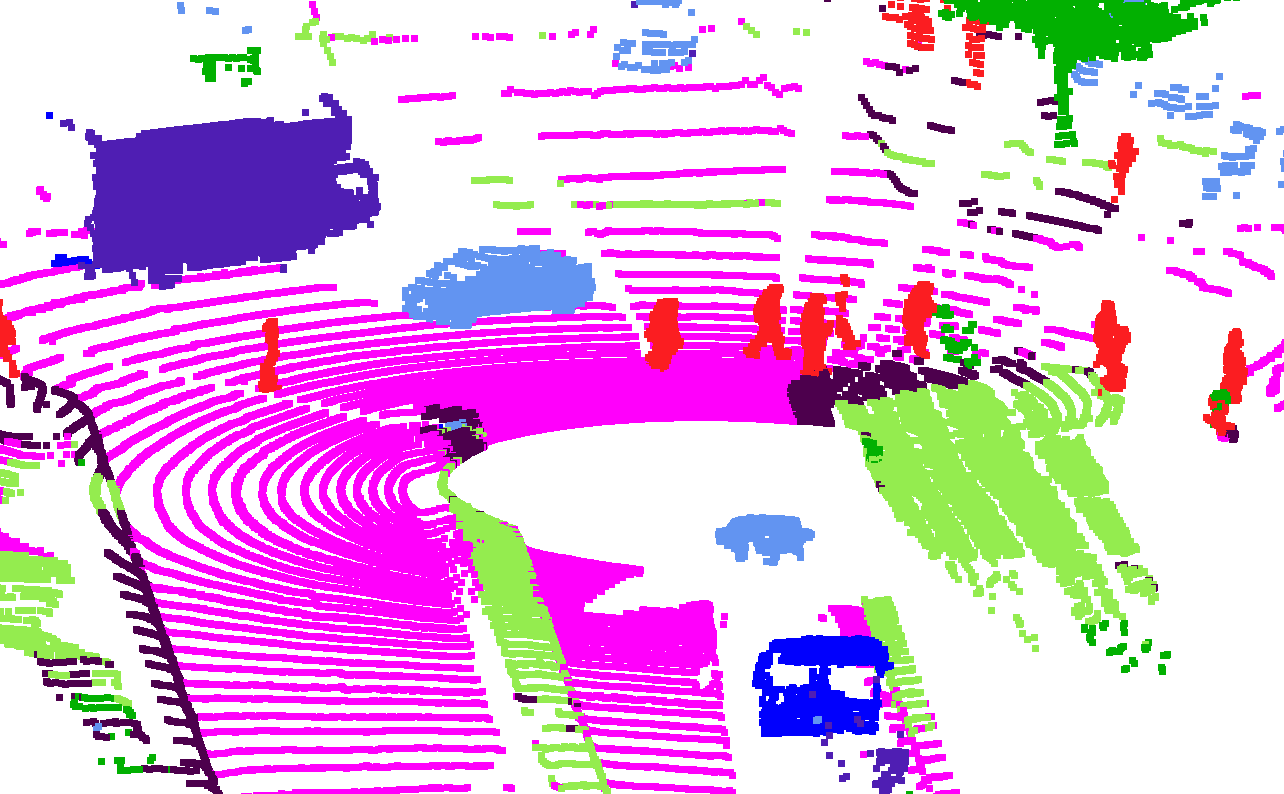}&

\begin{tikzpicture}
    \node[anchor=south west,inner sep=0] at (0,0) {\includegraphics[trim=40 0 20 10,clip,width=\widthimage\linewidth]{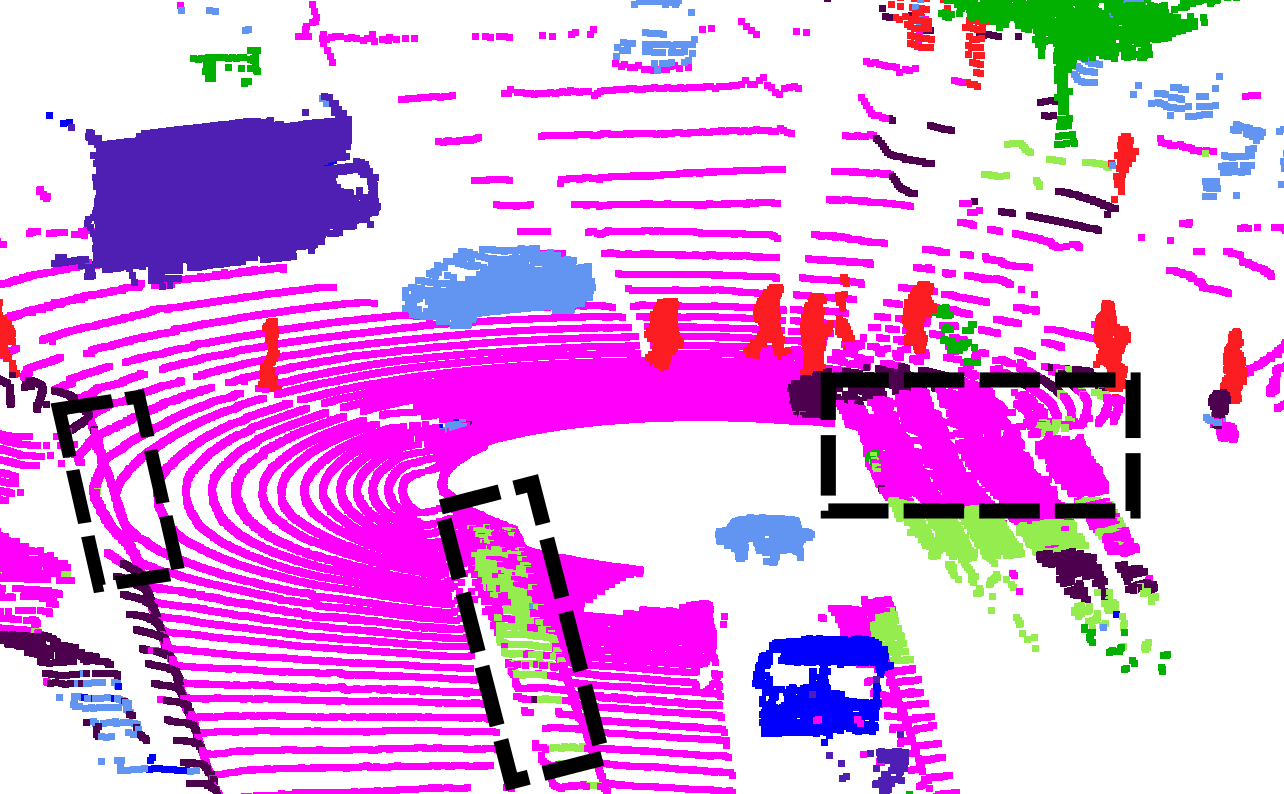}};

    
        

\end{tikzpicture}

\\
& GT & Src.-only (start point) & \methodstop & After 20k iterations \\[2mm]

\rotatebox{90}{\enspace \nstopd}
&
\includegraphics[trim=40 0 20 10,clip,width=\widthimage\linewidth]{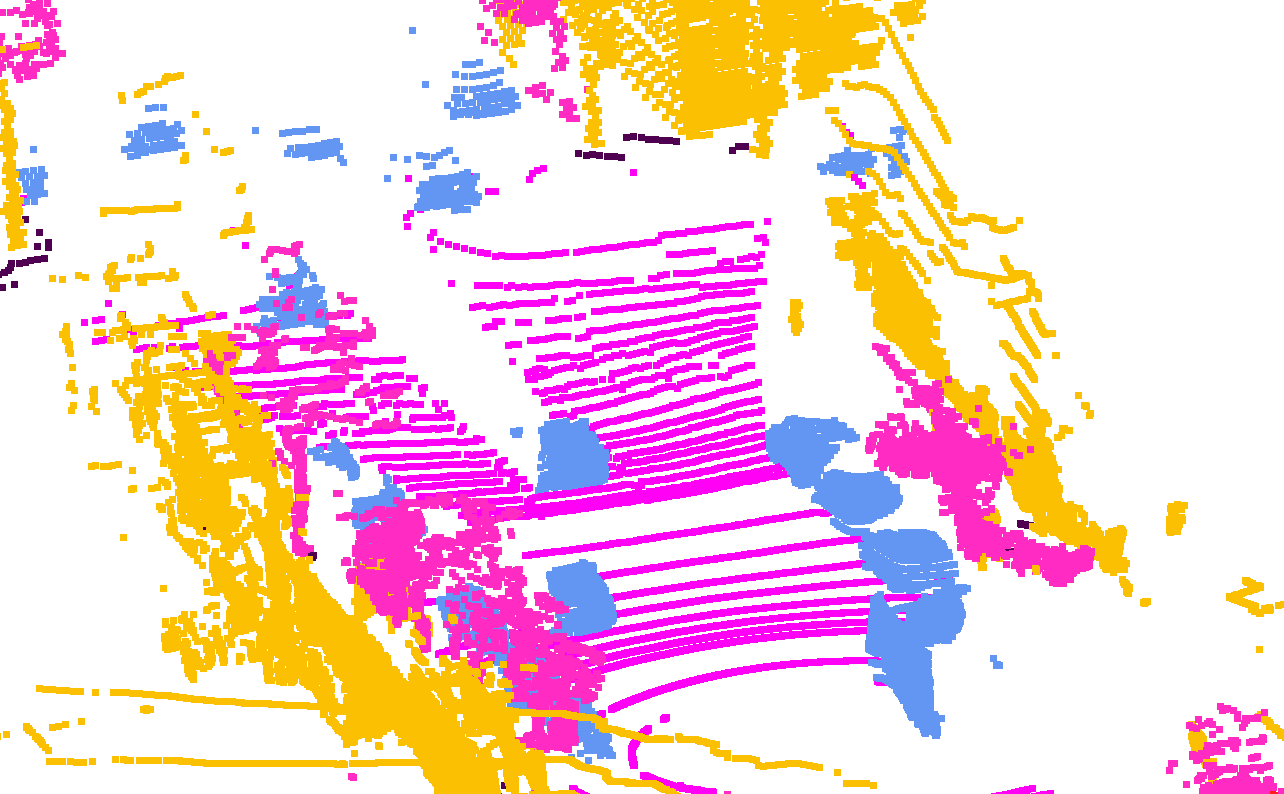}&
\includegraphics[trim=40 0 20 10,clip,width=\widthimage\linewidth]{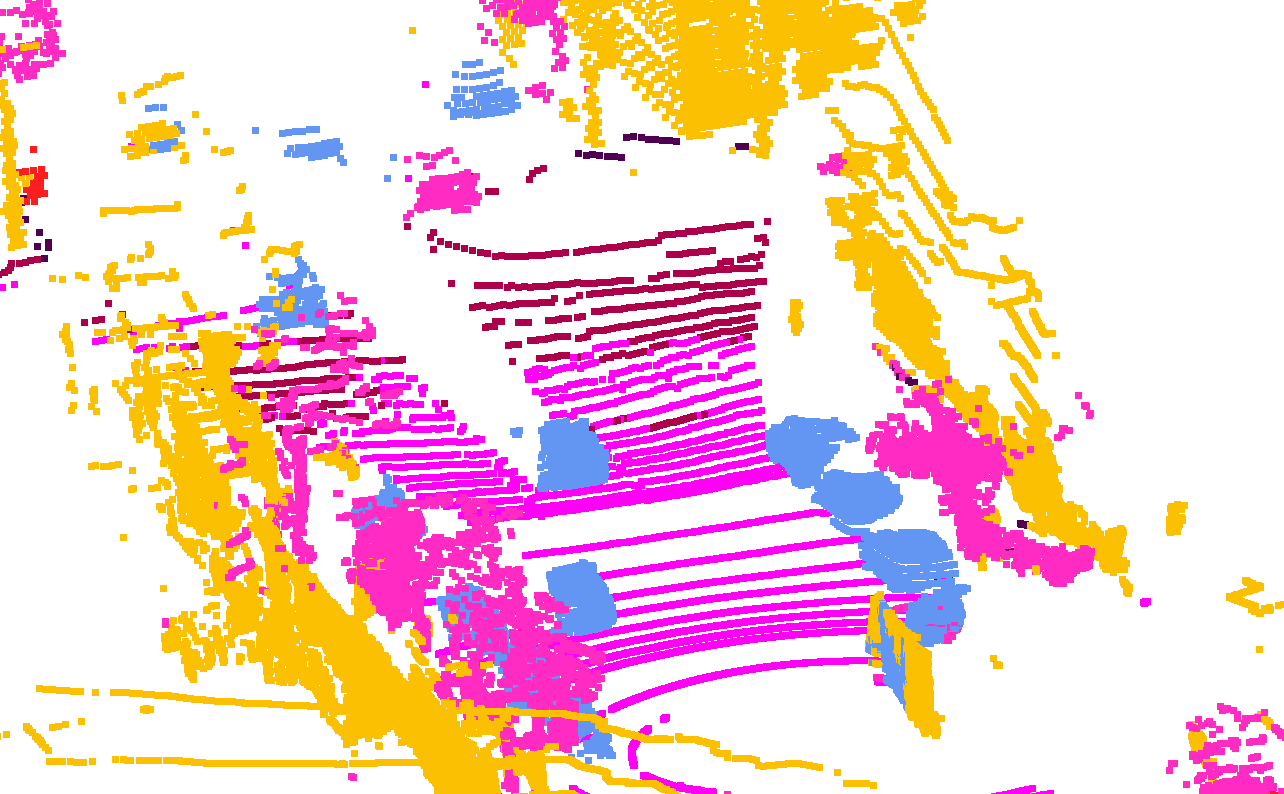}
 & 
\includegraphics[trim=40 0 20 10,clip,width=\widthimage\linewidth]{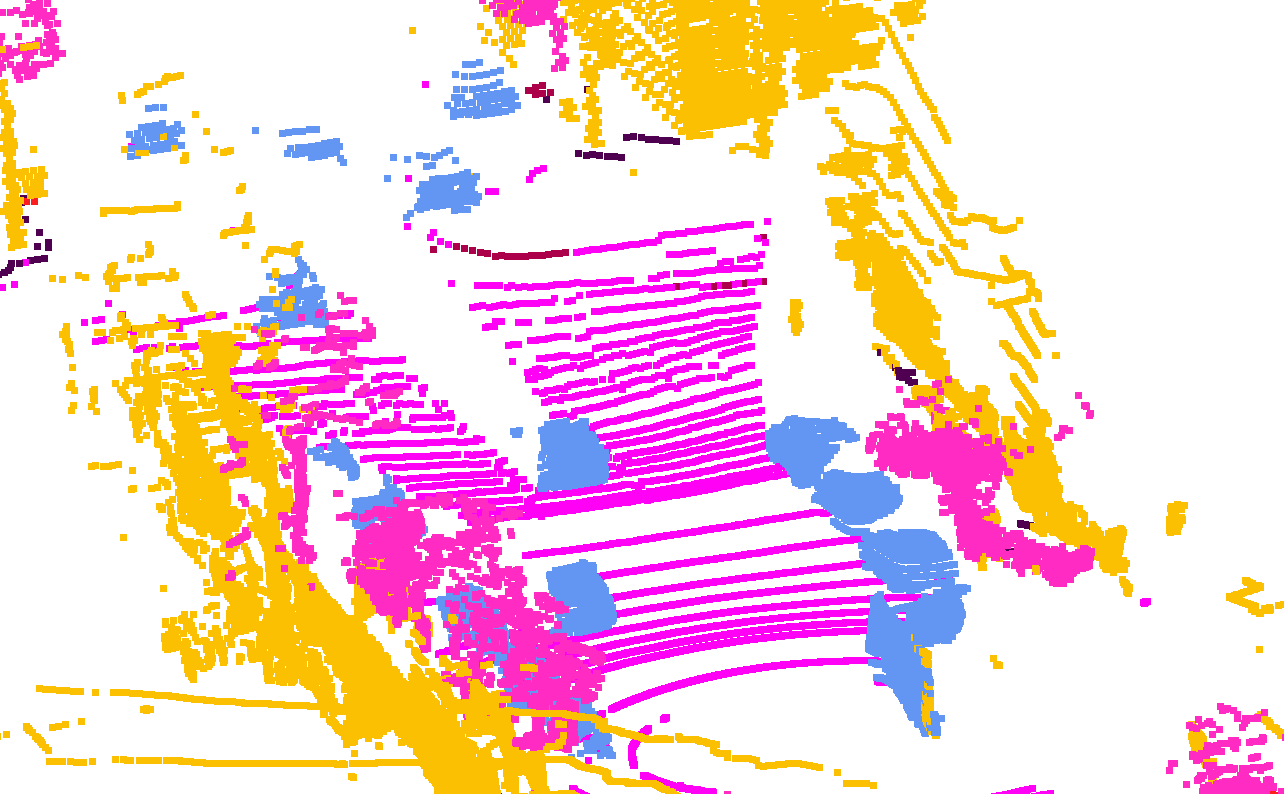}&
\begin{tikzpicture}
    \node[anchor=south west,inner sep=0] at (0,0) {\includegraphics[trim=40 0 20 10,clip,width=\widthimage\linewidth]{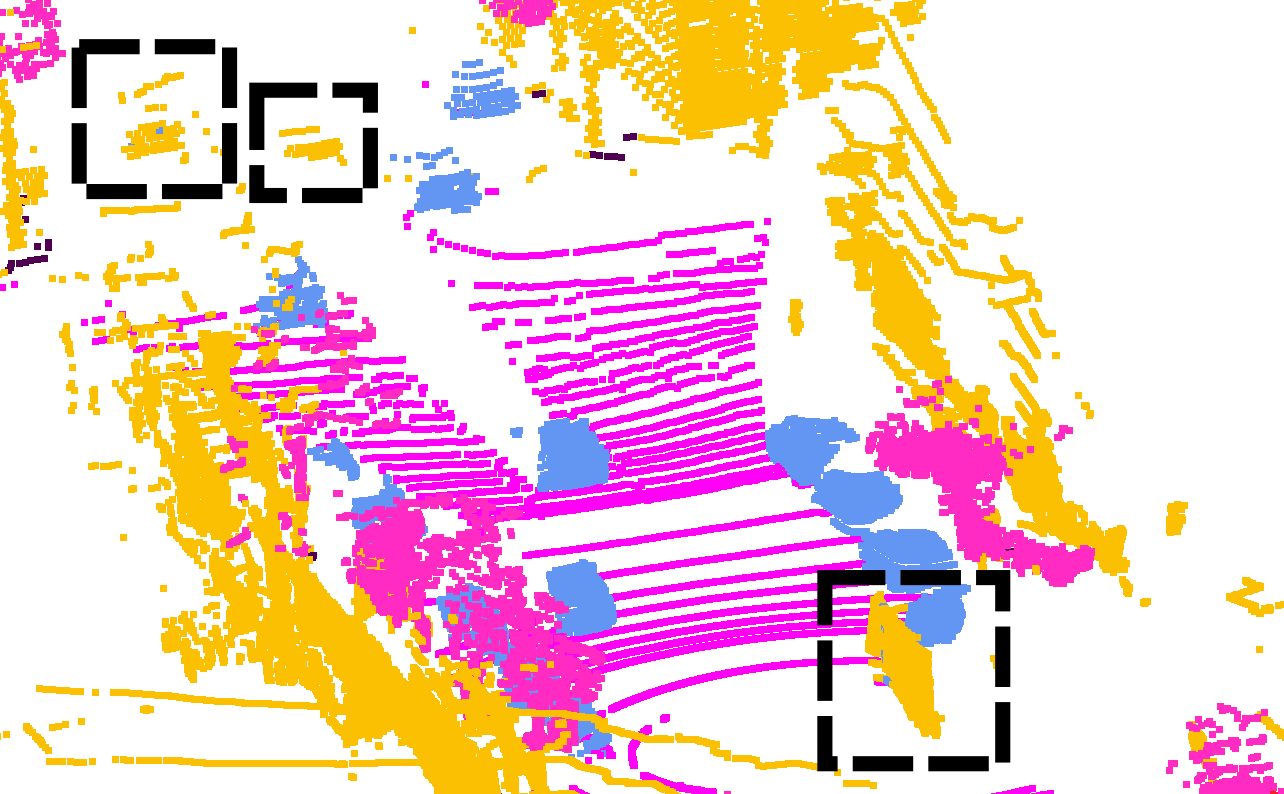}};


        

\end{tikzpicture}

\\
& GT & Src.-only (start point) & \methodstop & After 20k iterations \\

\end{tabular}

\caption{Examples of results with \methodstop: ground truth (GT), initial model trained only on source data, training with our training scheme when using our stopping criterion, and ``full'' training for 20k iterations. ``Ignore'' points are removed for a better visualisation. Notable errors due to degradation are marked with a dashed rectangle. Due to different class mappings, coloring can vary between the different settings.}
\label{fig:app:qualitative_complete_ours}
\end{figure*}

In~\cref{fig:app:qualitative_complete_ours}, we show qualitative results for each domain adaptation setting: ground-truth labels (GT), the source-only result, the result obtained by our training scheme with \methodstop, and the result obtained after 20k iterations. These representations highlight that the stopping criterion achieves a significant, qualitatively visible improvement.

As can be seen, the improvements of our training scheme in combination with our stopping criterion over the source-only model are dominated by changes in the ``Road'', ``Sidewalk'', and ``Terrain'' classes. If the training is pushed to 20k iterations, these large classes are little degraded, while objects of other classes like cars or pedestrians can be totally misclassified. One exception is the \nstoposs{} setting, where we can observe a total collapse into a binary classification after training for 20k iterations.


\clearpage

\section{Datasets and class mappings}
\label{sec:datasets_overview}

\cref{tab:datasets2} summarizes the main characteristics of the datasets we used in experiments, including details about the lidars used for data capture. As can be seen, there is a lot of variety among the lidar sensors, not counting variations that are not even reported here, such as sensor height or laser range. This sensor gap yields significant dissimilarities at point cloud level. Considering on top of that the geographical variety of the driving landscapes over 3 continents, including synthetic scenery, the total domain gap between most of these datasets can be considered as severe.

Note that the number of classes we report is the number used for the standard benchmarking of semantic segmentation on each dataset, which may be lower than the number of finer-grained classes actually annotated in the ground-truth data. Also, for SemanticKITTI, the class
of a moving object is merged with the class of the same static object.

In~\crefrange{tab:app:class_mapping_ns_sk}{tab:app:class_mapping_syn_poss}, we provide the exact class mapping. Unnamed classes are mapped to `Ignore'.

\begin{table*}[b]
    \centering
    \vspace*{-2mm}
\caption{Datasets used in our domain adaptation experiments. For each dataset, we provide: abbreviation in the paper, main reference, lidar sensor used for data capture, number of beams, vertical field of view (V.\,FoV), vertical resolution (V.\,res.), horizontal resolution (H.\,res.), number of classes used for standard benchmarking (which may be lower than the number of finer-grained actually annotated classes), number of frames for training and/or testing, and region of the world where the data was captured. The V.\,FoV of the Pandora (Pandar40) lidar is variable, denser when closer to horizontality: 0.33° for the FoV -6° to +2°, and 1° for the FoV -16° to -6° and +2° to +7°. The V.\,FoV of the Pandar64 is even more variable: 0.167° (-6° to +2°), 1° (-14° to -6°, +2° to +3°), 2° (+3° to +5°), 3° (+5° to +11°), 4° (+11° to +15°), 5° (-19° to -14°), 6° (-25° to -19°).}
\label{tab:datasets2}
    \tabcolsep 0.8mm
\scalebox{0.6}{\begin{tabular}{l@{\,}l@{}rlcrllcrrl}
\toprule
\rowcolor{purple!10}           
Dataset  & & Ref. & Lidar & \!\!\!\!\!Beams\!\!\! & \multicolumn{1}{c}{V.\,FoV} & V.\,res. & H.\,res. & \!\!\!\!\!Classes\!\! & Train & Test~ & Region of the world
\\
\midrule
nuScenes & (NS) & \cite{caesar2020nuscenes} & Velodyne HDL-32E & 32 & -30.7° to +10.7° & 1.33° & 0.33° & 16 & {28,130} & -- & Boston, Singapore
\\
SynLiDAR & (SL) & \cite{xiao2022transfer} & \textit{synthetic} & 64 & -25.0° to +\hphantom{1}3.0° & & & 22 & {19,840} & -- & 3D experts using \\
&&&&&&&&&&&Unreal Engine~4
\\
SemanticPOSS & (SP) & \cite{pan2020semanticposs} & Pandora (Pandar40) & 40 & -16.0° to +\hphantom{1}7.0° & 0.20° & 0,33°/1° & 14 & 2,484 & 499 & Peking University \\
&&&&&&&&&&&(many dynamic objects)
\\
SemanticKITTI & (SK) & \cite{behley2019iccv} & Velodyne 
HDL-64E & 64 & -24.8° to +\hphantom{1}2.0° & 0.42° & 0.18° &  19 & 19,130 & 4,071 & Karlsruhe
\\
Pandaset & (PD) & \cite{xiao2021pandaset} & Pandar64 & 64 & -25.0° to +15.0° & 0.17° & 0.20°/6° & 37 & 3,800 & {2,280} & San Francisco, \\
&&&&&&&&&&&El Camino Real
\\
Waymo Open & (WO) & \cite{Ettinger_2021_ICCV} & Laser Bear Honeycomb & 64 & -17.6° to +\hphantom{1}2.4° & & & 23  & 23,691 & 5,976 & Phoenix, San Francisco, \\
&&&&&&&&&&&Mountain View
\\
\bottomrule
\end{tabular}}

\end{table*}

\begin{table}[!p]

\centering
\begin{minipage}{.5\textwidth}
\captionsetup{justification=centering}
    \captionof{table}{Class mapping \\ for \nstosk\ (from~\cite{yi2021complete}).}
    \label{tab:app:class_mapping_ns_sk}
    \centering
    \scalebox{0.6}{
    \begin{tabular}{c|c|c}
        \toprule
        \rowcolor{violet!10}           
        nuScenes & \nstosk & SemanticKITTI \\ 
         \bottomrule \vphantom{$X^{X^X}$}
        Car & Car &  Car\\
        \midrule
        Bicycle    & Bicycle & Bicycle \\
        \midrule
        Motorcycle & Motorcycle & Motorcycle \\
        \midrule
        Truck   & Truck & Truck \\
        \midrule
        Construction  &  \multirow{ 2}{*}{Other vehicle} &  Other-vehicle, \\
        vehicle, Bus & & Bus \\
        \midrule
        Pedestrian  & Pedestrian & Person \\
        \midrule
        \multirow{ 3}{*}{\shortstack{Driveable \\ Surface}}    & \multirow{ 3}{*}{\shortstack{Driveable \\ surface}} & Road,\\
        & & Parking,\\
        & & Lane marking \\
        \midrule
        Sidewalk   & Sidewalk &  Sidewalk \\
        \midrule
        Terrain    & Terrain &  Terrain\\
        \midrule
        Vegetation  & Vegetation & Vegetation, Trunk\\ 
        \bottomrule
    \end{tabular}
    }
\end{minipage}%
\begin{minipage}{.5\textwidth}
\vspace{-18pt}
\captionsetup{justification=centering}
    \captionof{table}{Class mapping \\ for \nstoposs\ (from~\cite{Sanchez_2023_ICCV}).}
    \label{tab:app:class_mapping_ns_poss}
    \centering
    \scalebox{0.6}{
    \begin{tabular}{c|c|c}
        \toprule  
        \rowcolor{violet!10}                      
            nuScenes & \nstoposs & SemanticPOSS \\ 
             \bottomrule \vphantom{$X^{X^X}$}
            Pedestrian & Person & Person \\
            \midrule
            Bicycle, Motorcycle& Bike & Rider, Bike \\
            \midrule
            Car, Bus, & \multirow{ 3}{*}{Car} & \multirow{ 3}{*}{Car} \\
            Constriction vehicle,&&\\
            Trailer,Truck&&\\
            \midrule
            Driveable surface,  & \multirow{ 2}{*}{Ground} & \multirow{ 2}{*}{Ground} \\
            Other flat, &&\\
             Sidewalk, Terrain&&\\
             \midrule
            Vegetation & Vegetation &  Plants \\
            \midrule
            Barrier,  & \multirow{ 3}{*}{Manmade} & Traffic sign, Pole,  \\
            Manmade,  &&  Garbage can, Building, \\
            Traffic cone & &  Cone/Stone, Fence\\
            \bottomrule
    \end{tabular}
    }   
\end{minipage}

\begin{minipage}{0.5\textwidth}
\vspace{-27pt}
\centering
\captionsetup{justification=centering}
    \captionof{table}{Class mapping \\ for \nstowy\ (from~\cite{kim2023single}).}
    \label{tab:app:class_mapping_ns_wy}
    \scalebox{0.6}{
    \begin{tabular}{c|c|c}
        \toprule
          
        \rowcolor{violet!10}                      
        nuScenes & \nstowy & Waymo Open \\ 
         \bottomrule \vphantom{$X^{X^X}$}
        Pedestrian & Person & Person \\
        \midrule
        Bicycle, Motorcycle& Bike & Rider, Bike \\
        \midrule
        Car, Bus, & \multirow{ 3}{*}{Car} & \multirow{ 3}{*}{Car} \\
        Constriction Vehicle,&&\\
        Trailer,Truck&&\\
        \midrule
        Driveable Surface, & \multirow{ 2}{*}{Ground} & \multirow{ 2}{*}{Ground} \\
         Other Flat, & & \\
         Sidewalk, Terrain&&\\
         \midrule
        Vegetation & Vegetation & Vegetation, Plant \\
        \midrule
        Barrier,  & \multirow{ 3}{*}{Manmade} & Traffic Sign, Pole,  \\
        Manmade,  &&  Garbage Can, Building, \\
        Traffic Cone & &  Cone/Stone, Fence\\
        \bottomrule
    \end{tabular}
    }
\end{minipage}
\begin{minipage}{0.49\textwidth}
\vspace{+45pt}
\captionsetup{justification=centering}
    \captionof{table}{Class mapping \\ for \nstopd\ (from~\cite{Sanchez_2023_ICCV}).}
    \label{tab:app:class_mapping_ns_pd}
    \scalebox{0.6}{
    \begin{tabular}{c|c|c}
        \toprule
        \rowcolor{violet!10}           
        nuScenes & \nstopd & Pandaset \\ 
        \bottomrule \vphantom{$X^{X^X}$}
        & \multirow{ 4}{*}{2-wheeled} & Bicycle, Motorcycle,   \\
        Bicycle,  && Motorized scooter\\
        Motorcycle && Pedicab, \\
        && Personal Mobility Device\\
        \midrule
        \multirow{ 2}{*}{Pedestrian} & \multirow{2}{*}{Pedestrian} & Pedestrian,\\ & &Pedestrian w/ objects \\
        \midrule
        \multirow{ 2}{*}{Driveable ground} & \multirow{ 2}{*}{Driveable ground} & Driveway, Road,\\
        && Road marking \\
        \midrule
        Sidewalk & Sidewalk & Sidewalk\\
        \midrule
        Other flat, Terrain& Other ground & Ground \\
        \midrule
          & \multirow{ 5}{*}{Manmade} &Building, Cones,      \\
        Barrier, && Construction Barriers/Signs, \\
         Manmade,&&  Other static object,\\
        Traffic cone&& Pylons, Road Barriers, \\
        && Rolling containers, Signs \\
        \midrule
        Vegetation & Vegetation & Vegetation\\
        \midrule
           &\multirow{ 5}{*}{4-wheeled} & Car, Construction vehicle,   \\
         Bus, Car,&& Emergency vehicle,\\ 
         Construction vehicle,&&Bus, Towed object, \\
        Trailer, Truck && Truck (all kinds of)\\   \\
        && Uncommon vehicle\\
        \bottomrule
    \end{tabular}
    }
\end{minipage}
\end{table}
\begin{table}[!p]
    \begin{minipage}{0.5\textwidth}
    \captionsetup{justification=centering}
    \captionof{table}{Class mapping for \\\synthtosk\ (from~\cite{saltori2022cosmix}).}
    \label{tab:app:class_mapping_syn_sk}
    \centering
    \scalebox{0.6}{
    \begin{tabular}{c|c}
        \toprule
        \rowcolor{violet!10}           
        SynLiDAR & \synthtosk{} \&  SemanticKITTI \\ 
         \bottomrule \vphantom{$X^{X^X}$}
       Car &Car \\
       \midrule
       Bicycle & Bicycle \\
       \midrule
       Motorcycle & Motorcycle  \\
       \midrule
       Truck & Truck  \\
       \midrule
       Bus, Other vehicle & Other vehicle  \\
       \midrule
       Person & Pedestrian  \\
       \midrule
       Bicyclist & Bicyclist  \\
        \midrule
       Motorcyclist & Motorcyclist  \\
        \midrule
      Road  & Road  \\
       \midrule
      Parking  & Parking  \\
       \midrule
       Sidewalk & Sidewalk  \\
        \midrule
       Other ground & Other ground  \\
        \midrule
      Building  & Building  \\
       \midrule
      Fence  & Fence  \\
       \midrule
      Vegetation  & Vegetation  \\
      \midrule
      Trunk  & Trunk  \\
      \midrule
      Terrain  & Terrain  \\
      \midrule
      Pole  & Pole  \\
      \midrule
      Traffic sign  & Traffic sign \\
        \bottomrule
    \end{tabular}
    }
    \end{minipage}
    \begin{minipage}{0.5\textwidth}
    \centering
    \vspace{-58pt}
    \captionsetup{justification=centering}
    \captionof{table}{Class mapping for \\ \synthtoposs\ (from~\cite{saltori2022cosmix}).}
    \label{tab:app:class_mapping_syn_poss}
    \scalebox{0.6}{
        
    \begin{tabular}{c|c}
        \toprule
          
        \rowcolor{violet!10}           
       
        SynLidar & \synthtoposs{} \&  SemanticPOSS \\ 
         \bottomrule \vphantom{$X^{X^X}$}
      Person  & Person  \\
      \midrule
      Bicyclist,Motorcyclist  & Rider  \\
      \midrule
       Car & Car  \\
       \midrule
       Trunk & Trunk  \\
       \midrule
       Vegetation & Plants \\
       \midrule
      Traffic sign  & Traffic sign  \\
      \midrule
      Pole  & Pole \\
      \midrule
      Garbage can  & Garbage can  \\
      \midrule
      Building  & Building  \\
      \midrule
      Traffic-cone  & Cone  \\
      \midrule
      Fence  & Fence \\
      \midrule
      Bicycle  & Bike  \\
      \midrule
      Road  & Ground  \\

        \bottomrule
    \end{tabular}

    }
    \end{minipage}
        
\end{table}

\end{document}